%% file: main.tex
\documentclass[10pt,journal, final]{IEEEtran}
\usepackage{amsmath,amsfonts}
\usepackage{algorithm}
\usepackage{algpseudocode}
\usepackage{array}
\usepackage{textcomp}
\usepackage{stfloats}
\usepackage{url}
\usepackage{verbatim}
\usepackage{graphicx}
\usepackage{cite}
\usepackage{xcolor}
\usepackage{caption}
\usepackage{subcaption}
\usepackage{multicol}
\usepackage{dsfont}

\renewcommand{\theenumi}{\Alph{enumi}}

\usepackage{booktabs}

\usepackage{amssymb}

\usepackage{mathtools}
\newcounter{tableeqn}[table]

\newcounter{tablesubeqn}[tableeqn]

\newcolumntype{C}[1]{>{\centering\arraybackslash}p{#1}}

\usepackage{pgfplots}
\usetikzlibrary{shapes, shadows, arrows}
\usepackage{svg}  
\usetikzlibrary{plotmarks}
\usetikzlibrary{arrows.meta}
\usepgfplotslibrary{patchplots}

\newtheorem{theorem}{Theorem}
\newtheorem{proposition}{Proposition}

\newtheorem{lemma}{Lemma}
\newtheorem{remark}{Remark}

\pgfplotsset{compat=1.3}

\begin{document}

\title{Consistent Estimation of a Class of Distances Between Covariance Matrices}

\author{Roberto~Pereira,~\IEEEmembership{Member,~IEEE,}
        Xavier~Mestre,~\IEEEmembership{Senior~Member,~IEEE,}
        and~David~Gregoratti,~\IEEEmembership{Senior~Member,~IEEE}
        
\thanks{R. Pereira and X. Mestre are with the Centre Tecnol\`ogic de Telecomunicacions de Catalunya (CTTC), 
Barcelona (Spain), 
\{roberto.pereira, xavier.mestre\}@cttc.cat}
\thanks{D. Gregoratti is with Software Radio Systems, Barcelona, Spain, david.gregoratti@srs.io}

\thanks{Part of this manuscript has been presented to IEEE ICASSP 2023.}

\thanks{This work has been partially supported by the European Commission under the Windmill project (contract 813999) and the Spanish Grant PID2021-128373OB-I00 funded by MCIN/AEI/10.13039/501100011033 and by “ERDF A way of making Europe”.}}

\markboth{IEEE Transactions on Information Theory}%
{Shell \MakeLowercase{\textit{et al.}}: A Sample Article Using IEEEtran.cls for IEEE Journals}

\maketitle
\begin{abstract}
This work considers the problem of estimating the distance between two covariance matrices directly from the data.   Particularly, we are interested in the family of distances that can be expressed as sums of traces of functions that are separately applied to each covariance matrix. 
This family of distances is particularly useful as it takes into consideration the fact that covariance matrices lie in the Riemannian manifold of positive definite matrices, thereby including a variety of commonly used metrics, such as the Euclidean distance, Jeffreys\textquotesingle~divergence, and the log-Euclidean distance. 
Moreover, a statistical analysis of the asymptotic behavior of this class of distance estimators has also been conducted.  Specifically, we present a central limit theorem that establishes the asymptotic Gaussianity of these estimators and provides closed form expressions for the corresponding means and variances. 
Empirical evaluations demonstrate the superiority of our proposed consistent estimator over conventional plug-in estimators in multivariate analytical contexts.
Additionally, the central limit theorem derived in this study provides a robust statistical framework to assess of accuracy of these estimators.

\end{abstract}


\section{Introduction}

Over the past years, there has been a surge of machine learning methods that take advantage of the structure of the data to learn some desired task. 
In general, when a sufficiently large amount of data is available, learning algorithms can approximate the topological space of the data, which is often beneficial.  
Specifically, in multivariate analysis, where samples are continuously generated by one or several processes, exploring the relationships and dependencies among different samples offers a better understanding of the underlying data structures, rather than individually  comparing these samples~\cite{ning2013geometry}. In general,  covariance matrices provide a concise descriptor of the multiple features present in the data (often detectable only across multiple samples, e.g., over time or frequency), making these second-order statistics valuable for data representation in numerous learning tasks, such as 
clustering~\cite{Tiomoko19, Hallac17, nasim2022millimeter}, classification~\cite{chang2003hyperspectral, Richiardi_ml_brain_Graphs, cvpr2012CovDescLearning} and statistical analysis~\cite{zheng2017riemannian, yang2015independence}. In this learning context, a fundamental task is to compare multiple covariance matrices. However, by doing so, one also has to account for the non-conformity of these matrices to the Euclidean space.
Specifically, by considering a more geometric approach, one can also observe covariance matrices as symmetric positive definite (SPD) matrices lying in a Riemann manifold~\cite{thanwerdas2023n}. Consequently, common definitions based on the Euclidean space are no longer applicable to covariance matrices and, when employed to compare these second order statistics,  are  known to lead to inadequate conclusions~\cite{horev2016geometry}.

Hence, recent efforts have been devoted to defining suitable distances between covariance matrices by leveraging the Riemannian geometry of positive definite matrices~\cite{dryden09}. 
Particularly, initial works focused on the affine-invariant Riemannian metric, which, for two covariance matrices $\mathbf{R}_1$, $\mathbf{R}_2$, is defined as 
\begin{equation} 
d^{AI}_M (1,2)= \frac{1}{M} \mathrm{tr}\left[\log^2(\mathbf{R}_2^{-1/2}\mathbf{R}_1\mathbf{R}_2^{-1/2})\right]
\end{equation} 
where the logarithm is applied matrix-wise.
This distance has been extensively employed in several computational methods such as feature coding~\cite{ilea2018covariance} and classification in brain computer interfaces~\cite{Barachant13}. Unfortunately, despite its large applicability, the use of this metric is usually associated to high computational costs, since its application typically requires an intensive use of matrix square roots, inverses and logarithms\footnote{For instance, estimating the Fréchet mean (a generalization of the centroid concept to metric spaces) often requires solving an optimization task~\cite{dryden09}.}.  
Consequently, more recent contributions have focused on developing alternative metrics that retain the desirable geometric properties of SPD matrices while offering improved computational efficiency. This is the case of the log-Euclidean metric proposed in \cite{arsigny06} in the context of diffusion tensor imaging, which is defined as
\begin{equation} 
\label{eq:defLE}
d_{M}^{LE} (1,2)=\frac{1}{M}\mathrm{tr}\left[  \left(  \log\mathbf{R}_{1}%
-\log\mathbf{R}_{2}\right)  ^{2}\right]. 
\end{equation}

The log-Euclidean metric was originally derived by endowing the manifold of positive definite matrices with an appropriate Lie group structure, together with a logarithmic scalar multiplication that gives the essential properties of a vector space \cite{arsigny07}. The result is a metric that, even if it is not affine invariant, is invariant under orthogonal and dilation transformations. Contrary to the affine-invariant metric, the log-Euclidean distance is more amenable from the computational complexity,  has a closed form solution for its (Fréchet) mean and always yields a positive definite Gaussian kernel~\cite{Jayasumana15}. Hence, it is often the distance of choice when comparing different covariance matrices. 
Nonetheless, similar to the distances described above, 
there exist a number of metrics which leverage  Riemann geometry to compare covariance matrices or SPD matrices in general. These metrics have been widely studied to understand their computational and geometrical properties~\cite{thanwerdas2023n}.

A general  challenge in applying these second-order learning approaches is the fact that covariance matrices are generally unknown,  
and consequently, the inherent distances must be estimated from the corresponding data. 
A straightforward solution is to \textit{plug-in} (replace) the true covariance matrix with its respective sample covariance matrix (SCM) estimated directly from the observations. When an infinitely large number of samples is available and the  observation dimension is fixed, the SCMs can closely approximate the eigenstructure  of the true (often called population) covariance matrices describing the system. Consequently, in these scenarios, the distance obtained by using the SCMs in the original definition may also be a close approximation of the true distance. 
Unfortunately, this ideal setting is only possible in stationary systems. In the more realistic scenario where observations are obtained from one or several non-stationary systems, the number of available samples is often not much larger than the corresponding observation dimension. As a result, the SCMs are only approximations of the true covariance matrix~\cite{marchenko1967distribution, bai2008clt} such that the traditional \textit{plug-in} estimators tend to poorly approximate the true distance~\cite{zarrouk2020performance, roberto21_globecom, pereira23tsp}. These poor estimations  often lead to undesired behaviors -- e.g., the distance between two sample covariance matrices generated by the same process may be larger than the distance between the ones generated by different processes~\cite{roberto22_icassp}.

One possible way to address this issue it to try and improve covariance matrix estimation in the presence of limited samples and high dimensionality. Typical approaches often involve considering some structure on the actual covariance matrix~\cite{vinogradova2015estimation}, 
some type of regularization~\cite{ieva16}, both of which often require some assumption on  the data. 
An alternative is to try and directly estimate the distance between (sample) covariance matrices when both sample size and observation dimension are large but comparable in magnitude, as in~\cite{Tiomoko19, tiomoko2019random,couillet2019random,pmlr-v97-tiomoko19a}. In these contributions, the main objective was to study the family of distances between covariances that can be expressed as functions of $\mathbf{R}_1^{-1}\mathbf{R}_2$. Unfortunately, there exist important distances such as the log-Euclidean metric 
that do not fall into this category. One of the objectives of this paper is to follow a similar approach and consider consistent estimators of a different class of metrics, namely the class of metrics that can be expressed as sums of traces of functions that are separately applied to each covariance, which include the log-Euclidean distance and the symmetrized Kullback-Leibler (KL) divergence. Differently from the straightforward \textit{plug-in} estimator described above, these consistent estimators are designed to approximate the true distance even when the sample sizes are of the same order of magnitude as the observation dimension.

 Part of the results presented here were published without complete proofs in \cite{roberto23_icassp}.  In this work, we consider the same family of distances and further elaborate on our previous contribution by establishing a Central Limit Theorem (CLT) that characterizes the asymptotic distribution of these consistent estimators.  
 Furthermore, extended proofs of both the closed form expressions for the consistent estimators (Appendix \ref{sec:AppendixIntegralLog}) and their asymptotic means and variances (Appendices  \ref{sec:proofclt} and \ref{sec:simplifiedIntegralsVar}) are provided. 
 Some of the derivations of the second part of the paper strongly rely on the CLT established in~\cite{pereira23tsp} for the \textit{plug-in} estimators. In spite of the formal similarity between the two problems, some work is needed in order to transform the proposed consistent estimators into a suitable form (up to error terms rapidly converging to zero) so that the results in~\cite{pereira23tsp} can be applied. Interestingly enough, the obtained CLT turns out to be simpler than the one in~\cite{pereira23tsp}, in the sense that the asymptotic means and variances have more compact analytical forms.

\section{Statistical Model of the Observations and Family of Distances}
\label{sec:preliminaries}

Let us consider two sets of multivariate observations associated to the covariance matrices $\mathbf{R}_1$ and $\mathbf{R}_2$, of size $M \times M$. We assume that each one of these observation sets respectively consists of $N_1, N_2$ observations of dimensionality $M$, which we denote by $\mathbf{y}_j(n) \in \mathbb{K}^{M}, n =1,\ldots,N_j $, $j \in \{1,2\}$ with 
$\mathbb{K} \in \{\mathbb{R},\mathbb{C}\}$
depending on whether the observations 
are real- or complex-valued. Moreover, let us also define the matrix  $\mathbf{Y}_j$ of size $M \times N_j$,  which contains the
observations associated to each of these processes as columns, that is
\[
\mathbf{Y}_{j}=\left[
\begin{array}
[c]{ccc}
\mathbf{y}_{j}(1) & \ldots & \mathbf{y}_{j}(N_{j})
\end{array}
\right]
\]
for $j\in\{1,2\}$. We are interested in the family of  distances\footnote{Notice that in order for $d_M$ to be a distance, it has to satisfy some properties (e.g., positiveness and triangle inequality). It is clear that there might exist choices of $f_1, f_2$ that will lead to some $d_M$ that will not satisfy these properties. In this work, however, we loosely denote the family of functions $d_M$ as a family of (squared) distances without necessarily worrying about asserting these properties. } between  covariance matrices that can be mathematically
expressed as
\begin{equation}
{d}_{M}(1,2)=\sum_{l=1}^{L}\frac{1}{M}\mathrm{tr}\left[  f_{1}^{(l)}\left(
\mathbf{{R}}_{1}\right)  f_{2}^{(l)}\left(  \mathbf{{R}}_{2}\right)
\right]  
\label{eq:family_distances}
\end{equation}
for certain functions $f_{1}^{(l)}, f_{2}^{(l)}$
, $l=1,\ldots,L$ (see also \textbf{(As4)} in Section~\ref{sec:consistent_estimator} for details). In general terms, these functions can be understood as scalar functions applied to the eigenvalues of the Hermitian matrices ${\mathbf{R}}_j$, $j \in \{1,2\}$, so that (with some abuse of notation) we will also denote by $f_{1}^{(l)}, f_{2}^{(l)}: \mathbb{R}\rightarrow\mathbb{R}$ the scalar version directly applied the eigenvalues. 
Let us particularize this definition to some meaningful choices for the considered distances.

This family of functions covers a large class of covariance distances, including functionals of the type
$$
{d}_{M}^{f}(1,2)=\frac{1}{M}\mathrm{tr}\left[  \left( f_{1}(\mathbf{{R}}_{1})  - f_{2}(  \mathbf{{R}}_{2}) \right)^2 \right]
$$
which includes multiple generalizations of the conventional Euclidean distance. Particularly, for $f_j(\mathbf{R}_j) = \mathbf{R}_j$ we recover the traditional Euclidean distance between covariance matrices.  
Likewise, the choice $f_{j}({\mathbf{R}}_j)=\log ({\mathbf{R}}_j)$ will lead
to the log-Euclidean distance as defined in (\ref{eq:defLE}).
Finally, after proper normalization,  the symmetrized version of the Kullback-Leibler divergence between two multivariate Gaussians  (usually referred to as Jeffreys' divergence \cite{Ali66}) can be expressed by
\begin{equation}
{d}_M^{KL}(1,2) = \frac{1}{2M}\mathrm{tr}\left[\mathbf{{R}}_{1}\mathbf{{R}}_{2}^{-1} \right] + \frac{1}{2M}\mathrm{tr}\left[\mathbf{{R}}_{2}\mathbf{{R}}_{1}^{-1} \right] - 1 \label{eq:defJefferiesdivergence}
\end{equation}
which also conforms to the general expression in (\ref{eq:family_distances}).  
We emphasize that the quantities above are not the only ones that can be obtained from~(\ref{eq:family_distances}), instead, they are used here as illustrative examples of how broad this definition is (see, for instance, the power-Euclidean distance in \cite{dryden2010power}, the family of $(\alpha, \beta)$-divergences in~\cite{zhang2004divergence} and other definitions in~\cite{couillet2019random, pereira23tsp}). 

Our estimators will be built on the \emph{sample} covariance matrices $\hat{\mathbf{R}}_1$, $\hat{\mathbf{R}}_2$, where
\begin{equation}
    \hat{\mathbf{R}}_j = \frac{1}{N_j}\mathbf{Y}_j\mathbf{Y}_j^H,
\end{equation}
for $j \in \{1,2\}$. In this scenario, the family of distances defined in (\ref{eq:family_distances}) can be trivially estimated as 
\begin{equation}
    \tilde{d}_{M}(1,2)=\sum_{l=1}^{L}\frac{1}{M}\mathrm{tr}\left[  f_{1}^{(l)}\left(
\hat{\mathbf{R}}_{1}\right)  f_{2}^{(l)}\left(  \hat{\mathbf{R}}_{2}\right)
\right]  
\label{eq:definitionhatd}
\end{equation}
where we have replaced $\mathbf{R}_j$ by their sample estimators $\hat{\mathbf{R}}_j$ obtained directly from the data. Hereafter we will refer to (\ref{eq:definitionhatd}) as the traditional \textit{plug-in} distance estimator. As described above, one of the main problems with this approach is that, for $N_1, N_2$ large and comparable to $M$, we often find that $\tilde{d}_M(1,2)$ is an inconsistent estimator of $d_M(1,2)$. 

The sample covariance matrix is an object that has received a lot of attention in the random 
matrix theory literature. For example, the empirical eigenvalue distribution of $\hat{\mathbf{R}}_j$ is know to converge almost surely to a non random limit when both $M,N_j$ tend to infinity at the same rate \cite{marchenko1967distribution,silverstein95}. More recent 
contributions have formulated central limit theorems of linear spectral statistics of this matrices, both for the Gaussian \cite{bai2008clt} and the non-Gaussian context \cite{Najim16} and extensions have been proposed when the different observations present a certain degree of correlation (doubly correlated models) \cite{Bai19double}. When the observations follow a
Gaussian distribution, one can systematically use Gaussian tools (i.e. the integration by parts formula and the Poincaré-Nash inequality) to derive central limit theorems in a variety of contexts. This was the approach followed in \cite{pereira23tsp}, which studied the fluctuations of the statistic in (\ref{eq:definitionhatd}) in the large dimensional regime. This paper goes one step further and studies consistent estimators of the original distances in (\ref{eq:family_distances}).

More specifically, the general objectives of this paper can be summarized as follows.
\begin{enumerate}
\item For a certain collection of scalar analytic functions $f_{1}^{(l)}$ and $f_{2}^{(l)}$, propose an estimator of $d_{M}(1,2)$, denoted by $\hat{d}_{M}(1,2)$, that is consistent (i.e., $d_{M}(1,2)-\hat{d}_{M}(1,2)\rightarrow0$) when the observation dimension ($M$) and the number of samples per observation ($N_{1},N_{2}$) grow to infinity at the same rate.

\item Assuming that the data is Gaussian distributed, describe the asymptotic
fluctuations of $\hat{d}_{M}(1,2)$ around $d_{M}(1,2)$.

\item Using the above description of the asymptotic fluctuations, provide a statistical framework to study the suitability of a metric on performing a clustering task. 

\end{enumerate}
We develop each of these objectives in the following sections aiming to gain a comprehensive understanding of their characteristics and applications. 
For the sake of clarity, we report only the details that are necessary to make this paper self-contained\footnote{The code and other experiments related to this publication are publicly available in \url{https://github.com/robertomatheuspp/RMTClustering}.}.

\section{Consistent estimator of $d_{M}(1,2)$}
\label{sec:consistent_estimator}

The  objective of this section is to 
propose a general consistent estimator $\hat{d}_M(1,2)$ of the functional ${d}_{M}(1,2)$ in its broadest form. Given the general form of the corresponding consistent estimator $\hat{d}_M(1,2)$, we will particularize this result to several interesting choices of metrics, namely the Euclidean, symmetrized Kullback-Leibler, and log-Euclidean distances between covariance matrices. 
We will rely on the following assumptions.

\noindent\textbf{(As1)} The observation matrix $\mathbf{Y}_{j}$ can be expressed as $\mathbf{Y}_{j}~=~\mathbf{R}_{j}^{1/2} \mathbf{X}_{j}$ for $j \in \{1,2\}$, where $\mathbf{X}_j$ are $M \times N_j$ random matrices with all entries being independent and identically distributed with zero mean and unit variance. 
Moreover, we introduce a binary variable $\varsigma$ to distinguish between real- and complex-valued observations. Specifically, when $\varsigma = 1$,  the observations are real-valued. In contrast,  for $\varsigma = 0$,  the observations are complex random variables with real and imaginary parts that are independent and identically distributed with zero mean and variance $1/2$.

\noindent\textbf{(As2)} The set of distinct eigenvalues of $\mathbf{R}_{j}$ are denoted by $0<\gamma_{1}^{(j)}<\ldots<\gamma_{\bar{M}_{j}}^{(j)}$ ($j=1,2$)\ and have
multiplicity $K_{1}^{(j)},\ldots,K_{\bar{M}_{j}}^{(j)}$, where $\bar{M}_{j}$
represents the total number of distinct eigenvalues. These quantities can differ\footnote{In other words, the results in this paper hold regardless of how the eigenvalues, their number or their multiplicities evolve for increasing $M$, as long as the eigenvalues themselves belong to a compact set of $\mathbb{R}^+$ independent of $M$. In particular, $\bar{M}_j$, $\gamma_{l}^{(j)}$ and $K_l^{(j)}$ may vary with $M$.} for different values of $M$, but we always have $\sup_{M} \gamma_{\bar{M}_j}^{(j)}<\infty$ and $\inf_{M}\gamma_{1}^{(j)}>0$.

\noindent\textbf{(As3)} The number of samples $N_j$ for each observation set depends on the dimensionality $M$. Specifically, as $M$ tends to infinity, $N_j$ also tend to infinity in such a way that the ratio $c_{M,j}=M/N_j$ converges to a constant  $0<c_j<\infty$, where  $c_j \neq 1$. Furthermore we will assume that either $\sup_M c_{M,j}<1$ (oversampled regime) or $\inf_M c_{M,j}>1$ (undersampled regime).

In order to introduce the general form of the consistent estimator of $d_M$, we need to recall some common definitions in the context of random matrix theory. Let us start by considering the quantity $\omega_{j}\left(  z\right)$, which is given by one of the solutions to the polynomial equation
\begin{equation}
z=\omega_{j}\left(  z\right)  \left(  1-\frac{1}{N_{j}}\sum_{m=1}^{\bar{M}
_{j}}K_{m}^{(j)}\frac{\gamma_{m}^{(j)}}{\gamma_{m}^{(j)}-\omega_{j}\left(
z\right)  }\right)  . \label{eq:defw(z)}
\end{equation}
More specifically, if $z\in\mathbb{C}^{+}$ (upper complex semiplane), $\omega_{j}\left(  z\right)  $ is the only solution located in $\mathbb{C} ^{+}$. If $z\in\mathbb{C}^{-}$(lower complex semiplane), $\omega_{j}\left(z\right)  $ is the only solution in $\mathbb{C}^{-}$. Finally, if $z$ is real valued, $\omega_{j}\left(  z\right)  $ is defined as the only real valued solution such that
\begin{equation}
\frac{1}{N_{j}}\sum_{m=1}^{\bar{M}_{j}}K_{m}^{(j)}\left(  \frac{\gamma_{m}^{(j)}}{\gamma_{m}^{(j)}-\omega_{j}\left(  z\right)  }\right)  ^{2}<1.
\label{eq:conditionwrealvalued}
\end{equation}
It is well known that function $\omega_j(z)$ is a key element that allows us to describe the asymptotic behavior of the eigenvalues and eigenvectors of $\hat{\mathbf{R}}_j$ under the above set of assumptions.  
Indeed, consider the two resolvents
\begin{align}
\hat{\mathbf{Q}}_{j}(z) &= \left(\hat{\mathbf{R}}_{j}-z\mathbf{I}_{M}\right)^{-1} \\
 \mathbf{Q}_{j}(\omega)  &=\left(  \mathbf{R}_{j}-\omega\mathbf{I}_{M}\right)^{-1}.
\end{align}
It can be shown from~\cite{silverstein95} that, if $\mathbf{A}_{M}$ denotes a sequence of
deterministic $M\times M$ matrices with bounded spectral norm, we have 
\begin{equation}
\frac{1}{M}\mathrm{tr}\left[  \mathbf{A}_{M}\hat{\mathbf{Q}}_{j}(z)\right]
-
\frac{1}{M}\mathrm{tr}\left[  \mathbf{A}_{M}\mathbf{\bar{Q}}_{j}(z)\right]
\rightarrow 0
\label{eq:almostSureConvergenceTr}
\end{equation}
for any fixed $z \in \mathbb{C}^+$, where 
\begin{equation}
\mathbf{\bar{Q}}_{j}\left(  z\right)
=\frac{\omega_{j}\left(  z\right)  }{z}\mathbf{Q}_{j}\left(  \omega_{j}\left(
z\right)  \right).
\label{eq:defQbar}
\end{equation}
This result is particularly useful and basically states that, for any fixed $z$ outside the real axis, $z \hat{\mathbf{Q}}_{j}(z)$ is asymptotically equivalent to a  $\omega \mathbf{Q}_{j}(\omega)$ after an appropriate
change of variables.

\begin{remark}
\label{remark:asymptEq}
For two sequences of $M \times M$ matrices ${\mathbf{T}}_{M}$ and ${\mathbf{S}}_{M}$ of increasing dimension, we will say that they are asymptotically equivalent (and write ${\mathbf{T}}_{M}\asymp{\mathbf{S}}_{M}$)\ if $M^{-1}\mathrm{tr}[  \mathbf{A}_{M}(  {\mathbf{T}}_{M}-{\mathbf{S}}_{M})  ]  \rightarrow 0$ as $M\rightarrow\infty$ for any sequence of $M\times M$ matrices  $\mathbf{A}_{M}$ with bounded spectral norm.
\end{remark}

In particular, the above asymptotic characterization can be written in short as $\hat{\mathbf{Q}}_j(z) \asymp \bar{\mathbf{Q}}_j(z)$ almost surely. 
This asymptotic equivalence can be immediately used to find an estimator of $d_{M}$ in (\ref{eq:family_distances}) that is consistent under \textbf{(As1)}-\textbf{(As3)}. To that effect, we will further assume that the functions $f_{j}^{(l)}(\omega)$ are sufficiently regular, in the sense that they are analytic in a sufficiently large region of the complex plane. One possibility would be to assume analycity in $\mathbb{C}\backslash\mathbb{R}^{-}$ (that is, the whole complex plane except for the negative real axis). However, in practice this would rule out a number of situations in which we can achieve a consistent estimator even if the number of available samples is lower than the observation dimension, that is $N_{j}<M$\ (undersampled case). So, instead of that, we will consider analytical functions on the whole complex plane except for only a subset of the negative real axis.

\noindent\textbf{(As4)} For $j\in\{1,2\}$ and $l=1,\ldots,L,$ the functions $f_{j}
^{(l)}(\omega)$ are analytic on the set $\mathbb{C}\backslash(-\infty
,\mu_{\inf}^{(j)}]$, where $\mu_{\inf}^{(j)}=\inf_{M}\mu_{0}^{(j)}$ and
$\mu_{0}^{(j)}$ is the smallest solution to the equation
\begin{equation} \label{eq:equationmus}
0=\mu\left(  1-\frac{1}{N_{j}}\mathrm{tr}\left[  \mathbf{R}_{j}\mathbf{Q}
_{j}\left(  \mu\right)  \right]  \right)  .
\end{equation}
In particular, if $\sup_{M}M/N_{j}<1$ (oversampled regime) we have $\mu_{\inf}^{(j)}=0$ whereas
$\mu_{\inf}^{(j)}<0$ if $\inf_{M}N_{j}/M>1$ (undersampled regime).

A couple of comments regarding \textbf{(As4)} are in order. First of all, \textbf{(As4)} is implicitly stating that in the undersampled regime  ($\mu_{\inf}^{(j)}<0$) the functions $f_{j}^{(l)}(\omega)$ are required to be analytic at the origin. This is in fact quite restrictive, since some important distance metrics such as the KL metric (for which $f_{j}^{(l)}(\omega)=\omega^{-1}$ for some $l$) or the log-Euclidean metric (for which $f_{j}^{(l)}(\omega)=\log\omega$ for some $l$) do not have this property. This is in fact a well-known limitation of the proposed approach.
On the other hand, the regularity of the functions $f_{j}^{(l)}(\omega)$ outside $(-\infty,\mu_{\inf}^{(j)}]$ can be somewhat relaxed by considering a wider class of sufficiently smooth functions (not necessarily analytic) on the same region. However, we choose to keep analycity here in order to simplify the proofs and because this property clearly holds for all the covariance distance measures presented above.

\subsection{Consistent Estimator}

Under the above assumptions, we can now express (by Cauchy integration, and using the
analycity of $f_{j}^{(l)}(\omega)$)
\begin{equation}
f_{j}^{(l)}\left(  \mathbf{R}_{j}\right)  =\frac{1}{2\pi\mathrm{j}}
\oint\nolimits_{\mathrm{C}_{\omega_j}^{(l)}}f_{j}^{(l)}(\omega)\mathbf{Q}
_{j}\left(  \omega\right)  d\omega\label{eq:f(R)integralform}
\end{equation}
where $\mathrm{C}_{\omega_j}^{(l)}$ is a negatively oriented simple closed contour
enclosing all the eigenvalues of $\mathbf{R}_{j}$ and which does not enclose the point $\{\mu_{0}^{(j)}\}$. This choice of contour ensures that $\mathrm{C}_{\omega_j}^{(l)}$ remains within the region of analycity of $f_{j}^{(l)}(\omega)$. 

\begin{remark}
    Unless stated otherwise, all the contours in this paper are assumed to be negatively (counter-clockwise) oriented. For simplicity, this will not be reflected in the notation in what follows. 
\end{remark}

Moreover, it has been shown~\cite{Mestre08TIT} that a particularly useful way of parametrizing this contour is by using the function $\omega_{j}\left(  z\right)  $ introduced above. Indeed, it is well known that we can build $\mathrm{C}_{\omega_j}^{(l)}$ as $\mathrm{C}_{\omega_j}^{(l)}=\omega_{j}(  \mathrm{C}_j^{(l)}) $ where the contour $\mathrm{C}_j^{(l)}$
does not enclose $\{0\}$ and encloses the interval  (independent of $M$) $\mathcal{S}_{j}=[\theta_{j}^{-},\theta_{j}^{+}]$, where
$\theta_{j}^{-}=\inf_{M} [ \gamma_{1}^{(j)}\, (  1-\sqrt{c_{M,j}})^{2}  ]$
and
$\theta_{j}^{+}=\sup_{M}[\gamma_{\bar{M}_j}^{(j)}\, (  1+\sqrt
{c_{M,j}})^{2}]. 
$ 
In this case, $\mathrm{C}_{\omega_j}^{(l)}=\omega_{j}(  \mathrm{C}_j^{(l)})  $ has the required properties, namely it is a negatively oriented contour that encloses all the eigenvalues of $\mathbf{R}_{j}$ and not $\mu_{0}^{(j)}.$ 
In addition to this, it can be seen \cite{Mestre08TIT} that for $N_{j}<M$ (undersampled situation) the contour $\mathrm{C}_{\omega_j}^{(l)}$ encloses\footnote{
This is the main reason why we need the functions $f_j^{(l)}(\omega)$ to be analytic in a neighbourhood of zero (or, more precisely, outside $(-\infty,\mu_{\inf}^{(0)})$) in the undersampled regime, which effectively limits the existence of the proposed consistent estimators for the KL and the log-Euclidean distances unless $N_j>M$.} the point $\{0\}$, whereas if $N_{j}>M$ (oversampled situation)\ the contour $\mathrm{C}_{\omega_j}^{(l)}$ does not enclose $\{0\}$. In either one of these scenarios, thanks to \textbf{(As4)}, we can always express (\ref{eq:f(R)integralform}) as 
\begin{equation}
f_{j}^{(l)}\left(  \mathbf{R}_{j}\right)  =\frac{1}{2\pi\mathrm{j}}\oint\nolimits_{\mathrm{C}_j^{(l)}}f_{j}^{(l)}(\omega_{j}\left(  z\right))\mathbf{Q}_{j}\left(  \omega_{j}\left(  z\right)  \right)  \omega_{j}^{\prime}\left(  z\right)  dz \label{eq:identityFintegral}
\end{equation}
where $\omega_{j}^{\prime}\left(  z\right)$ denotes the derivative of $\omega_{j}\left(  z\right)$. 
Moreover, since the contour $\mathrm{C}_j^{(l)}$ is chosen independently of $M$,  
we can derive asymptotic equivalents for $f_{j}^{(l)}\left( \mathbf{R}_{j}\right)$ by studying the asymptotic equivalents of the quantities inside the integrand of (\ref{eq:identityFintegral}). 
The following proposition formulates the result, which essentially asserts the existence of a function of the sample covariance matrix that is asymptotically equivalent to $f_{j}^{(l)}(\mathbf{R}_{j})$ 
with probability one.

\begin{proposition}
\label{prop:generalEstimator}Under \textbf{(As1)}-\textbf{(As4)} we have
\begin{equation}
f_{j}^{(l)}\left(  \mathbf{R}_{j}\right)  \asymp 
\frac{1}{2\pi\mathrm{j}}\oint
\nolimits_{\mathrm{C}_j^{(l)}}\hat{h}_{j}^{(l)}(z)\hat{\mathbf{Q}}_{j}(z)dz,
\label{eq:asymptEqF}
\end{equation}
almost surely. Here, $\hat{h}_{j}^{(l)}(z)$ denotes the random function
\begin{equation}
\hat{h}_{j}^{(l)}(z)=f_{j}^{(l)}\left(\hat{\omega}_{j}\left(  z\right)  \right)\frac
{z}{\hat{\omega}_{j}\left(  z\right)  }\hat{\omega}_{j}^{\prime}\left(
z\right) \label{eq:defhhat}
\end{equation}
where $\hat{\omega}_{j}\left(  z\right)  $ denotes the consistent estimator of
$\omega_{j}\left(  z\right)  $ given by
\begin{equation}
\hat{\omega}_{j}\left(  z\right)  =\frac{z}{1-\frac{1}{N_{j}}\mathrm{tr}
\left[  \hat{\mathbf{R}}_{j}\hat{\mathbf{Q}}_{j}(z)\right]  }
\label{eq:defomegahat}
\end{equation}
and where $\hat{\omega}_{j}^{\prime}\left(  z\right)  $ represents its derivative,
namely
\begin{equation}
\hat{\omega}_{j}^{\prime}\left(  z\right)  =\frac{1-\frac{M}{N_{j}}+z^{2}
\frac{1}{N_{j}}\mathrm{tr}\left[  \hat{\mathbf{Q}}_{j}^{2}(z)\right]
}{\left(  1-\frac{1}{N_{j}}\mathrm{tr}\left[  \hat{\mathbf{R}}_{j}
\hat{\mathbf{Q}}_{j}(z)\right]  \right)  ^{2}}. \label{eq:defomegahatprime}
\end{equation}
Furthermore, the right hand side of (\ref{eq:asymptEqF}) has bounded spectral norm with probability one for all $M$ sufficiently large. 
\end{proposition}

\begin{IEEEproof}
See Appendix \ref{sec:proofPropEstimator}.
\end{IEEEproof}

Proposition \ref{prop:generalEstimator} provides in (\ref{eq:asymptEqF}) the
basic piece that can be used to build estimators of the general distances in
the form of (\ref{eq:family_distances}). Indeed, it is a direct consequence of
Proposition  \ref{prop:generalEstimator} that $d_{M}(1,2)-\hat{d}_{M}(1,2)\rightarrow0$ almost surely, where
\begin{align}
\hat{d}_{M}& (1,2)  =\sum_{l=1}^{L} \frac{-1}{4\pi^2} \oint\nolimits_{\mathrm{C}_1^{(l)}}\oint
\nolimits_{\mathrm{C}_2^{(l)}} \hat{h}_1(z_1) \hat{h}_2(z_2)\times
\nonumber \\
&\hspace{3em} \times \frac{1}{M}\mathrm{tr}\left[ 
 \hat{\mathbf{Q}}_{1}(z_1)  
\hat{\mathbf{Q}}_{2}(z_2)  \right]  dz_1 dz_2
\nonumber \\ 
&=\sum_{l=1}^{L} \frac{-1}{4\pi^{2}}\oint\nolimits_{\mathrm{C}_1^{(l)}}\oint
\nolimits_{\mathrm{C}_2^{(l)}}  f_{1}^{(l)}(\hat{\omega}_{1}(  z_{1}  )) f_{2}^{(l)}(\hat{\omega}_{2}( z_{2})) \times
\nonumber \\
&
\hspace{1em}\times \frac{1}{M}\mathrm{tr}\left[  \hat{\mathbf{Q}}_{1}(z_{1})\hat{\mathbf{Q}}_{2}(z_{2}))\right]  \frac{z_{1}z_{2}\hat{\omega}_{1}^{\prime}\left(  z_{1}\right)  \hat{\omega}_{2}^{\prime}\left(  z_{2}\right)}{\hat{\omega}_{1}\left(  z_{1}\right)  \hat{\omega}_{2}\left(  z_{2}\right)}dz_{1}dz_{2}. 
\label{eq:dhatintegral}
\end{align}
Hence, $\hat{d}_{M}(1,2)$ provides a general expression for a consistent estimator
of a functional having the form in (\ref{eq:family_distances}). Interestingly enough, this general expression can be particularized to the most common distances in the literature. This will be further evaluated in the next subsections. We begin with the most conventional Euclidean and symmetrized KL distances, which are considered here in order to illustrate the integration tricks that can be used to solve the above integrals. The
consistent estimator for the log-Euclidean distance requires some additional
work and is provided at the end of this section.

\subsection{Estimation of the Euclidean distance}
\label{sec:consistent_estimator:eu}
The Euclidean distance takes the form in (\ref{eq:family_distances}) with
\[
\sum_{l=1}^{L}f_{1}^{(l)}(\omega_{1})f_{2}^{(l)}(\omega_{2})=w_1^2 + w_2^2 - 2w_1w_2.
\]
In order to solve the integral in (\ref{eq:dhatintegral}) we start by observing that the function 
\[
\hat{\mathbf{Q}}_{j}(z)z\hat{\omega}_{j}^{\prime}\left(  z\right)
=z\hat{\mathbf{Q}}_{j}(z)\frac{1-\frac{M}{N_{j}}+z^{2}\frac{1}{N_{j}
}\mathrm{tr}\left[  \hat{\mathbf{Q}}_{j}^{2}(z)\right]  }{\left(  1-\frac
{1}{N_{j}}\mathrm{tr}\left[  \hat{\mathbf{R}}_{j}\hat{\mathbf{Q}}
_{j}(z)\right]  \right)  ^{2}}
\]
presents all its poles inside the contour $\mathrm{C}_j^{(l)}$. Indeed, by definition, $\mathrm{C}_j^{(l)}$ encloses all the eigenvalues of $\hat{\mathbf{R}}_{j}$ almost surely for all large $M$ (see \cite{Bai98}), as well as the solutions to the equation $N_{j}^{-1}\mathrm{tr}[ \hat{\mathbf{R}}_{j}\hat{\mathbf{Q}}_{j}(z)] = 1$ (see \cite{Mestre08TIT}).
Furthermore, in the undersampled case, the function presents a removable singularity at zero, so that effectively all the singularities are located inside the contour $\mathrm{C}_j^{(l)}$. We can therefore enlarge the contour $\mathrm{C}$ without changing the value of the integral, and then apply the change of variables $z\mapsto\zeta=z^{-1}$, which only has a single singularity at $\zeta=0$. This can be done to show that, when $f_{j}^{(l)}(\omega)=\omega$, we have  
\begin{multline*}
\frac{1}{2\pi\mathrm{j}}\oint\nolimits_{\mathrm{C}_j^{(l)}}f_{j}^{(l)}(\hat{\omega}_{j}\left(  z\right)  )\frac{z\hat{\omega}_{j}^{\prime}\left(z\right)  }{\hat{\omega}_{j}\left(  z\right)  }\hat{\mathbf{Q}}_{j}
(z)dz =  
\\= 
\frac{1}{2\pi\mathrm{j}}\oint\nolimits_{C_{0}}\hat{\omega}_j^\prime\left(\zeta^{-1}\right)\hat{\mathbf{Q}}_{j}\left(\zeta^{-1}\right)\frac{1}{\zeta^{3}}d\zeta
\end{multline*}
where $C_{0}$ denotes a negatively oriented contour that encloses only $\zeta=0$ and no other singularity. Noting
that $\lim_{\zeta\rightarrow0}\zeta^{-1}\hat{\mathbf{Q}}_{j}(\zeta
^{-1})=-\mathbf{I}_{M}$ we see that the above integrand presents a second
order pole at $\zeta=0$ and therefore one easily computes (for the case
$f_{j}^{(l)}(\omega)=\omega$)
\[
\frac{1}{2\pi\mathrm{j}}\oint\nolimits_{\mathrm{C}^{-}}f_{j}^{(l)}(\hat
{\omega}_{j}\left(  z\right)  )\frac{z\hat{\omega}_{j}^{\prime}\left(
z\right)  }{\hat{\omega}_{j}\left(  z\right)  }\hat{\mathbf{Q}}_{j}
(z)dz=\hat{\mathbf{R}}_{j}.
\]
Proceeding in exactly the same way for $f_{j}^{(l)}(\omega)=\omega^{2}$ we see
that the pole at $\zeta=0$ has now order three, and therefore (after some
algebra)
\begin{multline*}
\frac{1}{2\pi\mathrm{j}}\oint\nolimits_{\mathrm{C}_j^{(l)}}f_{j}^{(l)}
(\hat{\omega}_{j}\left(  z\right)  )\frac{z\hat{\omega}_{j}^{\prime}\left(
z\right)  }{\hat{\omega}_{j}\left(  z\right)  }\hat{\mathbf{Q}}_{j}(z)dz 
= \\
= \frac{1}{2\pi\mathrm{j}}\oint\nolimits_{C_{0}}
\frac{\hat{\omega}_j^\prime\left(\zeta^{-1}\right)\hat{\mathbf{Q}}_{j}\left(\zeta^{-1}\right)}{\hat{\omega}_j\left(\zeta^{-1}\right)}
\frac{1}{\zeta^{3}}d\zeta
\\
=\hat{\mathbf{R}}_{j}^{2}-\left(  \frac{1}{N_{j}}\mathrm{tr}\left[
\hat{\mathbf{R}}_{j}\right]  \right)  \hat{\mathbf{R}}_{j}.
\end{multline*}
As a consequence of the above two integrals, we can conclude that the
estimator in (\ref{eq:dhatintegral}) particularizes to the Euclidean distance
as
\begin{multline*}
\hat{d}_{M}^{E}(1,2)=\frac{1}{M}\mathrm{tr}\left[  \left(  \hat{\mathbf{R}}
_{1}-\hat{\mathbf{R}}_{2}\right)  ^{2}\right]  -\frac{1}{MN_{1}}
\mathrm{tr}^{2}\left[  \hat{\mathbf{R}}_{j}\right]  
\\
-\frac{1}{MN_{2}
}\mathrm{tr}^{2}\left[  \hat{\mathbf{R}}_{2}\right]
\end{multline*}
which corresponds to the conventional estimator corrected by the square of the normalized trace of the two sample covariance matrices. Obviously, the estimator becomes the conventional one if both $N_{1},N_{2}$ increase but $M$ remains fixed. 
The expression of $\hat{d}_{M}^{E}(1,2)$ could also have been obtained by identifying the asymptotic bias term of the \textit{plug-in} estimator in \cite{pereira23tsp} and compensating it by using the fact that $M^{-1}\mathrm{tr}[\hat{\mathbf{R}}_j] \asymp M^{-1}\mathrm{tr}[{\mathbf{R}}_j]$.

\subsection{Estimation of the symmetrized KL divergence}
\label{sec:consistent_estimator:kl}

The symmetrized KL divergence corresponds to the definition in
(\ref{eq:family_distances}) with
\[
\sum_{l=1}^{L}f_{1}^{(l)}(\omega_{1})f_{2}^{(l)}(\omega_{2})=\frac{1}{2}
\frac{\omega_{2}}{\omega_{1}}+\frac{1}{2}\frac{\omega_{1}}{\omega_{2}}-1.
\]
We start by noticing that in this case the function $\omega^{-1}$ is not holomorphic at the origin, which implies that we can only tolerate $\mu_{\inf}^{(j)}=0$ in \textbf{(As4)}. In particular, this implies that we can only obtain a consistent estimator for the oversampled case (namely $N_{1}>M$ and $N_{2}>M$).

Here again, we need to solve the different integrals in (\ref{eq:dhatintegral}) for the functions $f_{1}^{(l)}(\omega)=\omega$ (already considered above), $f_{1}^{(l)}(\omega)=1$ and $f_{1}^{(l)}(\omega)=\omega^{-1}$. Let us first consider the simple case of $f_{1}^{(l)}(\omega)=1$. To solve the corresponding
integral, we can simply notice that
\[
\frac{z\hat{\omega}_{j}^{\prime}\left(  z\right)  }{\hat{\omega}_{j}\left(
z\right)  }\hat{\mathbf{Q}}_{j}(z)=\hat{\mathbf{Q}}_{j}(z)\frac{1-\frac
{M}{N_{j}}+z^{2}\frac{1}{N_{j}}\mathrm{tr}\left[  \hat{\mathbf{Q}}_{j}
^{2}(z)\right]  }{1-\frac{1}{N_{j}}\mathrm{tr}\left[  \hat{\mathbf{R}}
_{j}\hat{\mathbf{Q}}_{j}(z)\right]  }
\]
presents all its poles inside the contour $\mathrm{C}_j^{(l)}$ (note that there is no singularity at zero because $\hat{\mathbf{Q}}_{j}(0)=\hat{\mathbf{R}}_{j}^{-1}$ which is well-defined since we are considering the oversampled situation so that $\hat{\mathbf{R}}_{j}$ is invertible with probability one). Hence, we can enlarge the contour $\mathrm{C}_j^{(l)}$ and consider again the change of variable  $\zeta=z^{-1}$, after which the integrand will only have a singularity at $\zeta=0$. Consequently, we can find
\begin{multline*}
\frac{1}{2\pi\mathrm{j}}\oint\nolimits_{\mathrm{C}_j^{(l)}}\frac{z\hat{\omega}_{j}^{\prime}\left(  z\right)  }{\hat{\omega}_{j}\left(  z\right)}\hat{\mathbf{Q}}_{j}(z)dz 
=\\
= \frac{1}{2\pi\mathrm{j}}\oint\nolimits_{C_{0}}
\hat{\omega}_{j}\left(\zeta^{-1}\right)\hat{\omega}_{j}^\prime\left(\zeta^{-1}\right)\hat{\mathbf{Q}}_{j}\left(\zeta^{-1}\right)
\frac{1}{\zeta^{2}}d\zeta
=
\mathbf{I}_{M}.
\end{multline*}

Regarding the integral for $f_{1}^{(l)}(\omega)=\omega^{-1}$, we consider the function
\begin{equation}
\frac{\hat{\mathbf{Q}}_{j}(z)z\hat{\omega}_{j}^{\prime}\left(  z\right)}{\hat{\omega}_{j}^{2}\left(  z\right)  }
=
\frac{\hat{\mathbf{Q}}_{j}(z)}{z}\left(  1-\frac{M}{N_{j}}+z^{2}\frac{1}{N_{j}}\mathrm{tr}\left[\hat{\mathbf{Q}}_{j}^{2}(z)\right]  \right) 
\label{eq:KLintegrand}
\end{equation}
and observe again that all its singularities are inside the contour $\mathrm{C}_j^{(l)}$ except for a simple pole at $z=0\,$. Therefore, we can deform the contour $\mathrm{C}_j^{(l)}$ into a larger one $\mathcal{C}_j^{(l)}$ that now encloses $\mathcal{S}_{j}\cup\{0\}$ and write (for the case $f_{j}^{(l)}(\omega)=\omega^{-1}$)
\begin{multline*}
\frac{1}{2\pi\mathrm{j}}\oint\nolimits_{\mathrm{C}_j^{(l)}}f_{j}^{(l)}(\hat{\omega}_{j}\left(  z\right)  )\frac{z\hat{\omega}_{j}^{\prime}\left(z\right)  }{\hat{\omega}_{j}\left(  z\right)  }\hat{\mathbf{Q}}_{j}(z)dz =
\\
=
\left(  1-\frac
{M}{N_{j}}\right)  \hat{\mathbf{R}}_{j}^{-1}
+\frac{1}{2\pi\mathrm{j}}\oint\nolimits_{\mathcal{C}_j^{(l)}}
\frac{\hat{\mathbf{Q}}_{j}(z)z\hat{\omega}_{j}^{\prime}\left(  z\right)}{\hat{\omega}_{j}^{2}\left(  z\right)  }dz
\end{multline*}
where the first term is the residue of (\ref{eq:KLintegrand})\ at zero. We can now see that the integral on the right hand side of the above equation  is zero by enlarging the contour and
applying the change of variable $\zeta=z^{-1}$, after which the corresponding
integrand becomes analytic at zero. We can therefore conclude that
\[
\frac{1}{2\pi\mathrm{j}}\oint\nolimits_{\mathrm{C}_j^{(l)}}f_{j}^{(l)}(\hat
{\omega}_{j}\left(  z\right)  )\frac{z\hat{\omega}_{j}^{\prime}\left(
z\right)  }{\hat{\omega}_{j}\left(  z\right)  }\hat{\mathbf{Q}}_{j}
(z)dz=\left(  1-\frac{M}{N_{j}}\right)  \hat{\mathbf{R}}_{j}^{-1}.
\]
With this, we have now all the ingredients to evaluate the integral at (\ref{eq:dhatintegral}), which provides a consistent estimator for the symmetrized KL distance between covariance matrices, namely
\begin{multline*}
\hat{d}_{M}^{KL}(1,2)=\left(  1-\frac{M}{N_{1}}\right)  \frac{1}{2M}\mathrm{tr}\left[  \hat{\mathbf{R}}_{1}^{-1}\hat{\mathbf{R}}_{2}\right] 
+ \\ 
+\left(1-\frac{M}{N_{2}}\right)  \frac{1}{2M}\mathrm{tr}\left[  \hat{\mathbf{R}}_{2}^{-1}\hat{\mathbf{R}}_{1}\right]  -1.
\end{multline*}
The above expression of $\hat{d}_{M}^{KL}(1,2)$ could also have been obtained by identifying the asymptotic bias term of the \textit{plug-in} estimator in \cite{pereira23tsp} and compensating it by using the fact that $N_j^{-1}\mathrm{tr}[\hat{\mathbf{R}}_j^{-1}] \asymp (N_j-M)^{-1}\mathrm{tr}[{\mathbf{R}}_j^{-1}]$.
We emphasize, however, that directly attempting to correct the asymptotic bias term only works in very specific cases. In general terms (such as the log-Euclidean metric) this approach is not viable and one must directly try to solve the integrals in (\ref{eq:dhatintegral}).

\subsection{Estimation of the Log-Euclidean distance}

The log-Euclidean distance takes the form in (\ref{eq:family_distances}) with
\[
\sum_{l=1}^{L}f_{1}^{(l)}(\omega_{1})f_{2}^{(l)}(\omega_{2})=\left(
\log\omega_{1}-\log\omega_{2}\right)  ^{2}
\]
and therefore to evaluate the integral in (\ref{eq:dhatintegral}) one must
consider the two functions $f_{j}^{(l)}(\omega)=$ $\log\omega$ and
$f_{j}^{(l)}(\omega)=$ $\left(  \log\omega\right)  ^{2}$. Observe now that
these two functions are analytic everywhere except for the negative real axis
(including zero). Hence, in \textbf{(As4)} we must have $\mu_{\inf}^{(j)}=0$,
implying that $\mu_{0}^{(j)}=0$ for all $M$ and hence $N_{j}>M$ (oversampled regime).

The first integral (with $f_{j}^{(l)}(\omega)=$ $\log\omega$) was
already solved in \cite{mestre12}. In order to present the closed form solution
for this integral, let us denote by $\hat{\lambda}_{1}^{(j)}\leq\ldots
\leq\hat{\lambda}_{M}^{(j)}$ and $\hat{\mathbf{e}}_{1}^{(j)},\ldots
,\hat{\mathbf{e}}_{M}^{(j)}$ the eigenvalues and associated eigenvectors of
the sample covariance matrix $\hat{\mathbf{R}}_{j}$. Using the results in \cite{mestre12}, we can clearly see that
\[
\frac{1}{2\pi\mathrm{j}}\oint\nolimits_{\mathrm{C}_j^{(l)}}\log(\hat{\omega}
_{j}\left(  z\right)  )\frac{z\hat{\omega}_{j}^{\prime}\left(  z\right)
}{\hat{\omega}_{j}\left(  z\right)  }\hat{\mathbf{Q}}_{j}(z)dz=\sum_{k=1}
^{M}\beta_{k}^{(j)}\hat{\mathbf{e}}_{k}^{(j)}\left(  \hat{\mathbf{e}}
_{k}^{(j)}\right)  ^{H}
\]
where the coefficients $\beta_{k}^{(j)}$, $k=1,\ldots,M$, are defined as
\begin{multline*}
\beta_{k}^{(j)} =\left(  1+\sum_{\substack{m=1\\m\neq k}}^{M}\frac{\hat{\lambda}_{k}^{(j)}}{\hat{\lambda}_{m}^{(j)}-\hat{\lambda}_{k}^{(j)}}-\sum_{m=1}^{M}\frac{\hat{\mu}_{k}^{(j)}}{\hat{\lambda}_{m}^{(j)}-\hat{\mu}_{k}^{(j)}}\right)  \log\hat{\lambda}_{k}^{(j)}
\\
+\left(  \sum_{\substack{r=1\\r\neq k}}^{M}\frac{\hat{\lambda}_{r}^{(j)}}{\hat{\lambda}_{r}^{(j)}-\hat{\lambda}_{k}^{(j)}}\log\hat{\lambda}_{r}^{(j)}-\sum_{r=1}^{M}\frac{\hat{\mu}_{r}^{(j)}}{\hat{\mu}_{r}^{(j)}-\hat{\lambda}_{k}^{(j)}}\log\hat{\mu}_{r}^{(j)}\right)
+1
\end{multline*}
and where we have denoted by $\hat{\mu}_{1}^{(j)}<\ldots<\hat{\mu}_{M}^{(j)}$ the $M$
solutions to the polynomial equation 
 \[
 1=\frac{1}{N_{j}}\sum_{k=1}^{M}\frac{\hat{\lambda}_{k}^{(j)}}{\hat{\lambda
 }_{k}^{(j)}-\hat{\mu}}.
 \]

Moving forward to the case $f_j^{(l)}(\omega) = \log^2(\omega)$, let $\alpha^{(j)}$ denote the value of the corresponding integral, namely 
\begin{equation}
\alpha^{(j)}=\frac{1}{2\pi\mathrm{j}}\oint\nolimits_{\mathrm{C}_j^{(l)}}\log
^{2}(\hat{\omega}_{j}\left(  z\right)  )\frac{z\hat{\omega}_{j}^{\prime
}\left(  z\right)  }{\hat{\omega}_{j}\left(  z\right)  }\frac{1}{M}
\mathrm{tr}\left[  \hat{\mathbf{Q}}_{j}(z)\right]  dz \label{eq:integrallog2}
\end{equation}
Using very similar arguments as above, it is shown in Appendix
\ref{sec:AppendixIntegralLog} that  $\alpha^{(j)}$ takes the form
\begin{multline}
\alpha^{(j)}=
\\
\left(  \frac{N_{j}}{M}-1\right)  \sum_{r=1}^{M}
\left(
\left(1+\log\hat{\mu}_{r}^{(j)}\right)  ^{2}-\left(  1+\log\hat{\lambda}_{r}^{(j)}\right)^{2}
\right)
\\
+
\frac{1}{M}\sum_{k=1}^{M}\left(  1+\log\hat{\lambda}_{k}^{(j)}\right)^{2}{-}\left(  \frac{N_{j}}{M}-1\right)  \log^{2}\left(  1-\frac{M}{N_{j}}\right)  + 1
\\
+\frac{2}{M}\sum_{k=1}^{M}\sum_{r=1}^{M}\left[  \Phi_{2}\left(  \frac{\hat{\mu}_{r}^{(j)}}{\hat{\lambda}_{k}^{(j)}}\right)  -\Phi_{2}\left(\frac{\hat{\lambda}_{r}^{(j)}}{\hat{\lambda}_{k}^{(j)}}\right)  \right] 
 \\
+\frac{2}{M}\sum_{k=1}^{M}\Biggl(  \sum_{\substack{r=1\\r\neq k}}^{M}\log\frac{\hat{\lambda}_{r}^{(j)}}{\hat{\lambda}_{k}^{(j)}}\log\frac{\hat{\lambda}_{k}^{(j)}}{\left\vert \hat{\lambda}_{k}^{(j)}-\hat{\lambda}_{r}^{(j)}\right\vert } 
\\ 
-\sum_{r=1}^{M}\log\frac{\hat{\mu}_{r}^{(j)}}{\hat{\lambda}_{k}^{(j)}}\log\frac{\hat{\lambda}_{k}^{(j)}}{\left\vert \hat{\lambda}_{k}^{(j)}-\hat{\mu}_{r}^{(j)}\right\vert }\Biggl) 
\label{eq:definition_alpha_j}
\end{multline}
where we have defined
\begin{equation}
\Phi_{2}(x) 
=\left\{
\begin{array}
[c]{ccc}
\mathrm{Li}_{2}\left(  x\right)  &  & x<1\\
\frac{\pi^{2}}{3}-\frac{1}{2}\log^{2}x-\mathrm{Li}_{2}\left(  x^{-1}\right)  &
& x\geq1
\end{array}
\right.  \label{eq:definitionPhi(x)Li2}
\end{equation}
and where $\mathrm{Li}_{2}\left(  x\right)  =-\int_{0}^{x} y^{-1}\log(1-y) dy$ is the di-logarithm function.

Combining the two expressions above, we obtain a closed form expression for the consistent estimator of the log-Euclidean distance between covariance matrices, namely
\[
\hat{d}_{M}^{LE}(1,2)=\alpha^{(1)}+\alpha^{(2)}-\frac{2}{M}\sum_{k=1}^{M}\sum_{m=1}^{M}\beta_{k}^{(1)}\beta_{m}^{(2)}\left\vert \left(  \hat{\mathbf{e}}_{k}^{(1)}\right)  ^{H}\hat{\mathbf{e}}_{m}^{(2)}\right\vert ^{2}
\]
where $\alpha^{(j)}$ and $\beta_{k}^{(j)}$, $k=1,\ldots,M$ are defined as above.

It is interesting to compare the above expression with the consistent estimator of the affine-invariant metric defined in \cite[eq. B7]{couillet2019random}. Both estimators are defined through the evaluation of the dilogarithm function at certain quotients between eigenvalues $\hat{\lambda}^{(j)}_k$ and roots $\hat{\mu}^{(j)}_k$ and are therefore comparable in terms of computational complexity. It would be interesting to study which one of the two estimators offers a superior performance in practical applications. Since both estimators are consistent, a rigorous analysis would require an asymptotic characterization of the fluctuations of the affine invariant metric estimator, which has not been addressed so far. 

\section{A central limit theorem on the proposed estimators}
\label{sec:clt} 


The objective of this section is to characterize the asymptotic fluctuations of the distance estimators $\hat{d}_{M}$ around the true values $d_{M}$. The goal is to provide an analytical framework that allows us to compare the performance of the different estimated distances in practical applications, such as clustering observations according to their covariance. To that effect, instead of considering a single distance between two subsets of observations, we will consider here a collection of estimators among multiple observation sets. 

Indeed, consider here a collection of sets of observations indexed by $\mathcal{J}$, that is $\mathbf{Y}_j \in \mathbb{K}^{M \times N_j}$ where $j \in \mathcal{J}$. Let $\hat{\mathbf{d}}_M$ denote an $R$-dimensional column vector containing the estimated covariance distances between $R$ different pairs of observations, that is 
\begin{equation} \label{eq:defDistvect}
\hat{\mathbf{d}}_M = \left[ \hat{d}_M(i_1,j_1), \ldots \hat{d}_M(i_R,j_R) \right]^T
\end{equation}
where $i_r, j_r \in \mathcal{J}$, $i_r \neq j_r$, $r=1,\ldots,R$. Likewise, we will denote by $\mathbf{d}_{M}$ the $R$-dimensional column vector containing the actual distances between these covariances. 
We will basically show that the random vector $M(  \hat{\mathbf{d}}_{M}-\mathbf{d}_{M})  $ asymptotically fluctuates according to a multivariate Gaussian distribution with
certain mean and covariance that will be characterized below. 
In order to present the main result, we need to introduce some quantities that will be used in the formulation of the main central limit theorem. 


We define the asymptotic (second order) mean of this random vector as an $R$-dimensional column vector $\bar{\boldsymbol{\mathfrak{m}}}_M$, which has its $r$th entry equal to 
\begin{align}
&\left[\bar{\boldsymbol{\mathfrak{m}}}_{M}\right]_r   =
\nonumber \\ &=\sum_{l=1}^{L}\frac{\varsigma}{2\pi\mathrm{j}}\oint\nolimits_{\mathrm{C}_{\omega_{i_r}}^{(l)}}f_{i_r}^{(l)}(\omega_{i_r})\frac{\mathrm{tr}\left[  \mathbf{\mathbf{R}}_{i_r}^{2}\mathbf{Q}_{i_r}^{3}\left(  \omega_{i_r}\right)  f_{j_r}^{(l)}\left(  \mathbf{R}_{j_r}\right)\right]  }{N_{i_r}(1-\Gamma_{i_r}(\omega_{i_r}))}d\omega_{i_r} 
\nonumber\\
& +\sum_{l=1}^{L}\frac{\varsigma}{2\pi\mathrm{j}}\oint\nolimits_{\mathrm{C}_{\omega_{j_r}}^{(l)}}f_{j_r}^{(l)}(\omega_{j_r}) \frac{\mathrm{tr}\left[  \mathbf{\mathbf{R}}_{j_r}^{2}\mathbf{Q}_{j_r}^{3}\left(  \omega_{j_r}\right)  f_{i_r}^{(l)}\left(  \mathbf{R}_{i_r}\right)  \right]  }{N_{j_r} (1-\Gamma_{j_r}(\omega_{j_r}))}d\omega_{j_r} \label{eq:2ndOrderMeanSimplified}
\end{align}
where we have introduced the function (for $j \in \mathcal{J}$)
\begin{equation}
\Gamma_{j}\left(  \omega\right)  =\frac{1}{N_{j}}\mathrm{tr}\left[
\mathbf{R}_{j}^{2}\mathbf{Q}_{j}^{2}(\omega)\right] . \label{eq:defGamma}
\end{equation}

Likewise, we will define the asymptotic covariance matrix $\bar{\boldsymbol{\Sigma}}_M$ as an $R \times R$ matrix with $(r,s)$th entry equal to\footnote{We are using again the shorthand notation $\omega_{j}=\omega_{j}(z_{j})$ and $\tilde{\omega}_{j}=\omega_{j}({\tilde{z}}_{j})$.} 
\begin{multline}
    \left[\bar{\boldsymbol{\Sigma}}_M\right]_{r,s} 
    = \\
    = \sum_{l_r,l_s=1}^{L}\frac{1+\varsigma}{(2\pi\mathrm{j})^4} \oint\nolimits_{\mathrm{C}^{(l_r)}_{\omega_{i_r}}}\oint\nolimits_{\mathrm{C}^{(l_r)}_{\omega_{j_r}}}\oint\nolimits_{\mathrm{C}^{(l_s)}_{\omega_{i_s}}}\oint\nolimits_{\mathrm{C}^{(l_s)}_{\omega_{j_s}}} 
     f_{i_r}^{(l_r)}(\omega_{i_r})
     \times \\ \times
     f_{j_r}^{(l_r)}(\omega_{j_r}) f_{i_s}^{(l_s)}(\tilde{\omega}_{i_s}) f_{j_s}^{(l_s)}(\tilde{\omega}_{j_s}) 
     \times \\\times 
    \bar{\sigma}^2_{i_r,j_r,i_s,j_s}(\omega_{i_r},\omega_{j_r},\tilde{\omega}_{i_s},\tilde {\omega}_{j_s}) d\omega_{i_r} d\omega_{j_r} d\tilde{\omega}_{i_s} d\tilde{\omega}_{j_s}
    \label{eq:asympvar}
\end{multline}
where $\bar{\sigma}^2_{i,j,m,n}(\omega_{i},\omega_{j},\tilde{\omega}_{m},\tilde{\omega}_{n}) $ is defined as
\begin{align}
\bar{\sigma}^{2}_{i,j,m,n}&(  \omega_{i},\omega_{j},\tilde{\omega}_{m},\tilde{\omega}_{n}) =
\nonumber\\
&= 
\bar{\sigma}_{i}^{2}\left(  \omega_{i},\tilde{\omega}_{m};\mathbf{Q}_{j}\left(
\omega_{j}\right)  ,\mathbf{Q}_{n}(\tilde{\omega}_{n})\right) \delta_{i=m}  \nonumber \\&
\hspace{1em} + \bar{\sigma}_{j}^{2}\left(  \omega_{j},\tilde{\omega}_{n};\mathbf{Q}_{i}\left(
\omega_{i}\right)  ,\mathbf{Q}_{m}(\tilde{\omega}_{m})\right) \delta_{j=n} 
\nonumber \\& 
\hspace{1em}+ \bar{\sigma}_{i}^{2}\left(  \omega_{i},\tilde{\omega}_{n};\mathbf{Q}_{j}\left(
\omega_{j}\right)  ,\mathbf{Q}_{m}(\tilde{\omega}_{m})\right) \delta_{i=n} 
\nonumber \\&
\hspace{1em}+ \bar{\sigma}_{j}^{2}\left(  \omega_{j},\tilde{\omega}_{m};\mathbf{Q}_{i}\left(
\omega_{i}\right)  ,\mathbf{Q}_{n}(\tilde{\omega}_{n})\right) \delta_{j=m} 
\nonumber \\& 
\hspace{1em} + {\varrho}_{i,j}(\omega_i,\omega_j,\tilde{\omega}_m,\tilde{\omega}_n) \delta_{i=m}\delta_{j=n} 
\nonumber \\&
\hspace{1em} + \varrho_{i,j}(\omega_i,\omega_j,\tilde{\omega}_n,\tilde{\omega}_m) \delta_{i=n}\delta_{j=m} 
\end{align}
with the following additional definitions. The functions $\bar{\sigma}_{j}^{2}\left(  \omega,\tilde{\omega};\mathbf{A},\mathbf{B}\right) $ for $j \in \mathcal{J} $ are defined as
 \begin{multline}
\bar{\sigma}_{j}^{2}\left(  \omega,\tilde{\omega};\mathbf{A},\mathbf{B}\right)  
=\frac{1}{1-\Gamma_{j}(\omega,\tilde{\omega})}\frac{1}{N_{j}}\mathrm{tr}
\Biggl[  \mathbf{R}_{j}\mathbf{Q}_{j}\left(  \omega\right)  \mathbf{Q}
_{j}\left( \tilde{ \omega}\right) 
\times  \\  \times
  \mathbf{A}\mathbf{R}_{j}\mathbf{Q}_{j}\left(  \omega\right)  \mathbf{Q}
_{j}\left( \tilde{\omega} \right) \mathbf{B}  \Biggl] +  \\
+\frac{1}{\left(  1-\Gamma_{j}(\omega,\tilde{\omega})\right)  ^{2}}
\frac{1}{N_{j}}\mathrm{tr}\left[  \mathbf{R}_{j}^{2}\mathbf{Q}_{j}^{2}\left(
\omega\right)  \mathbf{Q}_{j}\left( \tilde{\omega} \right)  \mathbf{A} \right]  
\times  \\  \times
\frac{1}{N_{j}}\mathrm{tr}\left[
\mathbf{R}_{j}^{2}\mathbf{Q}_{j}\left(  \omega\right)  \mathbf{Q}_{j}
^{2}\left( \tilde{\omega} \right) \mathbf{B} \right] 
\end{multline} 
 where we have introduced the quantities (with some abuse of notation)
\begin{equation}\label{eq:defGammabivariate}
\Gamma_{j}(\omega,\tilde{\omega})=\frac{1}{N_{j}}\mathrm{tr}\left[
\mathbf{R}_{j}^{2} \mathbf{Q}_j\left(\omega\right) \mathbf{Q}_j\left(\tilde{\omega}\right)
\right].
\end{equation}
Regarding the functions ${\varrho}_{i,j}(\omega_i,\omega_j,\tilde{\omega}_i,\tilde{\omega}_j) $, they are defined as 
\begin{multline}
{\varrho}_{i,j}(\omega_i,\omega_j,\tilde{\omega}_i,\tilde{\omega}_j)
=
\\
=
\frac{\mathrm{tr}^{2}\left[  \mathbf{R}_{i}\mathbf{Q}_{i}\left(  \omega
_{i}\right)  \mathbf{Q}_{i}\left( \tilde{\omega}_{i}\right)  \mathbf{R}
_{j}\mathbf{Q}_{j}\left(  \omega_{j}\right)  \mathbf{Q}_{j}\left(  \tilde{\omega}_{j}\right)  \right]  }{N_{i}N_{j}\left(  1-\Gamma_{i}(\omega
_{i},\tilde{\omega}_{i})\right)  \left(  1-\Gamma_{j}(\omega_{j},\tilde{\omega}
_{j})\right)  }.  
\end{multline}

We have now all the ingredients to present the following result, which is
proven in Appendix \ref{sec:proofclt}.
\begin{theorem} 
\label{th:cltEstimators}
In addition to \textbf{(As1)}-\textbf{(As4)}, assume that the observations are
Gaussian distributed and that the minimum eigenvalue of $\bar{\boldsymbol{\Sigma}}_M$ is bounded away from zero. Then, the random vector 
\[
\bar{\boldsymbol{\Sigma}}_M^{-1/2}\left[M(\hat{\mathbf{d}}_M - \mathbf{d}_M)-\bar{\boldsymbol{\mathfrak{m}}}_{M}\right] 
\]
converges in law to a multivariate standard Gaussian distribution. 
\end{theorem}
\begin{IEEEproof}
    See Appendix \ref{sec:proofclt}. 
\end{IEEEproof}
Even if the expression obtained above appears to be difficult to evaluate due to the presence of the contour integrals, one can  typically simplify these expressions using conventional residue calculus. This is illustrated next for the three distances that have been considered above. In particular, the fourfold integral in (\ref{eq:asympvar}) can easily be simplified so that an alternative expression can be given with a single integral only, which can easily be evaluated using numerical integration. More details are given in Appendix \ref{sec:simplifiedIntegralsVar}. 

In any case, and in order to simplify the exposition, in the remainder of this section we assume $R=1$ and $\mathcal{J} = \{1,2\}$, the results being trivially extrapolated to the more general case of $R>1$ and multiple sample covariance matrix.
\begin{remark}
    The proof of Theorem \ref{th:cltEstimators} builds upon a very similar result derived in \cite[Theorem 2]{pereira23tsp}, which is a Central Limit Theorem for the \textit{plug-in} estimator in (\ref{eq:definitionhatd}). Hence, most of the work behind the proof in Appendix~\ref{sec:simplifiedIntegralsVar} is to transform the original estimator into a more convenient form (up to certain errors that converge to zero fast enough) so that \cite[Theorem 2]{pereira23tsp} can be applied. In this sense, it is interesting to compare the result obtained here with the ones in \cite{pereira23tsp}. Surprisingly enough, the expression for the asymptotic mean and variances of the consistent estimators are very similar to the ones in \cite[equations (17) and (19)]{pereira23tsp} with the exception that the quantities $\Omega_j(\omega_j;\mathbf{A})$ in \cite{pereira23tsp} (see (\ref{eq:DefOmega}) in Appendix~\ref{sec:proofclt}) are replaced by the matrices $\mathbf{R}_j$ in the corresponding quantities here. Hence, it turns out that the expression for the asymptotic (second order) mean and variance of the consistent estimators are somewhat simpler than the ones for the \textit{plug-in} estimators.
\end{remark}

\subsection{Particularization to the Euclidean distance}
For the conventional Euclidean norm we have $f(\omega_{1},\omega_{2}
)=(\omega_{1}-\omega_{2})^{2}$ and both the integrands in (\ref{eq:2ndOrderMeanSimplified}) and (\ref{eq:asympvar}) have all the singularities inside the corresponding contours. The strategy to solve these integrals is
therefore to apply the change of variable $\omega_j \mapsto \zeta_j = \omega_j^{-1}$ after enlarging the contours, so that the resulting integrands after the change of variable have only a singularity at $\zeta_j = 0$. Using this technique, one can show that the asymptotic (second order) mean takes the form
\begin{equation*}
\bar{\mathfrak{m}}_{M}^{E}   =\varsigma\left(  \frac{1}{N_{1}}\mathrm{tr}\left[  \mathbf{\mathbf{R}}
_{1}^{2}\right]  +\frac{1}{N_{2}}\mathrm{tr}\left[  \mathbf{R}_{2}^{2}\right]
\right)  .
\end{equation*}
Regarding the asymptotic variance, one can use exactly the same 
integration technique to show that 
\begin{align*}
\frac{\bar{\boldsymbol{\Sigma}}_{M}}{1+\varsigma}    &=2  \left(  \frac{1}{N_{1}
}\mathrm{tr}\left[  \mathbf{R}_{1}^{2}\right]  \right)  ^{2}
+ 4 \frac{1}{N_{1}}\mathrm{tr}\left[  \mathbf{R}_{1} \boldsymbol{\Delta}  \mathbf{R}_{1} \boldsymbol{\Delta} \right] 
\\&  
+2 \left(  \frac{1}{N_{2}}\mathrm{tr}\left[
\mathbf{R}_{2}^{2}\right]  \right)  ^{2} + 4 \frac
{1}{N_{2}}\mathrm{tr}\left[  \mathbf{R}_{2}\boldsymbol{\Delta}  \mathbf{R}_{2} \boldsymbol{\Delta} \right] 
\\& 
+4  \frac{1}{N_{1}N_{2}}\mathrm{tr}^{2}\left[
\mathbf{R}_{1}\mathbf{\mathbf{R}}_{2}\right] 
\end{align*}
where $\boldsymbol{\Delta} = \mathbf{R}_1 -\mathbf{R}_2$. 
Obviously, all the terms are non-negative, so that in order to show that $\liminf_{M\rightarrow\infty} \bar{\boldsymbol{\Sigma}}_M>0$ it is sufficient to see that at least one of them is bounded away from zero. In particular, using the fact that the eigenvalues of $\mathbf{R}_1$ are located inside a compact interval of $\mathbb{R}^+$ independent of $M$, one trivially sees that the first term is bounded away from zero.  The terms involving $\mathbf{\Delta}$ are positive whenever $\mathbf{R}_1 \neq \mathbf{R}_2$,  otherwise the terms are zero, and the above simplifies to 
\begin{align*}
\frac{\bar{\boldsymbol{\Sigma}}_{M}}{1+\varsigma}  &= 2  \left(  \frac{1}{N_{1}}\mathrm{tr}\left[  \mathbf{R}_{1}^{2}\right]  \right)  ^{2}
  +2 \left(  \frac{1}{N_{2}}\mathrm{tr}\left[
\mathbf{R}_{2}^{2}\right]  \right)  ^{2} 
\nonumber\\
 &\hspace{7em}+4  \frac{1}{N_{1}N_{2}}\mathrm{tr}^{2}\left[
\mathbf{R}_{1}\mathbf{\mathbf{R}}_{2}\right] =
\nonumber\\
&=2\left(\frac{1}{N_1} + \frac{1}{N_2}\right)^2\mathrm{tr}^2[\mathbf{R}^2].
\end{align*}

\subsection{Particularization to the symmetrized KL divergence}
For the symmetrized KL divergence we need to particularize the above expressions ($\bar{\frak{m}}_M$ and $\bar{\mathbf{\Sigma}}_M$) to the case 
$$
f(\omega_{1},\omega_{2})=\frac
{1}{2}\left(  \frac{\omega_{1}}{\omega_{2}}+\frac{\omega_{2}}{\omega_{1}
}\right)  -1.
$$ 
We recall that the estimator for this particular distance has only been defined in the oversampled case ($N_j > M$), and in this case the contour $\mathrm{C}_{\omega_j}$ does not enclose $\{0\}$ \cite{Mestre08tsp}. Hence, the integrands in (\ref{eq:2ndOrderMeanSimplified}) and (\ref{eq:asympvar}) have all the singularities inside the contour, except for a potential singularity at zero. The integration strategy therefore consists in enlarging the contour $\mathrm{C}_{\omega_j}$ so that it also encloses this singularity, compensating the result by adding the corresponding residue at zero (note that the original contour is always negatively oriented). The resulting integral can therefore be solved applying the change of variables $\omega_j \mapsto \zeta_j = \omega_j^{-1}$. 

Using this integration technique one can easily show that the asymptotic (second order) mean takes the form
\begin{equation*}
\bar{\mathfrak{m}}_{M}^{KL}  =\frac{\varsigma}{2}\left(  \frac{1}{N_{1}-M}\mathrm{tr}\left[
\mathbf{R}_{1}^{-1}\mathbf{R}_{2}\right]  +\frac{1}{N_{2}-M}\mathrm{tr}\left[
\mathbf{R}_{1}\mathbf{R}_{2}^{-1}\right]  \right)  
\end{equation*}
whereas the asymptotic variance can be expressed as
\begin{multline*}
\frac{\bar{\boldsymbol{\Sigma}}_M}{(1+\varsigma) (N_{1}+N_{2}-M)}  =
\\
= -\frac{M}{2 N_{1}N_{2}}    + \frac{\mathrm{tr}\left[  \mathbf{R}_{1}\mathbf{R}_{2}
^{-1}\mathbf{R}_{1}\mathbf{R}_{2}^{-1}\right]}{4\left(
N_{2}-M\right)  N_{1}}  
+\frac{\mathrm{tr}\left[
\mathbf{R}_{1}^{-1}\mathbf{R}_{2}\mathbf{R}_{1}^{-1}\mathbf{R}_{2}\right]}{4\left(  N_{1}-M\right)  N_{2}} 
\\
   +\frac{1}{4 N_{1}}\left(  \frac
 {\mathrm{tr}\left[  \mathbf{R}_{1}\mathbf{R}_{2}^{-1}\right]  }{N_{2}
 -M}\right)  ^{2}+\frac{1}{ 4 N_{2}}\left(
 \frac{\mathrm{tr}\left[  \mathbf{R}_{1}^{-1}\mathbf{R}_{2}\right]  }{N_{1}
 -M}\right)  ^{2}.
\end{multline*}
One can easily show that this is positive by applying the inequality
$\alpha\mathrm{tr}\left[  \mathbf{A}\right]  +\beta\mathrm{tr}\left[
\mathbf{A}^{-1}\right]  \geq2M\sqrt{\alpha\beta}$ (valid for any positive-definite $M\times M$ matrix $\mathbf{A}$). Indeed, using this inequality we see that the sum of the first three terms of the above expression is non-negative. The fact that the other two terms are bounded away from zero follows easily from the fact that the eigenvalues of the covariance matrices are located in a compact interval of $\mathbb{R}^+$ independent of $M$. This shows that $\liminf_{M\rightarrow\infty}\sigma_M^2 >0$ and the CLT holds. 
Finally, when $\mathbf{R}_{1}=\mathbf{R}_{2}$ this particularizes to
\begin{align*}
\mathfrak{m}_{M}^{KL}  =\frac{\varsigma}{2}\left(  \frac{M}{N_{1}-M} +\frac{M}{N_{2}-M}  \right) 
\end{align*}
and
\begin{align*}
\frac{\bar{\boldsymbol{\Sigma}}_{M}}{1+\varsigma}   &=
\frac{1}{2}\frac{M^{2}}{N_{1}N_{2}}-\frac{1}{2}
\frac{M}{N_{1}}-\frac{1}{2}\frac{M}{N_{2}} +
\\
&+\frac{1}{4}\frac{\left(  N_{1}+N_{2}-M\right)
M}{N_{1}N_{2}}\left(  \frac{N_{1}^{2}}{\left(  N_{1}-M\right)  ^{2}}
+\frac{N_{2}^{2}}{\left(  N_{2}-M\right)  ^{2}}\right) 
\end{align*}
which is particularly useful when one does not have access to the true covariance matrices. 

\subsection{Particularization to the log-Euclidean distance}

Finally, the log-Euclidean distance we have $f(\omega_1,\omega_2) = (\log\omega_1 -\log\omega_2)^2$ and the above integral tricks are not useful anymore due to the fact that the integrands are not holomorphic on the whole negative axis. 
We can still find the solution to the asymptotic (second-order) mean by evaluating the residues at the poles given by the eigenvalues of
$\mathbf{R}_{j}$ and the solutions to $\Gamma_{j}(\omega_{j})=1$, which are
denoted $\theta_{m}^{(j)}$, $m=1,\ldots,2\bar{M}_{j}$. One can readily see
that (assuming that the roots $\theta_{m}^{(j)}$ are of multiplicity one) the asymptotic (second order) mean takes the form
\begin{align*}
\bar{\mathfrak{m}}_{M}^{LE} & = 
-\varsigma\sum_{m=1}^{\bar{M}_{1}}\frac{1}{K_{m}^{(1)}}
\mathrm{tr}\left[ \boldsymbol{\Pi}^{(1)}_m
\left( \log\mathbf{R}_{2}  - \log\gamma_{m}^{(1)}\mathbf{I}_{M} \right)^2
\right] 
\\ &
-\varsigma\sum_{m=1}^{\bar{M}_{2}}\frac{1}{K_{m}^{(2)}}\mathrm{tr}\left[
\boldsymbol{\Pi}^{(2)}_m
\left(
\log\mathbf{R}_{1}-\log\gamma_{m}^{(2)}\mathbf{I}_{M}\right)  ^{2}\right] 
\\
&  +\frac{\varsigma}{2}\sum_{m=1}^{2\bar{M}_{1}}
\frac{
\mathrm{tr}\left[  \mathbf{\mathbf{R}}_{1}^{2}\mathbf{Q}_{1}^{3}(\theta_{m}^{(1)})  \left( \log\mathbf{R}_{2} - \log\theta_{m}^{(1)}\mathbf{I}_{M} \right)  ^{2}\right]  
}{
\mathrm{tr}\left[\mathbf{\mathbf{R}}_{1}^{2}\mathbf{Q}_{1}^{3}(  \theta_{m}^{(1)})\right] 
}
\\ &
  +\frac{\varsigma}{2}\sum_{m=1}^{2\bar{M}_{2}}\frac{\mathrm{tr}\left[  \mathbf{R}_{2}^{2}\mathbf{Q}_{2}^{3}(  \theta
_{m}^{(2)})  \left(  \log\mathbf{R}_{1}-\log\theta_{m}^{(2)}
\mathbf{I}_{M}\right)  ^{2}\right]  }{\mathrm{tr}\left[
\mathbf{\mathbf{R}}_{2}^{2}\mathbf{Q}_{2}^{3}(  \theta_{m}^{(2)})
\right]  }
\end{align*}
where $\boldsymbol{\Pi}^{(j)}_M$ is the projection matrix onto the subspace spanned by the eigenvector(s) of $\mathbf{R}_j$ associated to the eigenvalue $\gamma_m^{(j)}$, assumed to have multiplicity $K_m^{(j)}$. 
In the particular case where $\mathbf{R}_{1}=\mathbf{R}_{2}=\mathbf{R}$ the first two terms become zero (note that $\boldsymbol{\Pi}_m^{(j)}\log\mathbf{R}_j=\log(\gamma^{(j)}_m)\boldsymbol{\Pi}_m^{(j)} $) and the above simplifies to
\[
\bar{\mathfrak{m}}_{M}^{LE}=\varsigma\sum_{m=1}^{2\bar{M}}\frac{\mathrm{tr}\left[  \mathbf{\mathbf{R}}^{2}\mathbf{Q}^{3}\left(  \theta
_{m}\right)  \left(  \log\theta_{m}\mathbf{I}_{M}-\log\mathbf{R}\right)
^{2}\right]  }{\mathrm{tr}\left[  \mathbf{\mathbf{R}}^{2}
\mathbf{Q}^{3}\left(  \theta_{m}\right)  \right]  }.
\]
We have not been able to come up with a compact expression for the asymptotic variance. We therefore will only evaluate it using numerical integration using the approach in Appendix \ref{sec:simplifiedIntegralsVar}, which shows that the fourfold integral can in fact be reduced to a single one.  


\section{Numerical Evaluation}
In order to validate the results presented above, we consider $R$ multidimensional observation sets $\mathbf{Y}_r \in \mathbb{R}^{ M\times N_r}, r = 1, \ldots, R$, associated to $R$ (possibly distinct) Toeplitz covariance matrices $\mathbf{R}_r, r = 1, \ldots, R$, with first rows $[\rho_r^0, \dots, \rho_r^{M-1}]$. We will carry out the analysis in three steps.  
First, through numerical evaluation, we assess the consistency of the asymptotic descriptors ($\bar{\mathfrak{m}}_M$ and $\bar{\boldsymbol{\Sigma}}_M$) of the proposed consistent estimators ($\hat{d}_M$) as defined in Section~\ref{sec:clt}.
Second, we compare the traditional \textit{plug-in} distances to these consistent estimators, where the former (represented by $\tilde{d}_M$ and ``TRAD'' in the figures) are obtained by directly replacing the covariance matrices $\mathbf{R}_1$, $\mathbf{R}_2$ with their respective estimates, $\hat{\mathbf{R}}_1$, $\hat{\mathbf{R}}_2$ in  definition  (\ref{eq:family_distances}). 
Finally, we employ these derived solutions to improve the quality of the metrics applied to the problem of clustering random observations.

\begin{figure}[t]
\hspace{-2.5em}
 \begin{subfigure}[b]{0.6\textwidth}
      \centering
{\input{figs/final_double_column/histograms/top_hist_eu_M_40_N1_400_N2_80_08_04}
\input{figs/final_double_column/histograms/bot_hist_eu_M_80_N1_800_N2_160_08_04}}
       \caption{EU} 
 \end{subfigure}
 
 \hspace{-2.5em}{
 \begin{subfigure}[b]{0.6\textwidth}
      \centering
{\input{figs/final_double_column/histograms/top_hist_kl_M_40_N1_400_N2_80_08_04}
\input{figs/final_double_column/histograms/bot_hist_kl_M_80_N1_800_N2_160_08_04}}
       \caption{KL} 
 \end{subfigure}
 }
 
 \hspace{-2.5em}{
 \begin{subfigure}[b]{0.6\textwidth}
      \centering
{\input{figs/final_double_column/histograms/top_hist_le_M_40_N1_400_N2_80_08_04}
\input{figs/final_double_column/histograms/bot_hist_le_M_80_N1_800_N2_160_08_04}}
       \caption{LE} 
 \end{subfigure}
 }
     \hspace*{0.05in}
      \vspace{-1.\baselineskip}
      \caption{
      Histogram of empirical distribution (in blue) and asymptotic descriptors (in orange) of different metrics EU, KL and LE arranged from top to bottom, respectively,   for fixed $\rho_1 = 0.8, \rho_2 = 0.4$ .
      }
\label{fig:results:histograms}
    \vspace{-1\baselineskip}
\end{figure}
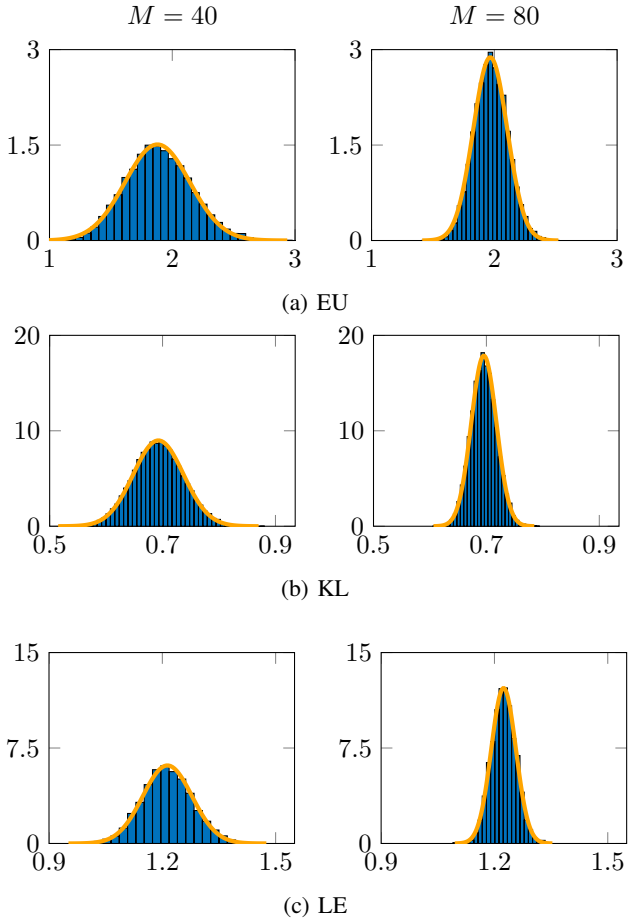

\subsection{Accuracy of the Asymptotic Distribution}

We start by comparing  the asymptotic distribution described in Theorem~\ref{th:cltEstimators} with the empirical distribution of the estimated distances with finite dimensions. Specifically, we are interested in the particularizations already discussed throughout this work,  namely Euclidean distance (EU), Symmetrized Kullback-Leibler (KL) and the log-Euclidean norm (LE) distances.   
Figure~\ref{fig:results:histograms} compares the histogram (in blue) and the asymptotic distributions (in orange) of these metrics for $c_1 = M/N_1=1/10,$ $c_2=M/N_2=1/2$ and some specific choices of $M$ and $\rho_1, \rho_2$. Observe that there is a very good match between the asymptotic and the empirical distribution regardless of the  considered metric and for relatively low system dimensions.

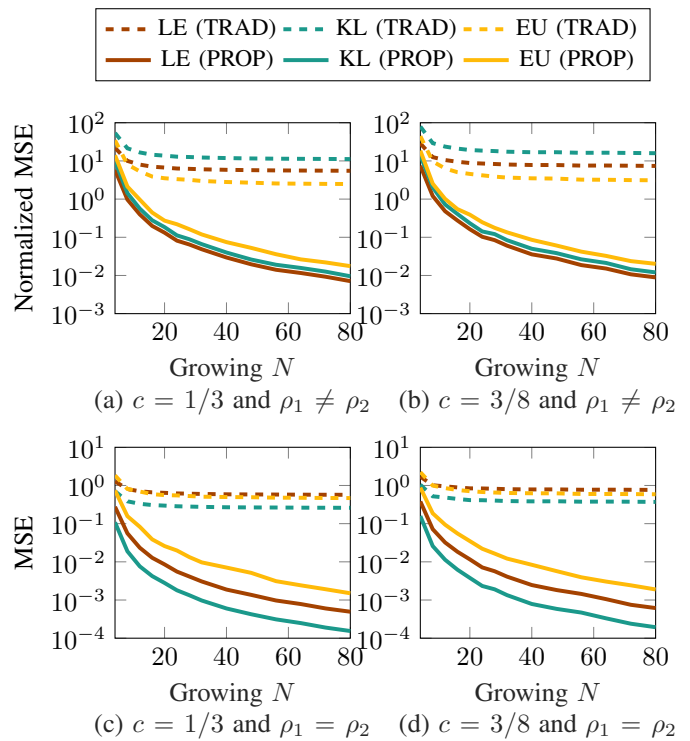
\begin{figure}[!b]
     \centering
     \hspace*{-0.15in} 
     \begin{subfigure}[b]{0.23\textwidth}
         \centering
        \input{figs/final_double_column/icassp/icassp_left}
     \end{subfigure}
     \hspace*{0.05in}
     \begin{subfigure}[b]{0.23\textwidth}
         \centering
        \input{figs/final_double_column/icassp/icassp_right}
     \end{subfigure}
      \caption{Relative MSE related to different metrics in different scenarios (a)-(d) with respect to the growth of $N=N_1=N_2$ ($x$--axis). In all these curves, the system dimension $M$ is scaled proportionally, so that $ c = M/N$ is constant.}
    \label{fig:results:growingsystem_tradvsprop}
    \vspace{-1\baselineskip}
\end{figure}

\subsection{Consistent Estimators vs Plug-in Distances}



We also compare the traditional \textit{plug-in} estimators to the proposed consistent estimators (\ref{eq:family_distances}) considering the different metrics studied in this work (EU, symmetrized KL, and LE distances).
We recall that both the traditional and our proposed methods rely on the information available in the sample covariance matrices $\hat{\mathbf{R}}_1$,  $\hat{\mathbf{R}}_2$ and try to approximate the true distance between the covariance matrices $\mathbf{R}_1$, $\mathbf{R}_2$. 
Figure~\ref{fig:results:growingsystem_tradvsprop}  illustrates the relative Mean Square Error (MSE) of these estimators compared to the true distance over $10^3$ samples for different choices of the coefficients $\rho_1, \rho_2$ and $c = M/N_1 = M/N_2$. 
For the proposed estimators (solid lines in the figures), when $\rho_1 \neq \rho_2$, the (Normalized) MSE between these quantities is given by
$$
\varepsilon_{\mathrm{PROP}} = \hat{\mathbb{E}}\left[\left(\frac{\hat{d}_M - d_M}{d_M}\right)^2\right],
$$
where the  empirical expectation ($\hat{\mathbb{E}}[\cdot]$) is obtained from $10^3$ simulations. Moreover, for $\rho_1 = \rho_2$, we have that $d_M = 0$ and the definition above becomes the absolute MSE value, i.e., $\varepsilon_{\mathrm{PROP}} = \hat{\mathbb{E}}[\hat{d}_M^2]$.   Following the same approach, one can also define $\varepsilon_\mathrm{TRAD}$ (dashed lines in the figures) by interchanging the proposed estimator $\hat{d}_M$ with the traditional \textit{plug-in} distance $\tilde{d}_M$. Figures~\ref{fig:results:growingsystem_tradvsprop}(a)-(b) represent the case where observations are drawn from  distinct processes ($\rho_1 = 0.3$ and $\rho_2 = 0.6$). 
In this scenario, it seems that the traditional estimators fail to converge to the actual distance between the two covariance matrices, whereas the MSE of the proposed estimators continuously decays with growing $M,N$. 
A similar behavior is displayed in Figures~\ref{fig:results:growingsystem_tradvsprop}(c)-(d), for the case where $\rho_1 = \rho_2 = 0.6$. In this  case, we have that $d_M = 0$ 
and hence the estimators should all converge to zero. We notice, however, that all three traditional \textit{plug-in} methods converge to another quantity away from zero, while our proposed estimators continuously decay as the system grows. Both these results corroborate the accuracy of the estimators proposed in this work, illustrating the advantage of the new estimators for relatively low values of $M,N$.

\subsection{Assessing Clustering Quality}

The results above also allow us to predict the expected behavior of a clustering algorithm based solely on the statistics of the underlying functionals.  This becomes particularly useful whenever one wants to theoretically evaluate the performance of a clustering algorithm before deploying it to real-world scenarios\footnote{We emphasize that the primary objective of this paper and this section is not to compare different metrics as this is highly problem dependent. Instead, here we showcase different applications of the framework we have provided above.}. Consider the problem of clustering  six sample covariance matrices $\hat{\mathbf{R}}_1,\ldots,\hat{\mathbf{R}}_6$, which are pairwise associated to the same covariance matrix (meaning that $\mathbf{R}_1=\mathbf{R}_2$, $\mathbf{R}_3=\mathbf{R}_4$ and $\mathbf{R}_5=\mathbf{R}_6$. 
Furthermore, assume for simplicity that $N_1=N_2$, $N_3=N_4$ and $N_5=N_6$. The asymptotic description above can be used to evaluate the probability of successful clustering of a particular estimated distance, understood to be the probability that all intra-cluster distances (distances built from sample covariance matrices associated to the same true covariance) are lower than all the inter-cluster distances (distances built from sample covariance matrices associated to different covariances). 


\begin{figure*}[!hbt]
     \centering
     \begin{subfigure}{0.9\textwidth}
{\centering\input{figs/results_prob_clust_emp_theo_metric_comp/legend_metrics}}
    \end{subfigure}
     \begin{subfigure}{0.3\textwidth}{\hspace{-2em}\input{figs/results_prob_clust_emp_theo_metric_comp/M_150_c1_0.67_c2_0.67_c3_0.67_c_pergroup}}
        \vspace{-2em}
        \caption*{\hspace{2em}(a) $c_1=\ldots=c_6=2/3$}
     \end{subfigure}
     \begin{subfigure}{0.32\textwidth}
    {\input{figs/results_prob_clust_emp_theo_metric_comp/M_150_c1_0.25_c2_0.33_c3_0.50_c_pergroup}}
       \vspace{-2em}
        \caption*{\hspace{2em}(b) $c_1=1/4, c_2=1/3, c_3=1/2$}
     \end{subfigure}
     \begin{subfigure}{0.32\textwidth}{\input{figs/results_prob_clust_emp_theo_metric_comp/M_150_c1_0.50_c2_0.33_c3_0.25_c_pergroup}}
       \vspace{-2em}
        \caption*{\hspace{2em}(c)  $c_1=1/2, c_2=1/3, c_3=1/4$}
     \end{subfigure}
      \caption{Empirical (solid lines) and theoretical (dashed lines) probability of correct clustering (y-axis) six sample covariance matrices into three groups for growing $M$ (x-axis) and fixed $\rho_1 = \rho_2=0.3, \rho_3=\rho_4 = 0.5, \rho_5 = \rho_6 = 0.7$ using proposed estimators.}
      \vspace{-1.5\baselineskip}
    \label{fig:results:prob_theoretical_vs_empirical}
 \end{figure*}
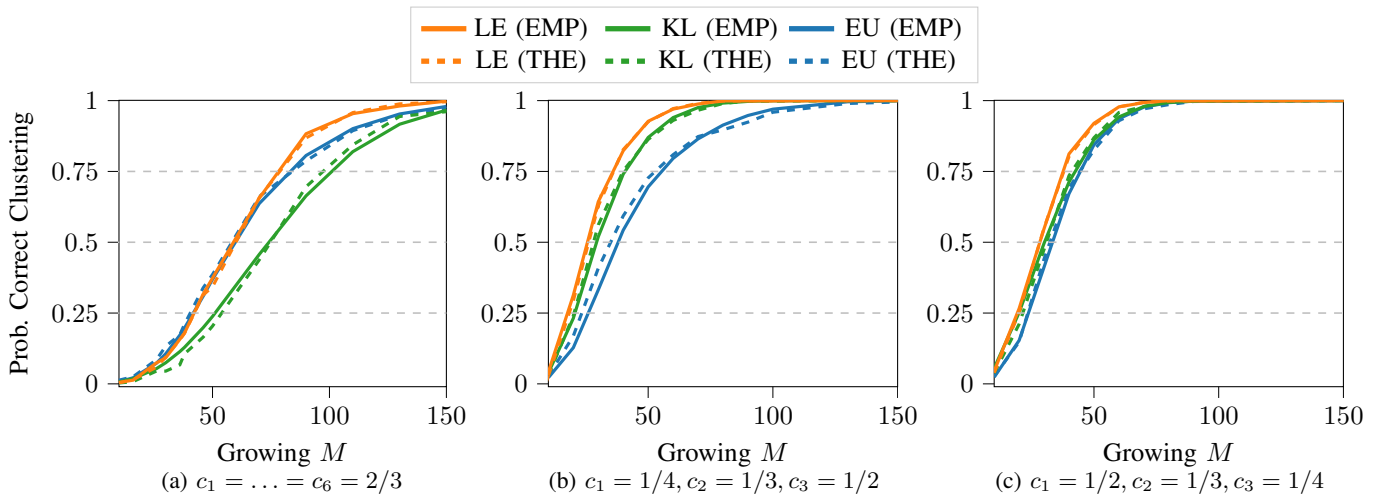

For the case of interest in this numerical example, we need to consider three intra-cluster distances 
and $12$ inter-cluster distances, denoted as {$\mathcal{D}_\mathrm{intra} = \{\hat{d}_M(1,2), \hat{d}_M(3,4), \hat{d}_M(5,6)\}$ and $\mathcal{D}_\mathrm{inter} = \left\{ \hat{d}_M(1,3),\hat{d}_M(1,4),\ldots \right\}$, respectively}. 
Depending on which is the minimum among the three intra-cluster distances, we can describe the clustering success probability as the probability of three disjoint events, that is
\begin{align*}
\mathbb{P}_\mathrm{succ} &= 
\mathbb{P} \left[ \{D_\mathrm{intra} < \hat{d}_M(1,2) \} 
\cap \{\hat{d}_M(1,2)<\min \mathcal{D}_\mathrm{inter} \}\right]  
\\
&+\mathbb{P} \left[  \{D_\mathrm{intra} < \hat{d}_M(3,4) \}  
\cap
\{\hat{d}_M(3,4)<\min \mathcal{D}_\mathrm{inter} \}\right]  
\\
&+\mathbb{P} \left[  \{D_\mathrm{intra} < \hat{d}_M(5,6) \} 
\cap \{\hat{d}_M(5,6)<\min \mathcal{D}_\mathrm{inter} \} \right].
\end{align*}
Each of these three probabilities can in turn be written as
$
\mathbb{P}\left( \mathbf{A}\hat{\mathbf{d}}_M  < 0\right)
$
where $\mathbf{d}_M$ is a column vector that contains all the distances as in (\ref{eq:defDistvect}) and $\mathbf{A}$ is a selection matrix with all the entries equal to zero except for one $+1$ and one $-1$ for each row, corresponding to the selection of distances to conform the different events that come into play into each argument of the three terms above. Since $\hat{\mathbf{d}}_M$ is asymptotically Gaussian distributed, so is the resulting column vector $\mathbf{A}\hat{\mathbf{d}}_M$. In fact, the transformed vector will also be asymptotically approximated as Gaussian distributed, with mean $\mathbf{A}(\mathbf{d}_M + \bar{\mathfrak{m}}_M / M)$ and covariance $\mathbf{A}(\bar{\mathbf{\Sigma}}_M / M^2)\mathbf{A}^T $. Hence, each of the three terms above can be evaluated by a multivariate Gaussian cumulative distribution function evaluated at zero. 
To evaluate the empirical probability of accurate clustering (referred to as ``Prob. Correct Clustering'' in the figures), we will utilize $10^3$ simulations. This clustering success rate is then defined as the percentage of realizations where the all estimated intra-cluster distances are lower than the smallest estimated inter-cluster distance.

We compare the empirical against the theoretical probability of correct clustering obtained from the different realizations of $\hat{\mathbf{R}}_j, j=1,\ldots, 6$ and by directly application of Theorem \ref{th:cltEstimators}, respectively. Specifically, Fig.~\ref{fig:results:prob_theoretical_vs_empirical} portrays these probabilities for fixed $\rho_1 = \rho_2=0.3, \rho_3=\rho_4 = 0.5, \rho_5 = \rho_6 = 0.7$, increasing $M, N_j$,  and different values of $c_j, j=1,\ldots, 6$. Recall from assumption \textbf{(As3)} that the constant $c_j$ is the limit of the quotient $M/N_j$, which is assumed fixed here. Note that there exists a very good alignment between empirical and theoretical probabilities of correctly clustering the sample covariance matrices, indicating a good prediction mechanism and the correctness of our results. This happens regardless of the scenario, i.e., different values of $c_j, j=1, \ldots, 6$. Moreover, during our experiments we noticed that, as the system grows large, the consistent estimator of the log-Euclidean distance outperformed all the other metrics. However,  this might not always be the case and in different applications, one metric may yield better results than another. In such scenarios, it becomes particularly useful to  perform a theoretical evaluation of various algorithms prior to their deployment in real-world applications.

\begin{figure*}
     \centering
     \begin{subfigure}{0.9\textwidth}
     \centering
{\input{figs/results_prob_clust_plug_vs_consistent/legend_multiple_setups}}
\end{subfigure}
     \begin{subfigure}{0.98\textwidth}
        {\hspace{-2em}\input{figs/results_prob_clust_plug_vs_consistent/tkz_eu_plugin_vs_consistent_rho_0.30_0.50_0.70}}{\input{figs/results_prob_clust_plug_vs_consistent/tkz_kl_plugin_vs_consistent_rho_0.30_0.50_0.70}}   {\input{figs/results_prob_clust_plug_vs_consistent/tkz_le_plugin_vs_consistent_rho_0.30_0.50_0.70}}
        \vspace{-2em}
        \caption*{(a) $\rho_1 = \rho_2=0.3, \rho_3=\rho_4 = 0.5, \rho_5 = \rho_6 = 0.7$ }
     \end{subfigure}
     \hfill
          \begin{subfigure}{0.98\textwidth}
        {\hspace{-2em}\input{figs/results_prob_clust_plug_vs_consistent/other_rho_val/tkz_eu_plugin_vs_consistent_rho_0.30_0.50_0.70_other}}{\input{figs/results_prob_clust_plug_vs_consistent/other_rho_val/tkz_kl_plugin_vs_consistent_esitmator_rho2}}   {\input{figs/results_prob_clust_plug_vs_consistent/other_rho_val/tkz_le_plugin_vs_consistent_rho_0.30_0.50_0.70_other}}
        \vspace{-2em}
        \caption*{(b) $\rho_1 = \rho_2=0.3, \rho_3=\rho_4 = 0.6, \rho_5 = \rho_6 = 0.9$ }
     \end{subfigure}
     \hfill
      \caption{Probability of correct clustering (y-axis) six SCMs into three groups for growing $M$ (x-axis). Results for traditional \textit{plug-in} estimator are depicted in dashed lines and consistent in solid lines. }
    \label{fig:results:prob_plugin_vs_consistent}
 \end{figure*}
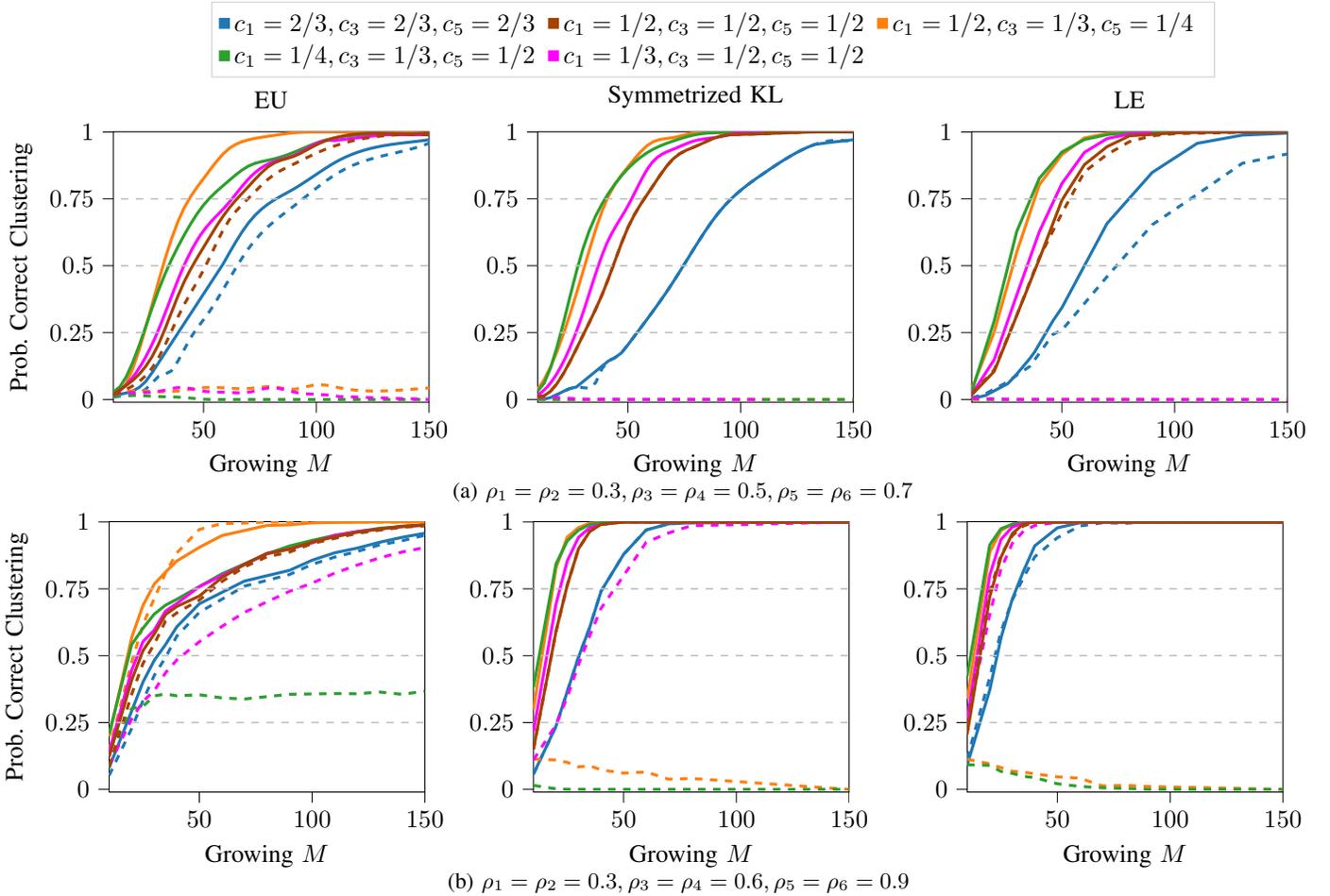

We also compare our proposed  estimators against the traditional \textit{plug-in} ones in Fig.~\ref{fig:results:prob_plugin_vs_consistent} for $\rho_1 = \rho_2=0.3, \rho_3=\rho_4 = 0.5, \rho_5 = \rho_6 = 0.7$ and $\rho_1 = \rho_2=0.3, \rho_3=\rho_4 = 0.6, \rho_5 = \rho_6 = 0.9$. 
The results for the \textit{plug-in} estimators (dashed lines in the figures) are empirically estimated from the $10^3$ different realizations of $\hat{\mathbf{R}}_j, j=1,\ldots,6$. For the sake of visualization, we present only the theoretical results of our proposed estimators (solid lines in the figures). 
By comparing the different metrics, we observe that our proposed estimator consistently outperforms the traditional \textit{plug-in} ones in the majority of the scenarios, as well as for large and small $M,N_j, j=1,\ldots, 6$.  This is a consequence of the design of our proposed method, which is tailored to better approximate the distance between the true covariance matrices. 
This behavior becomes more evident whenever the sample covariance matrices are built using different number of samples (e.g.,  $c_1=1/2, c_3=1/3, c_5=1/4$ and  $c_1=1/4, c_3=1/3, c_5=1/2$). In these scenarios, many of the traditional \textit{plug-in} distances fail to accurately cluster the sample covariance matrices, even for large $M$. Conversely, our proposed estimators consistently improve their probability of clustering  for growing $M, N_j, j=1,\ldots,6$. 
Finally, by comparing Fig.~\ref{fig:results:prob_plugin_vs_consistent}(a) and Fig.~\ref{fig:results:prob_plugin_vs_consistent}(b), we also observe that the \textit{plug-in} estimators are more sensitive to variations in $\rho_j, j=1,\ldots, 6$. For instance, in Fig.~\ref{fig:results:prob_plugin_vs_consistent}(a), for $c_1=c_2=1/3, c_3=c_4=1/2, c_5=c_6=1/2$ (magenta lines), all the \textit{plug-in} estimators completely fail to correctly cluster the different covariance matrices while in Fig.~\ref{fig:results:prob_plugin_vs_consistent}(b), these probabilities consistently grow with $M$. In other words, a slight variation in the values of $\rho_3=\rho_4$ and $\rho_5=\rho_6$ might severely change the  behavior of the \textit{plug-in} estimators, whereas our proposed estimators maintain a similar pattern, i.e., continuously increase its probability of correct clustering.

\section{Conclusion}
The estimation of the family of distances between sample covariance matrices that can be expressed as the sum of traces of analytic functions applied to each matrix separately has been studied. 
A general form for the consistent estimator of this particular family of distances has been proposed, along with its asymptotic characterization.
Moreover, we have particularized these results to three commonly used distances between covariance matrices, namely a symmetrized version of the Kullback–Leibler  divergence, the Euclidean distance and the Log-Euclidean distance. In particular, we have provided a closed form solution for the consistent estimators of all these three metrics and to their mean and variance (with exception to the Log-Euclidean distance whose fluctuations are estimated by solving an integral).  Results are established in the asymptotic regime where both the sample size and the observation dimension tend to infinity at the same rate. Numerical simulations confirm that our proposed estimators approximate the distance between covariance matrices better than the traditional \textit{plug-in} methods even for relatively low values of sample size and observation dimension. 
Finally, we also demonstrate the utility of the CLTs formulated in this study to estimate the probability of correct clustering sample covariance matrices using  different metrics.

\appendices

\section{Some useful lemmas}
The following lemmas, also presented in~\cite{pereira23tsp}, can be established using conventional tools in random matrix theory. We therefore omit their proofs.

\begin{lemma}
\label{lemma:UnifBounds}Let $j\in\{0,1\}$. Under \textbf{(As1)}-\textbf{(As3)} we have
\begin{gather}
\sup_{M}\sup_{z\in \mathrm{C}}\left\Vert \mathbf{Q}_{j}(\omega_{j}\left(  z\right)
)\right\Vert <+\infty\nonumber\\
0<\inf_{M}\inf_{z\in \mathrm{C}}\left\vert \omega_{j}\left(  z\right)  \right\vert
\leq\sup_{M}\sup_{z\in \mathrm{C}}\left\vert \omega_{j}\left(  z\right)  \right\vert
<+\infty\label{eq:boundednessw(z)}\\
\inf_{M}\mathrm{dist}\left\{  \omega_{j}\left(  z\right)  ,(-\infty,\mu_{\inf
}^{(j)}]\right\}  >0\nonumber
\end{gather}
Furthermore,
\begin{gather}
\sup_{M\geq M_{0}}\sup_{z\in \mathrm{C}}\left\Vert \hat{\mathbf{Q}}_{j}(z)\right\Vert
<+\infty\nonumber\\
0<\inf_{M\geq M_{0}}\inf_{z\in \mathrm{C}}\left\vert \hat{\omega}_{j}\left(  z\right)
\right\vert \leq\sup_{M\geq M_{0}}\sup_{z\in \mathrm{C}}\left\vert \hat{\omega}
_{j}\left(  z\right)  \right\vert <+\infty\label{eq:boundednessw(z)prime}\\
\inf_{M\geq M_{0}}\mathrm{dist}\left\{  \hat{\omega}_{j}\left(  z\right)
,(-\infty,\mu_{\inf}^{(j)}]\right\}  >0\nonumber
\end{gather}
with probability one for some $M_{0}$ sufficiently high.
\end{lemma}


\begin{lemma}
\label{lemma:UnifBounds2}Let $j\in\{0,1\}$. Under \textbf{(As1)}-\textbf{(As3)} we have
\[
\sup_{M}\sup_{z\in \mathrm{C}}
\frac{1}{N_{j}}
\mathrm{tr}\left[\mathbf{R}_j^2\mathbf{Q}^2\left(\omega_j(z)\right) \right] < 1.
\]

\end{lemma}



\section{Proof of Proposition \ref{prop:generalEstimator}}
\label{sec:proofPropEstimator}

We start by noticing that we can express $f_{j}^{(l)}\left(
\mathbf{R}_{j}\right)$ as in (\ref{eq:identityFintegral}), so that Proposition~\ref{prop:generalEstimator} will follow from proving that
\[
\sup_{z\in\mathrm{C}_j^{(l)}}\left\vert f_{j}^{(l)}(\omega_{j}\left(  z\right)
)\mathbf{Q}_{j}\left(  \omega_{j}\left(  z\right)  \right)  \omega_{j}
^{\prime}\left(  z\right)  -\hat{h}_{j}^{(l)}(z)\hat{\mathbf{Q}}
_{j}(z)\right\vert \asymp 0
\]
with probability one. 
%
From (\ref{eq:almostSureConvergenceTr})-(\ref{eq:defQbar}), we know that $\omega_{j}\left(
z\right)  \mathbf{Q}_{j}\left(  \omega_{j}\left(  z\right)  \right)  \asymp
z\hat{\mathbf{Q}}_{j}(z)$ for $z\in\mathbb{C}^{+}$. Then, by using the fact that
$$z\hat{\mathbf{Q}}_{j}(z)=-\mathbf{I}_{M}+\hat{\mathbf{R}}_{j}\hat{\mathbf{Q}
}_{j}(z)$$ (and equivalently for $\mathbf{Q}_{j}\left(  \omega_{j}\left(
z\right)  \right)$), we obtain $\mathbf{R}_{j}\mathbf{Q}_{j}\left(
\omega_{j}\left(  z\right)  \right)  \asymp\hat{\mathbf{R}}_{j}\hat{\mathbf
{Q}}_{j}(z)$. By Montel's theorem, one can extend this result to uniform
convergence on $\mathrm{C}_j^{(l)}$, so that
\[
\sup_{z\in\mathrm{C}_j^{(l)}}\left\vert \frac{1}{N_{j}}\mathrm{tr}\left[
\mathbf{R}_{j}\mathbf{Q}_{j}(\omega_{j}\left(  z\right)  )\right]  -\frac
{1}{N_{j}}\mathrm{tr}\left[  \hat{\mathbf{R}}_{j}\hat{\mathbf{Q}}
_{j}(z)\right]  \right\vert \rightarrow0
\]
almost surely. Now, noting that
\begin{equation} \label{eq:difomegas}
\omega_{j}\left(  z\right)  -\hat{\omega}_{j}\left(  z\right)  =\omega
_{j}\left(  z\right)  \frac{\mathrm{tr}\left[  \mathbf{R}_{j}\mathbf{Q}
_{j}(\omega_{j}\left(  z\right)  )-\hat{\mathbf{R}}_{j}\hat{\mathbf{Q}}
_{j}(z)\right]  }{N_{j}\left(  1-\frac{1}{N_{j}}\mathrm{tr}\left[
\hat{\mathbf{R}}_{j}\hat{\mathbf{Q}}_{j}(z)\right]  \right)  }
\end{equation}
and using the fact that\ $\sup_{z\in\mathrm{C}_j^{(l)}}\sup_{M}|\omega_{j}\left(
z\right)  |<+\infty$, together with (see Lemma \ref{lemma:UnifBounds})
\[
\inf_{z\in\mathrm{C}_j^{(l)}}\inf_{M}\left\vert 1-\frac{1}{N_{j}}\mathrm{tr}\left[
\hat{\mathbf{R}}_{j}\hat{\mathbf{Q}}_{j}(z)\right]  \right\vert >0,
\]
we can conclude that $\sup_{z\in\mathrm{C}_j^{(l)}
}|\omega_{j}\left(  z\right)  -\hat{\omega}_{j}\left(  z\right)
|\rightarrow0$ almost surely. Since these functions are holomorphic and the
above properties can be extended to an open subset that  includes $\mathrm{C}_j^{(l)}$, by
Montel\textquotesingle~s theorem, we automatically have $\sup_{z\in\mathrm{C}_j^{(l)}}|\omega
_{j}^{\prime}\left(  z\right)  -\hat{\omega}_{j}^{\prime}\left(  z\right)
|\rightarrow0$ with probability one. Consequently, we end up with
\begin{multline}
f_{j}^{(l)}(\omega_{j}\left(  z\right)  )\omega_{j}^{\prime}\left(  z\right)\mathbf{Q}_{j}\left(  \omega_{j}\left(  z\right)  \right)  -\hat{h}_{j}^{(l)}(z)\hat{\mathbf{Q}}_{j}(z)=\label{eq:integranddiff}
\\
\frac{\hat{h}_{j}^{(l)}(z)}{z}\left(  \omega_{j}\left(  z\right)\mathbf{Q}_{j}\left(  \omega_{j}\left(  z\right)  \right)  -z\hat{\mathbf{Q}}_{j}(z)\right) 
\\
+\left(  \frac{f_{j}^{(l)}(\omega_{j}\left(  z\right)  )\omega_{j}^{\prime}\left(  z\right)  }{\omega_{j}\left(  z\right)  }-\frac{f_{j}^{(l)}(\hat{\omega}_{j}\left(  z\right)  )\hat{\omega}_{j}^{\prime}\left(  z\right)}{\hat{\omega}_{j}\left(  z\right)  }\right) 
\times
\\
\times
\omega_{j}\left(  z\right)\mathbf{Q}_{j}\left(  \omega_{j}\left(  z\right)  \right).
\end{multline}
In what follows, we show that the terms on the right hand side of this equation are asymptotically equivalent to zero, which is sufficient to conclude the proof.

The first term is asymptotically equivalent to zero because 
$\omega_{j}\left(
z\right)  \mathbf{Q}_{j}\left(  \omega_{j}\left(  z\right)  \right)
-z\hat{\mathbf{Q}}_{j}(z)\asymp0$ and
\begin{equation}
\sup_{z\in\mathrm{C}_j^{(l)}}\left\vert \hat{h}_{j}^{(l)}(z)\right\vert =\sup
_{z\in\mathrm{C}_j^{(l)}}\left\vert f_{j}^{(l)}(\hat{\omega}_{j}\left(  z\right)
)\right\vert \sup_{z\in\mathrm{C}_j^{(l)}}\left\vert \frac{z\hat{\omega}_{j}^{\prime
}\left(  z\right)  }{\hat{\omega}_{j}\left(  z\right)  }\right\vert
<\infty\label{eq:boundedfprime}
\end{equation}
with probability one for all large $M$. 
Indeed, the second term on the right hand side of (\ref{eq:boundedfprime}) is finite (see Lemma\ \ref{lemma:UnifBounds}), and  the first one is also bounded because $\hat{\omega}_{j}\left(  z\right)  $ belongs to a compact interval
inside the analycity region of $F_{j}^{(l)}(\omega)\,$\ with probability one.

Now, the second term on the right hand side of
(\ref{eq:integranddiff}) can be studied by noting that 
$$\sup_{z\in\mathrm{C}_j^{(l)}
}\sup_{M \geq M_0}\left\Vert \omega_{j}\left(  z\right)  \mathbf{Q}_{j}\left(
\omega_{j}\left(  z\right)  \right)  \right\Vert <+\infty$$
for $M_0$ large enough and
\begin{multline*}
  \left\vert \frac{f_{j}^{(l)}(\omega_{j}\left(  z\right)  )\omega
_{j}^{\prime}\left(  z\right)  }{\omega_{j}\left(  z\right)  }-\frac
{f_{j}^{(l)}(\hat{\omega}_{j}\left(  z\right)  )\hat{\omega}_{j}^{\prime
}\left(  z\right)  }{\hat{\omega}_{j}\left(  z\right)  }\right\vert \leq
\\
 \leq \left\vert f_{j}^{(l)}(\hat{\omega}_{j}\left(  z\right)  )\right\vert
\left\vert \frac{\omega_{j}^{\prime}\left(  z\right)  }{\omega_{j}\left(
z\right)  }
-\frac{\hat{\omega}_{j}^{\prime}\left(  z\right)  }{\hat{\omega
}_{j}\left(  z\right)  }\right\vert +
\\
 +\left\vert \frac{\omega_{j}^{\prime}\left(  z\right)  }{\omega
_{j}\left(  z\right)  }\right\vert \left\vert f_{j}^{(l)}(\omega_{j}\left(
z\right)  )-f_{j}^{(l)}(\hat{\omega}_{j}\left(  z\right)  )\right\vert.
\end{multline*}
Then, reasoning as above, we immediately see that $$\sup_{M \geq M_0}\sup_{z\in\mathrm{C}_j^{(l)}}|f_{j}^{(l)}(\hat{\omega}_{j}\left(  z\right)  )|<\infty$$ 
and 
$$
\sup_{z\in\mathrm{C}_j^{(l)}}\left\vert\frac{\omega_{j}^{\prime}(z)}{\omega_{j}(z)}  -\frac{\hat{\omega}_{j}^{\prime}(z)}{\hat{\omega}_{j}(z)} \right\vert
\rightarrow 0
$$ with probability one. Regarding the second term, we
can use the fact that
\begin{multline*}
     f_{j}^{(l)}(\omega_{j}(z))-f_{j}^{(l)}(\hat{\omega}_{j}\left(
z\right)  )  =
\\
\frac{\hat{\omega}_j(z)-{\omega}_j(z)}{2\pi\mathrm{j}} 
\oint\nolimits_\gamma \frac{f_j^{(l)}(\zeta)}{\left( \zeta - \omega_j(z) \right) \left( \zeta - \hat{\omega}_j(z)  \right)} d\zeta  
\end{multline*}
where $\gamma$ is a negative contour in the region where $f^{(l)}_j(\zeta)$ is holomorphic that encloses both $\omega_j(z)$ and $\hat{\omega}_j(z)$. Thanks to the third and sixth properties of Lemma \ref{lemma:UnifBounds} it is possible to choose the contour $\gamma$ with these properties (possibly depending on $M$) such that $\inf_{M \geq M_0} \inf_{\zeta \in \gamma, z \in \mathrm{C}_j^{(l)}} \vert \zeta - {\omega}_j(z) \vert >0$ and $\inf_{M \geq M_0} \inf_{\zeta \in \gamma, z \in \mathrm{C}_j^{(l)}} \vert \zeta - \hat{\omega}_j(z) \vert >0$ almost surely for sufficiently large $M_0$. This implies that 
$$
\left\vert f_{j}^{(l)}(\omega_{j}(z))-f_{j}^{(l)}(\hat{\omega}_{j}\left(
z\right)  ) \right\vert
\leq K
|\omega_{j}\left(  z\right)  -\hat{\omega}_{j}\left(  z\right)  |
$$ 
for a certain $K>0$ independent of $M$. This directly shows that 
$$
\sup_{z \in \mathrm{C}_j^{(l)}} \left\vert f_{j}^{(l)}(\omega_{j}(z))-f_{j}^{(l)}(\hat{\omega}_{j}\left(
z\right)  ) \right\vert \rightarrow 0
$$
almost surely. 

Consider now the norm of $\hat{h}_{j}^{(l)}\left(  \hat{\mathbf{R}}
_{j}\right)  $ and observe that
\begin{multline*}
\left\Vert \hat{h}_{j}^{(l)}\left(  \hat{\mathbf{R}}_{j}\right)  \right\Vert
=
\sup_{\left\Vert \mathbf{u}\right\Vert =1}\mathbf{u}^{H}\hat{h}_{j}
^{(l)}\left(  \hat{\mathbf{R}}_{j}\right)  \mathbf{u}
\leq
\\
\leq
\sup_{\left\Vert \mathbf{u}\right\Vert =1}\frac{1}{2\pi}\oint
\nolimits_{\mathrm{C}_j^{(l)}}\left\vert \hat{h}_{j}^{(l)}(z)\right\vert \left\vert
\mathbf{u}^{H}\hat{\mathbf{Q}}_{j}(z)\mathbf{u}\right\vert \left\vert
dz\right\vert .
\end{multline*}
Now, obviously, $\vert \mathbf{u}^{H}\hat{\mathbf{Q}}_{j}(z)\mathbf{u}
\vert \leq\mathrm{dist}^{-1}(z,\mathcal{S}_{j}\cup\{0\})$ and, as claimed
in (\ref{eq:boundedfprime}), we have $\sup_{z\in\mathrm{C}_j^{(l)}} \vert  \hat
{h}_{j}^{(l)}(z) \vert <\infty$ almost surely for all large $M$. It
therefore follows that $\sup \Vert \hat{h}_{j}^{(l)} (  \hat{\mathbf
{R}}_{j}) \Vert <\infty$ with probability one for all large $M$.

\section{Solving the integral in (\ref{eq:integrallog2})}
\label{sec:AppendixIntegralLog}

In order to simplify the notation, whenever it is obvious, we will drop the dependence on $j\in\left\{
1,2\right\}  $ in all quantities (such as $\hat{\lambda}_{m}^{(j)}
$,~$\hat{\mathbf{e}}_{k}^{(j)}$, $\hat{\omega}_{j}\left(  z\right)  $, $N_{j}
$, or $\alpha^{(j)}$) within this appendix. On the other hand, we will
extensively use the fact that the eigenvalues $\hat{\lambda}_{k}$ are inside
the contour $\mathrm{C}$ almost surely for all large $M$. Hence, all the associated
statements should be understood to hold also with probability one and assuming
that $M$ is large enough (we will omit this detail throughout this appendix).

By using the expression for $\hat{\omega}\left(  z\right)  $ and $\hat{\omega
}^{\prime}\left(  z\right)  $ in (\ref{eq:defomegahat}) and
(\ref{eq:defomegahatprime}) respectively, we can immediately see that we need
to evaluate
\begin{multline*}
%
%
\alpha=\frac{1}{2\pi\mathrm{j}}\oint\nolimits_{\mathrm{C}^{-}}\log^{2}\left(\frac{1-\hat{\Psi}(z)}{z}\right)  \left(  \frac{1}{M}\sum_{k=1}^{M}\frac{1}{\hat{\lambda}_{k}-z}\right) \times
\\
\times 
\frac{1-\frac{M}{N}+\frac{1}{N}\sum_{m=1}^{M}\frac{z^{2}}{\left(  \hat{\lambda}_{m}-z\right) ^{2}}}{1-\hat{\Psi}(z)}dz
%
\end{multline*}
where we have defined
\[
\hat{\Psi}(z)=\frac{1}{N}\sum_{m=1}^{M}\frac{\hat{\lambda}_{m}}{\hat{\lambda
}_{m}-z}.
\]
To evaluate this integral, we first observe that $\log^{2}(  (1-\hat{\Psi}(z))/z)  =\log^{2}(  1-\hat{\Psi}(z))  -2\log z\log(  1-\hat{\Psi}(z) )  +\log^{2}(z)  $ and analyze
the three integrals separately. The first integral (containing $\log^{2}(  1-\hat{\Psi}(z))  $) is the one that is simpler to evaluate, because $\log(  1-\hat{\Psi}(z))  $ is holomorphic everywhere except for the intervals $\cup_{k=1}^{M}[\hat{\mu}_{k},\hat{\lambda}_{k}]$ where $\{\hat{\mu}_{k},k=1,\ldots,M\}$ are the solutions to the equation $\hat{\Psi}(\hat{\mu})=1$. Since these intervals are inside the contour $\mathrm{C}$ almost surely for all large $M$, one can conclude that the whole integrand is holomporphic outside $\mathrm{C}$. One can therefore enlarge $\mathrm{C}$ and consider the change of variable $\zeta=z^{-1}$, after which the only potential singularity will be at $\zeta=0$. It turns out that the resulting singularity at zero has residue equal to zero, so that the corresponding integral is zero as well, i.e., 
\begin{multline*}
\frac{1}{2\pi\mathrm{j}}\oint\nolimits_{\mathrm{C}^{-}}\log^{2}\left(  1-\hat{\Psi}(z)\right)  \left(  \frac{1}{M}\sum_{k=1}^{M}\frac{1}{\hat{\lambda}_{k}-z}\right) \times 
\\ \times
\frac{1-\frac{M}{N}+\frac{1}{N}\sum_{m=1}^{M}\frac{z^{2}}{\left(\hat{\lambda}_{m}-z\right)  ^{2}}}{1-\hat{\Psi}(z)}dz=0.
\end{multline*}
Now, the integral with the term $\log^{2}\left(  z\right)  $ is also easily solved by evaluating the residues at the singularities $\{\hat{\lambda}_{k},\hat{\mu}_{k}\}$ for $k=1,\ldots,M$, which are the only ones inside the contour $\mathrm{C}$. It follows that
\begin{multline*}
\frac{1}{2\pi\mathrm{j}}\oint\nolimits_{\mathrm{C}^{-}}\log^{2}z \left(  \frac{1}{M}\sum_{k=1}^{M}\frac{1}{\hat{\lambda}_{k}-z}\right) \times \\
\times \frac{1-\frac{M}{N}+\frac{1}{N}\sum_{m=1}^{M}\frac{z^{2}}{\left(  \hat {\lambda}_{m}-z\right)  ^{2}}}{1-\hat{\Psi}(z)}dz=
\\
=\left(  \frac{N}{M}-1\right)  \sum_{r=1}^{M}\log^{2}\left(  \hat{\mu}_{r}\right)  +2\frac{1}{M}\sum_{k=1}^{M}\log^{2}\left(  \hat{\lambda}
_{k}\right) +
\\
+2\frac{1}{M}\sum_{k=1}^{M}\log\left(  \hat{\lambda}_{k}\right)
-\frac{1}{M}\sum_{k=1}^{M}\sum_{\substack{r=1\\r\neq k}}^{M}\log^{2}\left(
\hat{\lambda}_{r}\right)  \frac{\hat{\lambda}_{r}}{\hat{\lambda}_{k}
-\hat{\lambda}_{r}} -
\\
-\frac{1}{M}\sum_{k=1}^{M}\log^{2}\left(  \hat{\lambda}_{k}\right)  \left(
\sum_{m=1}^{M}\frac{\hat{\lambda}_{m}}{\hat{\lambda}_{m}-\hat{\mu}_{k}}
-\sum_{\substack{m=1\\m\neq k}}^{M}\frac{\hat{\lambda}_{m}}{\hat{\lambda}
_{m}-\hat{\lambda}_{k}}\right)  .
\end{multline*}

It remains to compute the integral with the cross term $\log
z\log\left(  1-\hat{\Psi}(z)\right)  $. Let us denote
\begin{multline*}
\mathcal{I}=\frac{1}{2\pi\mathrm{j}}\oint\nolimits_{\mathrm{C}^{-}}\log z\log\left(
1-\hat{\Psi}(z)\right)  \left(  \frac{1}{M}\sum_{k=1}^{M}\frac{1}{\hat
{\lambda}_{k}-z}\right)  
\times \\ \times 
\frac{1-\frac{M}{N}+\frac{1}{N}\sum_{m=1}^{M}
\frac{z^{2}}{\left(  \hat{\lambda}_{m}-z\right)  ^{2}}}{1-\hat{\Psi}(z)}dz.
\end{multline*}
We observe that $\mathcal{I}=\mathcal{I}(1)$ where we have defined the
function $\mathcal{I}(x):[0,1]\rightarrow\mathbb{C}$ as
\begin{multline}
    \mathcal{I}(x)=\frac{1}{2\pi\mathrm{j}}\oint\nolimits_{\mathrm{C}^{-}}\log z\log\left(
1-x\hat{\Psi}(z)\right)  \times
\\ \times \left(  \frac{1}{M}\sum_{k=1}^{M}\frac{1}
{\hat{\lambda}_{k}-z}\right) 
\frac{1-\frac{M}{N}+\frac{1}{N}\sum_{m=1}
^{M}\frac{z^{2}}{\left(  \hat{\lambda}_{m}-z\right)  ^{2}}}{1-\hat{\Psi}
(z)}dz. \label{eq:definitionI(x)}
\end{multline}
The above function is continuously differentiable with respect to $x$, with
derivative
\begin{multline*}
\mathcal{I}^{\prime}(x)=\frac{1}{2\pi\mathrm{j}}\oint\nolimits_{\mathrm{C}^{+}}\log
z\left(  \frac{1}{M}\sum_{k=1}^{M}\frac{1}{\hat{\lambda}_{k}-z}\right)
 \times
\\ \times
\hat{\Psi}(z)   \frac{1-\frac{M}{N}+\frac{1}{N}\sum_{m=1}^{M}\frac{z^{2}}{\left(
\hat{\lambda}_{m}-z\right)  ^{2}}}{\left(  1-x\hat{\Psi}(z)\right)  \left(
1-\hat{\Psi}(z)\right)  }dz.
\end{multline*}
This is a consequence of the fact that the integrand of the above function is
uniformly bounded in $\mathrm{C}$, so that by the dominated convergence theorem we can
move the derivative with respect to $x$ inside the integration. The above
integral can easily be solved for $x\in(0,1)$ by noting that the only
singularities of the integrand inside $\mathrm{C}$ are the sample eigenvalues
$\hat{\lambda}_{m}$, the solutions to the equation $1=\hat{\Psi}(\hat{\mu})$,
namely $\hat{\mu}_{m}$, and the solutions to the equation $1=x\hat{\Psi}(z)$,
which will be denoted $\hat{\mu}_{m}\left(  x\right)  $, $m=1,\ldots,M.$ Using
conventional residue calculus, we can solve for any $x\in\left(  0,1\right)
$, leading to
\begin{multline*}
\mathcal{I}^{\prime}(x)    =\frac{1-N/M}{1-x}
\sum_{r=1}^{M}\log\hat{\mu}_{r} +\frac{1}{M}\sum_{r,k=1}^{M}\frac{\log\hat{\mu}_{r}\left(
x\right)  }{\hat{\lambda}_{k}-\hat{\mu}_{r}\left(  x\right)  }\hat{\mu}
_{r}^{\prime}\left(  x\right) 
\\
  +\frac{1}{\left(  1-x\right)  x}\left(
\frac{N}{Mx}-
1\right)  \sum_{r=1}^{M}\log\hat{\mu}_{r}\left(  x\right) 
  +\frac{1}{x}\frac{M+1}{M}\sum_{k=1}^{M}\log\hat{\lambda}_{k}
  \\
  +\frac{1}
{x}-\left(  \frac{1}{x}+1\right)  \frac{1}{x}\frac{N}{M}\sum_{k=1}^{M}\log
\hat{\lambda}_{k}
\end{multline*}
where we have used the fact that $\hat{\mu}_{m}\left(  x\right)  $ are
differentiable functions of $x$ with probability one, with derivative
\[
\hat{\mu}_{k}^{\prime}\left(  x\right)  =\left(  \frac{-x^{2}}{N}\sum
_{m=1}^{M}\frac{\hat{\lambda}_{m}}{\left(  \hat{\lambda}_{m}-\hat{\mu}
_{k}\left(  x\right)  \right)  ^{2}}\right)  ^{-1}.
\]
The above expression can be simplified by using the fact that (see \cite{Mestre08tsp,Schenck22})
\begin{equation}
1-\frac{M}{N}=\frac{\prod\nolimits_{r}\hat{\mu}_{r}}{\prod\nolimits_{r}
\hat{\lambda}_{r}}
\quad \text{and} \quad
1-\frac{xM}{N}=\frac{\prod\nolimits_{r}\hat{\mu
}_{r}\left(  x\right)  }{\prod\nolimits_{r}\hat{\lambda}_{r}}
\label{eq:identitiesProductmu}
\end{equation}
so that the derivative $\mathcal{I}^{\prime}(x)$ can alternatively be
expressed as
\begin{multline}
    \mathcal{I}^{\prime}(x)    =\frac{-1}{1-x}\left(  \frac{N}{M}-1\right)
\log\left(  1-\frac{M}{N}\right) 
\\
 +\frac{1}{\left(  1-x\right)  x}\left(
\frac{N}{Mx}-1\right)  \log\left(  1-\frac{xM}{N}\right)
\\
+\frac{1}{M}\sum_{k=1}^{M}\sum_{r=1}^{M}\frac{\log\hat{\mu}_{r}\left(
x\right)  }{\hat{\lambda}_{k}-\hat{\mu}_{r}\left(  x\right)  }\hat{\mu}
_{r}^{\prime}\left(  x\right)  
\\
+\frac{1}{x}\left(  1+\frac{1}{M}\sum_{k=1}
^{M}\log\hat{\lambda}_{k}\right)  .\label{eq:derivativeI(x)}
\end{multline}

From the above expression of the derivative $\mathcal{I}^{\prime}(x)$ it is easy to find a primitive as follows. The primitive of the first and the fourth term are trivial, so let us first focus on the second term. In order to obtain a primitive of this term, we recall the function $\Phi_{2}(x)$ introduced in (\ref{eq:definitionPhi(x)Li2}). By using the change of variables $t=1-\frac{xM}{N}$ and partial fraction decomposition one can show that
\begin{multline*}
\int\frac{1}{\left(  1-x\right)  x}\left(  \frac{N}{Mx}-1\right)  \log\left(
1-\frac{xM}{N}\right)  dx   =
\\
  =-\frac{N}{xM}\log\left(  1-\frac{xM}{N}\right)  +\log\left(  \frac
{1-\frac{xM}{N}}{\frac{xM}{N}}\right) 
\\
  +\left(  \frac{N}{M}-1\right)  \log\left(  1-\frac{xM}{N}\right)
\log\left(  x\frac{1-\frac{M}{N}}{1-x}\right) 
\\
  +\left(  \frac{N}{M}-1\right)  \left[  \Phi_{2}\left(  1-\frac{xM}
{N}\right)  -\Phi_{2}\left(  \frac{1-\frac{xM}{N}}{1-\frac{M}{N}}\right)
\right]  +K
\end{multline*}
where $K$ is an undetermined constant (its value may change from one line to the next) and where we have used the fact that
\begin{align}
\int\frac{\log t}{\lambda-t}dt  &  =\log t\log\left(  \frac{\lambda
}{\left\vert \lambda-t\right\vert }\right)  -\Phi_{2}\left(  \frac{t}{\lambda
}\right) + K \label{eq:identityLi2int}\\
\int\frac{\log t}{\left(  \lambda-t\right)  ^{2}}dt  &  =\frac{\log
t}{\lambda-t}-\frac{1}{\lambda}\log\left(  \frac
{t}{\left\vert \lambda-t\right\vert }\right) + K \nonumber
\end{align}
which can be readily proven by taking derivatives on both sides. 

Regarding the third term of (\ref{eq:derivativeI(x)}), one can easily show
using (\ref{eq:identityLi2int}) that
\begin{multline*}
\frac{1}{M}\sum_{k=1}^{M}\sum_{r=1}^{M}\int\frac{\log\hat{\mu}_{r}\left(
x\right)  }{\hat{\lambda}_{k}-\hat{\mu}_{r}\left(  x\right)  }\hat{\mu}
_{r}^{\prime}\left(  x\right)  dx=
\\
\frac{1}{M}\sum_{k=1}^{M}\sum_{r=1}^{M}
\log\hat{\mu}_{r}\left(  x\right)  \log \frac{\hat{\lambda}_{k}
}{\left\vert \hat{\lambda}_{k}-\hat{\mu}_{r}\left(  x\right)  \right\vert
} 
\\
-\frac{1}{M}\sum_{k=1}^{M}\sum_{r=1}^{M}\Phi_{2}\left(  \frac
{\hat{\mu}_{r}\left(  x\right)  }{\hat{\lambda}_{k}}\right)  +K.
\end{multline*}
Hence, putting everything together we can state that the primitive of
$\mathcal{I}^{\prime}(x)$ takes the form
\begin{multline*}
\mathcal{I}(x)   =\left(  \frac{N}{M}-1\right)  \log\left(  1-\frac{M}
{N}\right)  \log(1-x)  \\
+\left(  \frac{N}{M}-1\right)  \log\left(  1-\frac{xM}
{N}\right)  \log\left(  \left(  1-\frac{M}{N}\right)  \frac{x}{1-x}\right) \\
  +\left(  \frac{N}{M}-1\right)  \left[  \Phi_{2}\left(  1-\frac{xM}
{N}\right)  -\Phi_{2}\left(  \frac{1-\frac{xM}{N}}{1-\frac{M}{N}}\right)
\right]  
\\
-\frac{N}{xM}\log\left(  1-\frac{xM}{N}\right)  +\log\left(  \frac
{N}{M}\left(  1-\frac{xM}{N}\right)  \right) 
\\
+\frac{1}{M}\sum_{k=1}^{M}\sum_{r=1}^{M}\log\hat{\mu}_{r}\left(  x\right)
\log  \frac{\hat{\lambda}_{k}}{\left\vert \hat{\lambda}_{k}-\hat{\mu
}_{r}\left(  x\right)  \right\vert }
\\
  +\left(  \frac{1}{M}\sum_{k=1}^{M}\log\hat{\lambda}_{k}\right)  \log
x -\frac{1}{M}\sum_{k=1}^{M}
\sum_{r=1}^{M}\Phi_{2}\left(  \frac{\hat{\mu}_{r}\left(  x\right)  }
{\hat{\lambda}_{k}}\right)  + K.
\end{multline*}

The undetermined constant can be obtained by forcing $\mathcal{I}(0)=0$ (which
follows from the definition of $\mathcal{I}(x)$ in (\ref{eq:definitionI(x)})),
leading to
\begin{multline*}
K    =-\left(  \frac{N}{M}-1\right)  \left[  \Phi_{2}\left(  1\right)
-\Phi_{2}\left(  \frac{1}{1-\frac{M}{N}}\right)  \right]  -\log\left(
\frac{N}{M}\right) 
\\
  -\frac{\log\left(  N\right)  }{M}\sum_{k=1}^{M}\log\hat{\lambda}_{k}
-\frac{1}{M}\sum_{k=1}^{M}\sum_{\substack{r=1\\r\neq k}}^{M}\log\hat{\lambda
}_{r}\log\left(  \frac{\hat{\lambda}_{k}}{\left\vert \hat{\lambda}_{k}
-\hat{\lambda}_{r}\right\vert }\right)  
\\
+\frac{1}{M}\sum_{k=1}^{M}\sum_{r=1}^{M}\Phi_{2}\left(  \frac{\hat{\lambda}_{r}}{\hat{\lambda}_{k}}\right) - 1
\end{multline*}
where we have used the fact that $\hat{\mu}_{k}\left(  x\right)
\rightarrow\hat{\lambda}_{k}$ when $x\rightarrow0$, and also
\[
\lim_{x\rightarrow0}\frac{x\hat{\lambda}_{k}}{\hat{\lambda}_{k}-\hat{\mu}
_{k}\left(  x\right)  }=\lim_{x\rightarrow0}\left(  N-x\sum
_{\substack{m=1\\m\neq k}}^{M}\frac{\hat{\lambda}_{m}}{\hat{\lambda}_{m}
-\hat{\mu}_{k}\left(  x\right)  }\right)  =N.
\]
As a consequence of this, we can directly find
\begin{multline*}
I(1)  =\left(  \frac{N}{M}-1\right)\log\left(  1-\frac{M}
{N}\right) \left[\frac{1}{2} \log\left(  1-\frac
{M}{N}\right)  -  1 \right] 
\\
-1 
  +\frac{1}{M}\sum_{k=1}^{M}\sum_{r=1}^{M}\log\hat{\mu}_{r}\log\left(
\frac{\hat{\lambda}_{k}}{\hat{\lambda}_{k}-\hat{\mu}_{r}}\right) 
\\ 
-\frac{1}
{M}\sum_{k=1}^{M}\sum_{\substack{r=1\\r\neq k}}^{M}\log\hat{\lambda}_{r}
\log\left(  \frac{\hat{\lambda}_{k}}{\hat{\lambda}_{k}-\hat{\lambda}_{r}
}\right) 
\\
  +\frac{1}{M}\sum_{k=1}^{M}\sum_{r=1}^{M}\Phi_{2}\left(  \frac{\hat{\lambda
}_{r}}{\hat{\lambda}_{k}}\right)  -\frac{1}{M}\sum_{k=1}^{M}\sum_{r=1}^{M}
\Phi_{2}\left(  \frac{\hat{\mu}_{r}}{\hat{\lambda}_{k}}\right) 
\\
  -\log\left(  N\right)  \frac{1}{M}\sum_{k=1}^{M}\log\hat{\lambda}_{k}
\end{multline*}
where we have used the fact that $\mathrm{Li}_{2}\left(  1\right)  =\pi^{2}/6$
(so that $\Phi_{2}\left(  1\right)  =\pi^{2}/6$) together with the identity
(for $z\in(0,1)$)
\[
\Phi_{2}\left(  z\right)  +\Phi_{2}\left(  z^{-1}\right)  =\frac{\pi^{2}}
{3}-\frac{1}{2}\log^{2}z.
\]

The final expression for $\alpha$ is obtained by putting together all the above integrals 
and using the fact that \cite{mestre20tsp}
\[
-\frac{1}{N}\hat{\lambda}_{k}=\frac{\prod\nolimits_{r=1}^{M}\left(  \hat{\mu
}_{r}-\hat{\lambda}_{k}\right)  }{\prod\nolimits_{\substack{r=1\\r\neq k}
}^{M}\left(  \hat{\lambda}_{r}-\hat{\lambda}_{k}\right)  }
\]
and therefore
\begin{equation*}
\log N    =\sum_{r=1}^{M}\log\frac{\hat{\lambda}_{k}}{\left\vert \hat{\mu}_{r}
-\hat{\lambda}_{k}\right\vert }-\sum_{\substack{r=1\\r\neq k}}^{M}\log
\frac{\hat{\lambda}_{k}}{\left\vert \hat{\lambda}_{r}-\hat{\lambda}
_{k}\right\vert }.
\end{equation*}

\section{Proof of Theorem \ref{th:cltEstimators}}
\label{sec:proofclt}

We start by noting that the estimators $\hat{d}_{M}(i,j)$ can all be expressed as
\begin{multline*}
    \hat{d}_{M}(i,j)= \sum_{l=1}^{L} \frac{-1}{4\pi^{2}}\oint\nolimits_{\mathrm{C}_{i}^{(l)}}
\oint\nolimits_{\mathrm{C}_{j}^{(l)}}\hat{h}_i^{(l)}(z_i)\hat{h}_j^{(l)}(z_{j})
\times \\ \times
\frac{1}{M}
\mathrm{tr}\left[  \hat{\mathbf{Q}}_{i}(z_{i})\hat{\mathbf{Q}}_{j}
(z_{j})\right]  dz_{i}dz_{j}
\end{multline*}
with probability one for all large $M$, where $\hat{h}_j^{(l)}(z_{j})$ is defined in (\ref{eq:defhhat}). Now, let $h_j(z_j)$ denote the asymptotic equivalent of this quantity, defined as 
\begin{equation*}
h_{j}^{(l)}(z_{j})  = f_{j}^{(l)}(\omega_{j}\left(  z_{j}\right)
)\frac{z_{j}\omega_{j}^{\prime}\left(  z_{j}\right)  }{\omega_{j}\left(
z_{j}\right) } 
\end{equation*}
and note that, recalling the definition of $\mathbf{\bar{Q}}_{j}(z_{j})$ in
(\ref{eq:defQbar}) and by Cauchy integration, we can write the original distance that is to be estimated as 
\begin{multline}
    {d}_{M}(i,j)= \sum_{l=1}^{L} \frac{-1}{4\pi^{2}}\oint\nolimits_{\mathrm{C}_{i}^{(l)}}
\oint\nolimits_{\mathrm{C}_{j}^{(l)}} {h}_i^{(l)}(z_i) {h}_j^{(l)}(z_{j})
\times \\ \times
\frac{1}{M}\mathrm{tr}\left[  \bar{\mathbf{Q}}_{i}(z_{i})\bar{\mathbf{Q}}_{j}
(z_{j})\right]  dz_{i}dz_{j}.
\end{multline}
We begin by noticing that, using the short-hand notation $\mathbf{\bar{Q}}
_{j}=\mathbf{\bar{Q}}_{j}(z_{j})$ and $\hat{\mathbf{Q}}_{j}=\hat{\mathbf{Q}
}_{j}(z_{j})$, we have $\hat{\mathbf{Q}}_j \asymp \bar{\mathbf{Q}}_j$ almost surely according to (\ref{eq:almostSureConvergenceTr}). 
Using the fact that all
quantities inside the integral are bounded over the corresponding contours
that we can write
\begin{multline}    
M\left(  \hat{d}_{M}(i,j)  -d_{M}(i,j)\right) = 
\\
\sum_{l=1}^{L} \frac{-1}{4\pi^{2}}\oint\nolimits_{\mathrm{C}_{i}^{(l)}}\oint
\nolimits_{\mathrm{C}_{j}^{(l)}} h^{(l)}_i(z_{i})h_j^{(l)}(z_{j}) \hat{\xi}_{i,j}(z_i,z_j) dz_{i}dz_{j}  
 \\
 + \sum_{l=1}^{L} \frac{-1}{4\pi^{2}}\oint
\nolimits_{\mathrm{C}_{i}^{(l)}}\oint\nolimits_{\mathrm{C}_{j}^{(l)}}\mathrm{tr}\left[  \mathbf{\bar{Q}}_{i}\mathbf{\bar{Q}}_{j}\right]
\times \\ \times
\left(  \hat{h}_i^{(l)}(z_i) \hat{h}_j^{(l)}(z_j) -  {h}_i^{(l)}(z_i) {h}_j^{(l)}(z_j)  \right)  dz_{1}dz_{2} + \varepsilon_{i,j} 
\label{eq:error1inh}
\end{multline}
where we have defined 
\begin{equation} \label{eq:xiz1z2}
\hat{\xi}_{i,j}(z_i,z_j) = \mathrm{tr}\left[  \hat{\mathbf{Q}}_{i}\hat{\mathbf{Q}}
_{j}\right]  -\mathrm{tr}\left[  \mathbf{\bar{Q}}_{i}\mathbf{\bar{Q}}
_{j}\right].
\end{equation}
together with the error quantity
\begin{multline*}
    \varepsilon_{i,j} =  \sum_{l=1}^{L} \frac{-1}{4\pi^{2}}\oint
\nolimits_{\mathrm{C}_{i}^{(l)}}\oint\nolimits_{\mathrm{C}_{j}^{(l)}}
\Bigg(  \hat{h}_i^{(l)}(z_i) \hat{h}_j^{(l)}(z_j) 
\\
-{h}_i^{(l)}(z_i) {h}_j^{(l)}(z_j)  \Bigg) \hat{\xi}_{i,j}(z_i,z_j)  dz_{1}dz_{2}.
\end{multline*}
The following lemma shows that $\varepsilon_{i,j}$ converges to zero in probability, so that it does not contribute to the Central Limit Theorem.  
\begin{lemma}
    Under \textbf{(As1)}-\textbf{(As4)} we have $\varepsilon_{i,j} \rightarrow 0 $ in probability.
\end{lemma}
\begin{IEEEproof}
    Let us write $\theta^{(l)}_{i,j}(z_i,z_j) = M \left( \hat{h}_i^{(l)}(z_i) \hat{h}_j^{(l)}(z_j) -  {h}_i^{(l)}(z_i) {h}_j^{(l)}(z_j) \right) $. Now, for any given $\epsilon>0$ we observe that 
 \begin{multline*}
    \mathbb{P} \left( \left\vert \frac{1}{4 \pi^2} \oint\nolimits_{\mathrm{C}_i^{(l)}} \oint\nolimits_{\mathrm{C}_j^{(l)}} \frac{1}{M} \theta^{(l)}_{i,j}(z_i,z_j) \hat{\xi}_{i,j}(z_i,z_j) dz_i dz_j \right\vert > \epsilon \right) \leq
    \\ 
    \leq \mathbb{P} \Biggl( \frac{1}{4 \pi^2} \oint\nolimits_{\mathrm{C}_i^{(l)}} \oint\nolimits_{\mathrm{C}_j^{(l)}} \left\vert \frac{1}{M} \theta^{(l)}_{i,j}(z_i,z_j) \hat{\xi}_{i,j}(z_i,z_j)\right\vert 
     \times \\ \times 
     |dz_i| |dz_j|  > \epsilon \Biggl) \leq 
    \\
    \leq  \frac{1}{4 \pi^2 \epsilon M} \oint\nolimits_{\mathrm{C}_i^{(l)}} \oint\nolimits_{\mathrm{C}_j^{(l)}} \mathbb{E} \Bigg\vert {\theta}^{(l)}_{i,j}(z_i,z_j)
    \hat{\xi}_{i,j}(z_i,z_j) \mathds{1}_{\mathcal{L}_i} \mathds{1}_{\mathcal{L}_j}  \Bigg\vert 
     \times \\ \times |dz_i| |dz_j| 
    \end{multline*}
    where the last identity follows from Markov's inequality. In the above identity,  
    we have introduced $\mathcal{L}_i$ as the event that all the positive eigenvalues of $\hat{\mathbf{R}}_i$ fall within a sufficiently small blow-up of $\mathcal{S}_i$ and $\mathds{1}_{\{\cdot\}}$ denotes the indicator function.
    Therefore it is sufficient to show that $\mathbb{E}|{\theta}^{(l)}_{i,j}(z_i,z_j)\hat{\xi}_{i,j}(z_i,z_j) \mathds{1}_{\mathcal{L}_i} \mathds{1}_{\mathcal{L}_j} | < \infty$ 
    uniformly in $M$ and $(z_1,z_2) \in \mathrm{C}_i^{(l)} \times  \mathrm{C}_j^{(l)}$. By using the Cauchy-Schwarz inequality, we see that it is enough to prove that $\mathbb{E}[|{\theta}^{(l)}_{i,j}(z_i,z_j) |^2  \mathds{1}_{\mathcal{L}_i} \mathds{1}_{\mathcal{L}_j}]$ and $\mathbb{E}[|\hat{\xi}_{i,j}(z_i,z_j) |^2  \mathds{1}_{\mathcal{L}_i} \mathds{1}_{\mathcal{L}_j}]$ are both uniformly bounded. 

    Let us first consider the term $\mathbb{E}[|{\theta}^{(l)}_{i,j}(z_i,z_j) |^2  \mathds{1}_{\mathcal{L}_i} \mathds{1}_{\mathcal{L}_j}]$. Observe that we can express
\begin{align*}
 \theta^{(l)}_{i,j}(z_i,z_j)& =h_{j}^{(l)}(z_{j})M\left(  \hat{h}
_{i}^{(l)}(z_{i})-h_{i}^{(l)}(z_{i})\right) 
\\ & +h_{i}^{(l)}(z_{i})M\left(  \hat{h}_{j}^{(l)}(z_{j})-h_{j}^{(l)}(z_{j})\right) 
\\
&+ M \left(  \hat{h}
_{i}^{(l)}(z_{i})-h_{i}^{(l)}(z_{i})\right) \left(  \hat{h}_{j}^{(l)}(z_{j})-h_{j}^{(l)}(z_{j})\right) 
\end{align*}
where we have 
\[
\sup_{z_j \in \mathrm{C}_j^{(l)} } \vert h^{(l)}_j(z_j) \vert 
=\sup_{z_j\in\mathrm{C}_j^{(l)}}\left\vert f_{j}^{(l)}({\omega}_{j}(  z_j)
)\right\vert \sup_{z\in\mathrm{C}_j^{(l)}}\left\vert \frac{z{\omega}_{j}^{\prime
}\left(  z\right)  }{{\omega}_{j}\left(  z\right)  }\right\vert
<\infty
\]
for all $M$ sufficiently large, according to Lemmas \ref{lemma:UnifBounds} and \ref{lemma:UnifBounds2}. On the other hand, we also have 
\begin{multline*}
\vert\hat{h}_j^{(l)}(z)-h_j^{(l)}(z)\vert \leq |z| \left\vert f_j^{(l)}(\hat{\omega}_j(z) \right\vert \left\vert \frac{\omega^{\prime}_j(z)}{\omega_j(z)} -\frac{\hat{\omega}^{\prime}_j(z)}{\hat{\omega}_j(z)} \right\vert 
\\
+ |z| \left\vert \frac{\omega^{\prime}_j(z)}{\omega_j(z)} \right\vert \left\vert f_j^{(l)}(\omega_j(z)) - f_j^{(l)}(\hat{\omega}_j(z))\right\vert =
\\
=|z| \left\vert f_j^{(l)}(\hat{\omega}_j(z) \right\vert \left\vert \frac{\omega^{\prime}_j(z)}{\omega_j(z)} -\frac{\hat{\omega}^{\prime}_j(z)}{\hat{\omega}_j(z)} \right\vert 
\\
+ |z| \left\vert \frac{\omega^{\prime}_j(z)}{\omega_j(z)} \right\vert \left\vert \frac{\hat{\omega}_j(z)-{\omega}_j(z)}{2\pi\mathrm{j}} 
\oint\nolimits_\gamma \frac{f_j^{(l)}(\zeta)}{\left( \zeta - \omega_j(z) \right) \left( \zeta - \hat{\omega}_j(z)  \right)} d\zeta \right\vert 
\end{multline*}
where $\gamma$ is a negative contour in the region where $f^{(l)}_j(\zeta)$ is holomorphic that encloses both $\omega_j(z)$ and $\hat{\omega}_j(z)$. Now, according to Lemma \ref{lemma:UnifBounds} both $\omega_j(z)$ and $\hat{\omega}_j(z)$ are bounded away from $\mu_\mathrm{inf}^{(j)}$ uniformly in $M, z\in \mathrm{C}_j^{(l)}$ almost surely for all large $M$. Since these two quantities are also bounded in magnitude, one can choose the contour $\gamma$ independent of $M$ such that $\inf_{M>M_0} \inf_{\zeta \in \gamma, z \in \mathrm{C}_j^{(l)}} \min\{|\zeta -  \omega_j(z)|, |\zeta - \hat{\omega}_j(z) | \}>0$ with probability one for $M_0$ large enough. This implies that 
\begin{multline}
    \vert\hat{h}_j^{(l)}(z)-h_j^{(l)}(z)\vert \leq 
|z| \left\vert f_j^{(l)}(\hat{\omega}_j(z) \right\vert \left\vert \frac{\omega^{\prime}_j(z)}{\omega_j(z)}
-\frac{\hat{\omega}^{\prime}_j(z)}{\hat{\omega}_j(z)} \right\vert 
\times \\ \times
\left\vert \hat{\omega}_j(z)-{\omega}_j(z)\right\vert 
\end{multline}
for a certain positive $K>0$ independent of $z$ and $M$. 
Now, regarding the first term, we begin by noticing that one can express
\[
\frac{\hat{\omega}^{\prime}_j(z)}{\hat{\omega}_j(z)} = \frac{1}{z}+\frac{\frac{1}{N_j}\mathrm{tr}\left[\hat{\mathbf{R}}_j\hat{\mathbf{Q}}_j^2\right]}{1-\frac{1}{N_j}\mathrm{tr}\left[\hat{\mathbf{R}}_j^2\hat{\mathbf{Q}}_j^2\right]} =
\frac{1}{z} + \frac{\hat{\omega}_j(z)}{z}\frac{1}{N_j}\mathrm{tr}\left[\hat{\mathbf{R}}_j\hat{\mathbf{Q}}_j^2\right]
\]
whereas
\[
\frac{{\omega}^{\prime}_j(z)}{{\omega}_j(z)}
= \frac{1-\frac{1}{N_j}\mathrm{tr}\left[ \mathbf{R}_j \mathbf{Q}_j \right]}{z\left(1-\frac{1}{N_j}\mathrm{tr}\left[ \mathbf{R}_j
^2 \mathbf{Q}_j^2 \right]\right)}
=\frac{1}{z} + \frac{\omega_j(z)}{z}\frac{\frac{1}{N_j}\mathrm{tr}\left[\mathbf{R}_j
 \mathbf{Q}_j^2 \right]}{1-\frac{1}{N_j} \mathrm{tr}\left[\mathbf{R}_j
^2 \mathbf{Q}_j^2 \right]}.
\]
This implies that we can write 
\begin{multline*}
|z| \left\vert \frac{\omega^{\prime}_j(z)}{\omega_j(z)} -\frac{\hat{\omega}^{\prime}_j(z)}{\hat{\omega}_j(z)} \right\vert 
\leq
\left\vert \hat{\omega}_j(z)-\omega_j(z) \right\vert \left\vert \frac{1}{N_j} \mathrm{tr}\left[\hat{\mathbf{R}}_j \hat{\mathbf{Q}}_j^2 \right] \right\vert
\\
+ \left\vert \omega_j(z) \right\vert \left\vert  \frac{1}{N_j}\mathrm{tr}\left[\hat{\mathbf{R}}_j \hat{\mathbf{Q}}_j^2\right] - \frac{\frac{1}{N_j}\mathrm{tr}\left[\mathbf{R}_j
 \mathbf{Q}_j^2 \right]}{1-\frac{1}{N_j}\mathrm{tr}\left[\mathbf{R}_j
^2 \mathbf{Q}_j^2 \right]} \right\vert.
\end{multline*}
Now, using Lemmas \ref{lemma:UnifBounds} and \ref{lemma:UnifBounds2} together with the convexity of the square function we immediately see that 
\begin{multline} 
\label{eq:powerhdif}
\mathbb{E}\left[M^2|\hat{h}_j^{(l)}(z)-h_j^{(l)}(z)|^2 \mathds{1}_{\mathcal{L}_j} \right] \leq
\\ \leq 
K_1 \mathbb{E}\left[M^2|\hat{\omega}_j^{(l)}(z)-\omega_j^{(l)}(z)|^2 \mathds{1}_{\mathcal{L}_j} \right] 
 \\
 + K_2 \mathbb{E}\left[\left\vert  \mathrm{tr}\left[\hat{\mathbf{R}}_j \hat{\mathbf{Q}}_j^2\right] - \frac{\mathrm{tr}\left[\mathbf{R}_j
 \mathbf{Q}_j^2 \right]}{1-\frac{1}{N_j}\mathrm{tr}\left[\mathbf{R}_j
^2 \mathbf{Q}_j^2 \right]} \right\vert^2 \mathds{1}_{\mathcal{L}_j} \right]
\end{multline}
for two positive constants $K_1,K_2$ independent of $z$ and $M$. Now, the first term can be investigated using the identity in (\ref{eq:difomegas}), which directly shows that 
\begin{multline} \label{eq:powerWomegadif}
\mathbb{E}\left[\left\vert M\left(\hat{\omega}_j^{(l)}(z)-\omega_j^{(l)}(z)\right)\right\vert^2 \mathds{1}_{\mathcal{L}_j} \right]
\leq  \\ \leq 
\mathbb{E} \left[ \left\vert \frac{M}{N_j}\frac{\omega_j(z)\hat{\omega}_j(z)}{z} \mathrm{tr}\left[ \mathbf{R}_j\mathbf{Q}_j-\hat{\mathbf{R}}_j\hat{\mathbf{Q}}_j\right]  \right\vert^2 \mathds{1}_{\mathcal{L}_j} \right]
\leq  \\  \leq
K \mathbb{E}\left[\left\vert \mathrm{tr}\left[ \mathbf{R}_j\mathbf{Q}_j-\hat{\mathbf{R}}_j\hat{\mathbf{Q}}_j\right]   \right\vert^2 \mathds{1}_{\mathcal{L}_j}\right]
\end{multline}
where the last inequality follows from Lemmas \ref{lemma:UnifBounds} and \ref{lemma:UnifBounds2}. Now, the expressions in (\ref{eq:powerhdif}) and (\ref{eq:powerWomegadif}) can both be handled using \cite[Lemmas S-I \& S-II]{pereira23tsp}. For example, we have 
\begin{multline*}
\mathbb{E}\left[\left\vert \mathrm{tr}\left[ \mathbf{R}_j\mathbf{Q}_j-\hat{\mathbf{R}}_j\hat{\mathbf{Q}}_j\right]   \right\vert^2 \mathds{1}_{\mathcal{L}_j}\right] =
\\ = 
\left\vert  \mathrm{tr}\left[ {\mathbf{R}}_j{\mathbf{Q}}_j
- \mathbb{E}\left[\hat{\mathbf{R}}_j\hat{\mathbf{Q}}_j \mathds{1}_{\mathcal{L}_j}\right]
\right] \right\vert^2
+ \mathrm{var}\left[\mathrm{tr}\left[\hat{\mathbf{R}}_j\hat{\mathbf{Q}}_j\right]   \mathds{1}_{\mathcal{L}_j}\right]
\end{multline*}
and both therms are uniformly bounded according to \cite[Lemmas S-I]{pereira23tsp}. The term in (\ref{eq:powerWomegadif}) accepts a similar treatment. From all this, one can readily conclude that  $\mathbb{E}[|{\theta}^{(l)}_{i,j}(z_i,z_j) |^2  \mathds{1}_{\mathcal{L}_i} \mathds{1}_{\mathcal{L}_j}]$ is uniformly bounded. A similar bound can be found for $\mathbb{E}|\vert\hat{\xi}_{i,j}(z_i,z_j)\vert^2 \mathds{1}_{\mathcal{L}_i} \mathds{1}_{\mathcal{L}_j} | $ using the same approach. 
\end{IEEEproof}

Let us now go back to the proof of Theorem~\ref{th:cltEstimators}. As a direct consequence of the above result, we can drop the term $\varepsilon_{i,j}$ in (\ref{eq:error1inh}). Our next objective will be to show that the second term on the right hand side of (\ref{eq:error1inh}) can be expressed in a similar way as in the first one. Then, we will invoke a central limit theorem that was derived in \cite{pereira23tsp} for this type of statistic.
Let us therefore focus on the first step of the proof, which is summarized in the proposition that is presented below.

In order to introduce this result, we need some additional 
definitions. For $j \in \mathcal{J}$ and for a certain $M \times M$ deterministic matrix $\mathbf{A}$, consider the function
$\phi_{j}(\omega;\mathbf{A})$,  defined as
\begin{equation}
\phi_{j}(\omega;\mathbf{A})=\frac{\omega}{1-\Gamma_{j}\left(  \omega\right)
}\frac{1}{N_{j}}\mathrm{tr}\left[  \mathbf{R}_{j}\mathbf{Q}_{j}^{2}
(\omega)\mathbf{A}\right]  \label{eq:defphi}
\end{equation}
where $\Gamma_{j}\left(  \omega\right)$ is defined in (\ref{eq:defGamma}).
With these definitions, we are now ready to present the first 
step in the proof, which is summarized in the following proposition.
\begin{proposition} \label{prop:xi3terms}
Under assumptions \textbf{(As1)}-\textbf{(As4)} the quantity $M(  \hat{d}_{M}(i,j)-d_{M}(i,j))  - \hat{\chi}_M(i,j) $ converges in probability to zero, where
\begin{multline}
\hat{\chi}_M(i,j)=\sum_{l=1}^{L} \frac{-1}{4\pi^{2}}\oint
\nolimits_{\mathrm{C}_{i}^{(l)}}\oint\nolimits_{\mathrm{C}_{j}^{(l)}} h_i^{(l)}(z_i) h_j^{(l)}(z_j) 
\times \\ \times 
\hat{\xi}_{i,j}(z_i,z_j)
dz_{i}dz_{j} 
\\
-\sum_{l=1}^{L} \frac{-1}{4\pi^2}\oint\nolimits_{\mathrm{C}_{i}^{(l)}} \oint\nolimits_{\mathcal{Z}_{j}} f_i^{(l)}(\omega_i)\phi_i\left(\omega_i;f_j^{(l)}(\mathbf{R}_j)\right) 
\times \\ \times 
\hat{\xi}_{i,j}(z_i,z_j) dz_i dz_j 
\\
-\sum_{l=1}^{L} \frac{-1}{4\pi^2}\oint\nolimits_{\mathrm{C}_{j}^{(l)}} \oint\nolimits_{\mathcal{Z}_{i}} f_j^{(l)}(\omega_j)\phi_j\left(\omega_j;f_i^{(l)}(\mathbf{R}_i)\right)
\times \\ \times 
\hat{\xi}_{i,j}(z_i,z_j) dz_i dz_j  
\label{eq:defhatxi}
\end{multline}
where $\mathcal{Z}_j$, $j \in \mathcal{J}$, are negatively oriented simple contours enclosing $\mathcal{S}_j \cup \{0\}$.
\end{proposition}
\begin{IEEEproof}
It is sufficient to see that the second term on the
right hand side of (\ref{eq:error1inh}) coincides with the
sum of the second and third terms in the statement of the 
proposition up to a sequence of random variables that converge to zero in probability. 
To that effect, let us recall the definition 
\begin{equation}
    \theta^{(l)}_{i,j}(z_i,z_j) = M\left(  \hat{h}_{i}^{(l)}(z_{i})\hat{h}_{j}^{(l)}(z_{j})-h_{i}^{(l)}(z_{i})h_{j}^{(l)}(z_{j})\right) .
\end{equation}
It has been shown above that 
\begin{multline} 
\label{eq:decomposition2theta}
 \theta^{(l)}_{i,j}(z_i,z_j) =h_{j}^{(l)}(z_{j})M\left(  \hat{h}
_{i}^{(l)}(z_{i})-h_{i}^{(l)}(z_{i})\right) 
\\
+  h_{i}^{(l)}(z_{i})M\left(  \hat{h}_{j}^{(l)}(z_{j})-h_{j}^{(l)}(z_{j})\right)  +o_{p}(1)
\end{multline}
where here and in the rest of this proof we should understand $o_{p}(1)$ as
a function of $z_{i},z_{j}$ which converges in probability to zero uniformly
in $\mathrm{C}_{i}^{(l)}\times\mathrm{C}_{j}^{(l)}$. Now, using a Taylor approximation of
$f$ around $\omega_{j}\left(  z\right)  $, we see that
\[
N_{j}\left(  f_{j}^{(l)}(\hat{\omega}_{j})-f_{j}^{(l)}(\omega_{j})\right)
=f_{j}^{(l)\prime}(\omega_{j})N_{j}\left(  \hat{\omega}_{j}-\omega_{j}\right)
+R_j(z)
\]
where $f_{j}^{(l)\prime}(\omega_{j})$ is the derivative of $f_{j}^{(l)}
(\omega_{j})$ and where the residual takes the form
\[
R_j(z) = N_j \left(\hat{\omega}_j -\omega_j \right)^2 \frac{1}{2\pi\mathrm{j}} \oint\nolimits_\gamma \frac{f^{(l)}_j(\zeta)}{(\hat{\omega}_j-\zeta)(\zeta-{\omega}_j)^2} d\zeta
\]
where $\gamma$ is a negative contour on $\mathbb{C} \backslash (-\infty,\mu^{(j)}_{\inf} ]$ (where $f^{(l)}_j(\zeta)$ is holomorphic) that encloses both $\omega_j(z)$ and $\hat{\omega}_j(z)$. Thanks to Lemma \ref{lemma:UnifBounds} it is possible to choose this contour independently of $M$ such that the denominator of the integrand stays uniformly bounded away from zero. Therefore, using the fact that the family of random variables $N_j (\hat{\omega}_j-\omega_j)$, indexed by $M$ and $z \in \mathrm{C}_j^{(l)}$, is tight (see above), we can conclude that $\sup_{z \in \mathrm{C}_j^{(l)}} \vert R_j(z) \vert \rightarrow 0$ in probability. 

Now, observe that we can write 
\begin{multline*}
N_{j}\left(  \hat{h}_{j}^{(l)}(z)-h_{j}^{(l)}(z)\right)   = z\frac{f_{j}^{(l)}(\hat{\omega}_{j})}{\omega_{j}}N_{j}\left(
\hat{\omega}_{j}^{\prime}-\omega_{j}^{\prime}\right) 
\\
+z\left(  f_{j}^{(l)\prime}(\omega_{j})-\frac{f_{j}^{(l)}(\hat{\omega}
_{j})}{\hat{\omega}_{j}}\right)  \frac{\omega_{j}^{\prime}}{\omega_{j}}N_{j}\left(
\hat{\omega}_{j}-\omega_{j}\right) + z \frac{\omega_j^{\prime}}{\omega_j} R_j(z) 
\end{multline*}
where we have used the short-hand notation $\omega_{j}^{\prime}=\omega
_{j}^{\prime}\left(  z\right)  $ to denote the derivative of the function
$\omega_{j}\left(  z\right)  $. By conveniently grouping terms, we can alternatively write the above expression as
\begin{multline*}
N_{j}\left(  \hat{h}_{j}^{(l)}(z)-h_{j}^{(l)}(z)\right)   = z \frac{d}{dz}\left( \frac{f_j^{(l)}(\omega_j)}{\omega_j}N_j (\hat{\omega}_j-\omega_j) \right)  \\
+ z \frac{f_j^{(l)}(\hat{\omega}_j) - f_j^{(l)}(\omega_j)}{\omega_j}N_j(\hat{\omega}_j^\prime - \omega_j^\prime)
\\
+ z \left( \frac{f_j^{(l)}(\omega_j)}{\omega_j} -\frac{f_j^{(l)}(\hat{\omega}_j)}{\hat{\omega}_j} \right) \frac{\omega_j^\prime}{\omega_j} N_j (\hat{\omega}_j - \omega_j)
+ z \frac{\omega_j^{\prime}}{\omega_j} R_j(z) 
\end{multline*}
Following the above line of reasoning and the bounds in Lemma \ref{lemma:UnifBounds}, the last three terms of the above equation converge uniformly to zero in probability, so that
\begin{multline*}
    N_{j}\left(  \hat{h}_{j}^{(l)}(z)-h_{j}^{(l)}(z)\right)   = 
    \\z \frac{d}{dz}\left( \frac{f_j^{(l)}(\omega_j)}{\omega_j}N_j (\hat{\omega}_j-\omega_j) \right)  + o_p(1).
\end{multline*}
Now, consider a sequence of $M \times M$ deterministic matrices with bounded spectral norm $\mathbf{A}_M$. Using integration by parts. one can immediately write
\begin{multline} 
\label{eq:expansionIntegralAQ}
\frac{1}{2\pi\mathrm{j}}\oint\nolimits_{\mathrm{C}_j^{(l)}} N_j \left(\hat{h}_j^{(l)}(z)-h_j^{(l)}(z)\right) \frac{1}{N_j} \mathrm{tr} \left[\mathbf{A}_M\bar{\mathbf{Q}}_j(z)\right] dz =
\\ 
= \frac{1}{2\pi\mathrm{j}}\oint\nolimits_{\mathrm{C}_j^{(l)}}\frac{d}{dz} \left[\frac{f_j^{(l)}(\omega_j)}{\omega_j}N_j (\hat{\omega}_j-\omega_j) \right]  \omega_j
\times \\ \times 
\frac{1}{N_j} \mathrm{tr}\left[ \mathbf{A}_M {\mathbf{Q}}_j(\omega_j) \right]  dz + o_p(1)
\\= \frac{-1}{2\pi\mathrm{j}}\oint\nolimits_{\mathrm{C}_j^{(l)}}\frac{f_j^{(l)}(\omega_j)}{\omega_j}N_j (\hat{\omega}_j-\omega_j) 
\times \\ \times 
\frac{d}{dz} \left[ \omega_j\frac{1}{N_j} \mathrm{tr}\left[ \mathbf{A}_M {\mathbf{Q}}_j(\omega_j) \right] \right] dz + o_p(1)
\\= \frac{-1}{2\pi\mathrm{j}}\oint\nolimits_{\mathrm{C}_j^{(l)}}\frac{f_j^{(l)}(\omega_j)}{\omega_j^2}N_j (\hat{\omega}_j-\omega_j) \phi_j\left(\omega_j;\mathbf{A}_M\right) dz + o_p(1).
\end{multline}
We will use this result to simplify the second term on the right hand side of (\ref{eq:error1inh}). Indeed, using the decomposition of $\theta_{i,j}(z_i,z_j)$ in (\ref{eq:decomposition2theta}), this term can be written as
\begin{multline*}
\sum_{l=1}^{L} \frac{-1}{4\pi^{2}}\oint
\nolimits_{\mathrm{C}_{i}^{(l)}}\oint\nolimits_{\mathrm{C}_{j}^{(l)}} \frac{1}{M} \mathrm{tr}\left[  \mathbf{\bar{Q}}_{i}\mathbf{\bar{Q}}_{j}\right] \theta_{i,j}(z_i,z_j) dz_i dz_j = \\
 =  \sum_{l=1}^{L} \frac{1}{2\pi\mathrm{j}} \oint
\nolimits_{\mathrm{C}_{i}^{(l)}} \frac{1}{M} \mathrm{tr}\left[  \mathbf{\bar{Q}}_{i} f_j^{(l)} \left( \mathbf{R}_j \right) \right] 
\times \\ \times 
M\left(  \hat{h}_{i}^{(l)}(z_{i})-h_{i}^{(l)}(z_{i})\right) dz_i  
\\
+ \sum_{l=1}^{L} \frac{1}{2\pi\mathrm{j}} \oint\nolimits_{\mathrm{C}_{j}^{(l)}} \frac{1}{M} \mathrm{tr}\left[  f_i^{(l)}\right(\mathbf{R}_{i}\left)\mathbf{\bar{Q}}_{j}\right] 
\times \\ \times 
M\left(  \hat{h}_{j}^{(l)}(z_{j})-h_{j}^{(l)}(z_{j})\right) dz_j +o_p(1).
\end{multline*}
To convert the double integrals into single ones, we have used the fact that (introducing the short hand notation $\mathbf{Q}
_{j}=\mathbf{Q}_{j}(\omega_{j}(z_{j}))$)
\begin{align}
\frac{1}{2\pi\mathrm{j}}\oint\nolimits_{\mathrm{C}_{j}^{(l)}}h_{j}^{(l)}
(z_{j})\mathbf{\bar{Q}}_{j}dz_{j} &= 
\frac{1}{2\pi\mathrm{j}}\oint
\nolimits_{\mathrm{C}_{\omega_j}^{(l)}}f_{j}^{(l)}(\omega_{j})\mathbf{Q}_{j}d\omega_{j}
\\
&= f_{j}^{(l)}\left(  \mathbf{R}_{j}\right) \label{eq:integralhQ}
\end{align}
where we have used the conventional change of variable $z_j \mapsto \omega_j(z_j)$ and where $\mathrm{C}_{\omega_j}^{(l)} = \omega_j (\mathrm{C}_{j}^{(l)})$.
Hence, using the expansion in (\ref{eq:expansionIntegralAQ}) we immediately see that 
\begin{multline*}
\sum_{l=1}^{L} \frac{-1}{4\pi^{2}}\oint
\nolimits_{\mathrm{C}_{i}^{(l)}}\oint\nolimits_{\mathrm{C}_{j}^{(l)}} \frac{1}{M} \mathrm{tr}\left[  \mathbf{\bar{Q}}_{i}\mathbf{\bar{Q}}_{j}\right] \theta_{i,j}(z_i,z_j) dz_i dz_j = \\
 = - \sum_{l=1}^{L} \frac{1}{2\pi\mathrm{j}} \oint
\nolimits_{\mathrm{C}_{i}^{(l)}}
\frac{f_i^{(l)}(\omega_i)}{\omega_i^2} \phi_i\left(\omega_i;  f_j^{(l)} \left( \mathbf{R}_j \right)\right) N_i (\hat{\omega}_i-\omega_i) dz_i  \\
- \sum_{l=1}^{L} \frac{1}{2\pi\mathrm{j}} \oint
\nolimits_{\mathrm{C}_{j}^{(l)}}
\frac{f_j^{(l)}(\omega_j)}{\omega_j^2} \phi_j\left(\omega_j;  f_i^{(l)} \left( \mathbf{R}_i \right)\right) N_j (\hat{\omega}_j-\omega_j) dz_j  
\\
+o_p(1).
\end{multline*}
At this point, we only need to consider $N_j (\hat{\omega}_j -\omega_j)$, which can be transformed as
\begin{align*}
N_j (\hat{\omega}_j -\omega_j)&= \hat{\omega}_j\omega_j N_j \left(\frac{1}{\omega_j} -\frac{1}{\hat{\omega}_j} \right)
\\
&= {\hat{\omega}_j\omega_j} \mathrm{tr} \left[ \hat{\mathbf{Q}}_j - \bar{\mathbf{Q}}_j \right] 
\\
&= {\omega_j^2} \mathrm{tr} \left[ \hat{\mathbf{Q}}_j - \bar{\mathbf{Q}}_j \right]  +o_p(1).
\end{align*}
Now, let $\mathcal{Z}_{i}$ and $\mathcal{Z}_{j}$
denote two negatively oriented simple contours as in the statement of the proposition. 
We can now use the fact that (almost surely, for all $M$ sufficiently large)
\begin{equation} \label{eq:intToI}
\frac{1}{2\pi\mathrm{j}}\oint\nolimits_{\mathcal{Z}_{j}}\hat{\mathbf{Q}
}_{j}dz_{j}  = \frac{1}{2\pi\mathrm{j}}\oint\nolimits_{\mathcal{Z}_{j}}\mathbf{\bar{Q}
}_{j}dz_{j} =\mathbf{I}_{M}
\end{equation}
which directly leads to $$\mathrm{tr}[\hat{\mathbf{Q}}_i-\bar{\mathbf{Q}}_i] = \frac{1}{2\pi\mathrm{j}} \oint\nolimits_{\mathcal{Z}_{j}} \hat{\xi}_{i,j}(z_i,z_j) dz_j$$ 
with probability one for all $M$ large enough
and hence the statement of the proposition.  
\end{IEEEproof}

It directly follows from Proposition \ref{prop:xi3terms} that the asymptotic law of 
$M(\hat{d}_{M}(i,j)-d_{M}(i,j))$ coincides with that of $\hat{\chi}_M(i,j)$ in (\ref{eq:defhatxi}). 
To study the asymptotic behavior of this new sequence of random variables, we use a CLT derived in \cite{pereira23tsp}, which can be directly applied here. The CLT basically shows that random variables of the form
$\hat{\chi}_M(i,j)$ asymptotically fluctuate as Gaussian random variables with some predefined asymptotic (second order) mean and variance. Hence, the result in Theorem \ref{th:cltEstimators} will follow directly from \cite[Theorem 2]{pereira23tsp} after showing that the second order mean and asymptotic variance coincide with those in the statement of the theorem. We will analyze these two quantities separately. 

\subsection{Second order mean}

Let us first consider the second order mean vector, which is defined as $\bar{\boldsymbol{\mathfrak{m}}}_M$ and has its $r$th component equal to (see further \cite[Theorem 2]{pereira23tsp})
\begin{multline}
\{\bar{\boldsymbol{\mathfrak{m}}}_{M}\}_r  = \sum_{l=1}^{L} \frac{-\varsigma}{ 4\pi^2} \oint\nolimits_{\mathrm{C}_{i_r}^{(l)}}\oint\nolimits_{\mathrm{C}_{j_r}^{(l)}}\frac{\omega_{i_r}}{z_{i_r}
}\frac{\omega_{j_r}}{z_{j_r}} h^{(l)}_{i_r}(z_{i_r}) h^{(l)}_{j_r}(z_{j_r}) 
\times \\ \times 
\mathfrak{m}_{i_r,j_r}\left(  \omega_{i_r},\omega_{j_r}\right)  dz_{i_r}dz_{j_r} \\
-\sum_{l=1}^{L} \frac{-\varsigma}{ 4\pi^2} \oint\nolimits_{\mathrm{C}_{i_r}^{(l)}}\oint\nolimits_{\mathcal{Z}_{j_r}}\frac{\omega_{i_r}}{z_{i_r}
}\frac{\omega_{j_r}}{z_{j_r}} f^{(l)}_{i_r}(\omega_{i_r}) \phi_i\left( \omega_i;f^{(l)}_{j_r}\left(\mathbf{R}_{j_r}\right) \right)  
\times \\ \times 
\mathfrak{m}_{i_r,j_r}\left(  \omega_{i_r},\omega_{j_r}\right)  dz_{i_r}dz_{j_r} \\
-\sum_{l=1}^{L} \frac{-\varsigma}{ 4\pi^2} \oint\nolimits_{\mathrm{C}_{j_r}^{(l)}}\oint\nolimits_{\mathcal{Z}_{i_r}}\frac{\omega_{i_r}}{z_{i_r}
}\frac{\omega_{j_r}}{z_{j_r}} f^{(l)}_{j_r}(\omega_{j_r}) \phi_j\left( \omega_j;f^{(l)}_{i_r}\left(\mathbf{R}_{i_r}\right) \right)  
\times \\ \times 
\mathfrak{m}_{i_r,j_r}\left(  \omega_{i_r},\omega_{j_r}\right)  dz_{i_r}dz_{j_r} 
\label{eq:asymptMean}
\end{multline}
where $\phi_{j}(\omega;\mathbf{A})$ is defined in (\ref{eq:defphi}) and where we have introduced the bivariate functions 
$\mathfrak{m}_{i,j}\left(  \omega_{i},\omega_{j}\right)     =\mathfrak{m}_{i}\left(
\omega_{i},\mathbf{Q}_{j}\left(  \omega_{j}\right)  \right)  +\mathfrak{m}
_{j}\left(  \omega_{j},\mathbf{Q}_{i}\left(  \omega_{i}\right)  \right)$ where (for a $M \times M$ matrix $\mathbf{A}$) we have 
\begin{equation}
    \mathfrak{m}_{j}\left(  \omega_{j},\mathbf{A}\right)     =\frac{1}{N_{j}
}\frac{\mathrm{tr}\left[  \mathbf{\mathbf{R}}_{j}^{2}\mathbf{Q}_{j}^{3}\left(
\omega_{j}\right)  \Omega_{j}\left(  \omega_{j};\mathbf{A}\right)  \right]
}{1-\Gamma_{j}(\omega_{j})} \label{eq:defmofomegaA}
\end{equation} 
and with $\Omega_{j}(\omega;\mathbf{A})$ denoting
\begin{equation}
\Omega_{j}(\omega;\mathbf{A})=\mathbf{A}+\phi_{j}(\omega;\mathbf{A}.
)\mathbf{I}_{M} \label{eq:DefOmega}
\end{equation}
The above expression can be significantly simplified as follows. Consider the first term on the right hand side of the above expression, and observe that by using the change of variables $z \mapsto \omega_j(z)$ we can write (for general $i,j \in \mathcal{J}$)
\begin{multline*}
\frac{-1}{4\pi^2} \oint\nolimits_{\mathrm{C}_i^{(l)}} \oint\nolimits_{\mathrm{C}_j^{(l)}} 
\frac{\omega_i}{z_i} \frac{\omega_j}{z_j}h_i^{(l)}(z_i) h_j^{(l)}(z_j) \mathfrak{m}_{ij}(\omega_i,\omega_j)dz_i dz_j  =
\\
= \frac{-1}{4\pi^2} \oint\nolimits_{\mathrm{C}_{\omega_i}^{(l)}} \oint\nolimits_{\mathrm{C}_{\omega_j}^{(l)}} 
 f_i^{(l)}(\omega_i) f_j^{(l)}(\omega_j) \mathfrak{m}_{ij}(\omega_i,\omega_j)d\omega_i d\omega_j  
 \\
=\frac{1}{2\pi\mathrm{j}} \oint\nolimits_{\mathrm{C}_{\omega_i}^{(l)}} 
 f_i^{(l)}(\omega_i)  \mathfrak{m}_i\left(\omega_i, f_j^{(l)}\left(\mathbf{R}_j \right) \right) d\omega_i 
 \\
+\frac{1}{2\pi\mathrm{j}} \oint\nolimits_{\mathrm{C}_{\omega_j}^{(l)}} 
 f_j^{(l)}(\omega_j)  \mathfrak{m}_j\left(\omega_j, f_i^{(l)}\left(\mathbf{R}_i \right) \right) d\omega_j 
\end{multline*}
where we have used the linearity of $\Omega_j(\omega,\mathbf{A})$ in (\ref{eq:DefOmega}) as a function of $\mathbf{A}$. Regarding the second and third terms in (\ref{eq:asymptMean}), we simply note that (for any $j \in \mathcal{J}$)
\begin{multline*}
\frac{1}{2\pi\mathrm{j}}\oint\nolimits_{\mathcal{Z}_j} \frac{\omega_j}{z_j} \mathfrak{m}_{i,j}(\omega_i,\omega_j) dz_j =
\\
\mathfrak{m}_{i}(\omega_i,\mathbf{I}_M)
+\frac{1}{2\pi\mathrm{j}}\oint\nolimits_{\mathcal{Z}_j} \frac{\omega_j}{z_j} \mathfrak{m}_{j}(\omega_j,\mathbf{Q}(\omega_i)) dz_j 
\end{multline*}
where in the first term we have used (\ref{eq:intToI}). The integral of the second term on the right hand side of the above equation can easily be shown to be zero. Indeed, this can be shown by first applying the change of variables $z_j \mapsto \omega_j(z_j)$, so that 
\begin{multline*}
\frac{1}{2\pi\mathrm{j}}\oint\nolimits_{\mathcal{Z}_j} \frac{\omega_j}{z_j} \mathfrak{m}_{j}(\omega_j,\mathbf{Q}(\omega_i)) dz_j = 
\\
\frac{1}{2\pi\mathrm{j}}\oint\nolimits_{\mathcal{Z}_{\omega_j}} \frac{ \frac{1}{N_j} \mathrm{tr}\left[ \mathbf{R}_j^2\mathbf{Q}_j^3 \Omega_j (\omega_j; \mathbf{Q}_i)
\right] }{1-\frac{1}{N_j} \mathrm{tr}\left[ \mathbf{R}_j\mathbf{Q}_j\right] }  d\omega_j
\end{multline*}
where $\mathcal{Z}_{\omega_j} = \omega_j ( \mathcal{Z}_j)$. 
The resulting integral has all the singularities inside $\mathcal{Z}_{\omega_j}$, so that one can enlarge the contour to include $\{0\}$ if necessary and then apply a second change of variables $\omega_j \mapsto \zeta = \omega_j^{-1}$. The resulting integrand after the change of variables can be shown to be holomorphic at zero, implying that the integral is zero. With the above simplifications, the expression in (\ref{eq:asymptMean}) simplifies to
\begin{multline*}
\{\bar{\boldsymbol{\mathfrak{m}}}_{M}\}_r  = \sum_{l=1}^{L} 
\frac{\varsigma}{2\pi\mathrm{j}} \oint\nolimits_{\mathrm{C}_{\omega_{i_r}}^{(l)}} 
 f_{i_r}^{(l)}(\omega_{i_r})  
\mathfrak{m}_{i_r} \Bigg(\omega_{i_r}, f_{j_r}^{(l)}\left(\mathbf{R}_{j_r} \right)  
\\
 - \frac{\omega_{i_r}}{z_{i_r} \omega_{i_r}^{\prime}
}  \phi_{i_r} \left( \omega_{i_r};f^{(l)}_{j_r}\left(\mathbf{R}_{j_r}\right) \right) \mathbf{I}_M
 \Bigg) d\omega_{i_r} 
 \\
+ \sum_{l=1}^{L} \frac{\varsigma}{2\pi\mathrm{j}} \oint\nolimits_{\mathrm{C}_{\omega_{j_r}}^{(l)}} 
 f_{j_r}^{(l)}(\omega_{j_r})  \mathfrak{m}_{j_r} \Bigg(\omega_{j_r}, f_{i_r}^{(l)}\left(\mathbf{R}_{i_r} \right) 
 \\
 - \frac{\omega_{j_r}}{z_{j_r} \omega_{j_r}^{\prime}} \phi_{j_r}\left( \omega_{j_r};f^{(l)}_{i_r}\left(\mathbf{R}_{i_r}\right) \right) \mathbf{I}_M
 \Bigg) d\omega_{j_r} .
\end{multline*}
The expression of $\boldsymbol{\mathfrak{m}}_{M}$ in the statement of Theorem \ref{th:cltEstimators} is obtained by noticing that 
\begin{equation}
\Omega_{j}\left(  \omega_j;\mathbf{A}-\frac{\omega_{j}}{z_{j}\omega_{j}^{\prime
}}\phi_{j}(\omega_{j};\mathbf{A})\mathbf{I}_{M}\right)  =\mathbf{A}
\label{eq:OmegaIdentity}
\end{equation}
which directly implies that 
\[
\mathfrak{m}_j\left(\omega_j; \mathbf{A}-\frac{\omega_{j}}{z_{j}\omega_{j}^{\prime
}}\phi_{j}(\omega_{j};\mathbf{A})\mathbf{I}_{M}\right) = \bar{\mathfrak{m}}_j \left(\omega_j; \mathbf{A} \right)
\]
where $\bar{\mathfrak{m}}_j\left(\omega_j; \mathbf{A} \right)$ is defined as in (\ref{eq:defmofomegaA}) but replacing $\Omega_j(\omega_j;\mathbf{A})$ with $\mathbf{A}$. This directly leads to the expression of $\bar{\boldsymbol{\mathfrak{m}}}_M$ in the statement of Theorem \ref{th:cltEstimators}. 

\subsection{Asymptotic Variance}

Let us now consider the asymptotic covariance matrix $\bar{\boldsymbol{\Sigma}}_M$. 
With some abuse of notation, let us re-write (\ref{eq:defhatxi}) as 
\begin{equation} \label{eq:alternativexi}
\hat{\chi}_M(i,j) = \sum_{l=1}^L \frac{-1}{4\pi^2} \oint\nolimits \oint\nolimits g_{i,j}^{(l)}(z_i,z_j) \hat{\xi}_{i,j}(z_i,z_j) dz_i dz_j
\end{equation}
where we have defined the function 
\begin{align} 
\label{eq:defgfunction}
g_{i,j}^{(l)}(z_i,z_j) &= h_{i}^{(l)}(z_i)h_{j}^{(l)}(z_j)
- f_i^{(l)}(\omega_i)\phi_i\left(\omega_i; f_j^{(l)} \left( \mathbf{R}_j\right) \right)
\nonumber\\&
- f_j^{(l)}(\omega_i)\phi_j\left(\omega_j; f_i^{(l)} \left( \mathbf{R}_i\right) \right)
\end{align}
and where the integral contours in (\ref{eq:alternativexi}) are different depending on the term that is being integrated, i.e. $\mathrm{C}_i^{(l)} \times \mathrm{C}_j^{(l)}$, $\mathrm{C}_i^{(l)} \times \mathcal{Z}_j$ and $\mathcal{Z}_i \times \mathrm{C}_j^{(l)}$ for the first, second and third term in $g_{i,j}^{(l)}(z_i,z_j)$ respectively.
Accepting this abuse of notation, the $(r,s)$th entry of the asymptotic covariance matrix can be compactly expressed as \cite[Theorem 2]{pereira23tsp}
\begin{multline} \label{eq:asymptVariance}
 \{\bar{\boldsymbol{\Sigma}}_M\}_{r,s}  
  =\sum_{l_r,l_s=1}^{L} \frac{1+\varsigma}{\left(  2\pi\mathrm{j}\right)  ^{4}}
\oint\nolimits \oint\nolimits \oint\nolimits \oint\nolimits
\frac{\omega_{i_r}\omega_{j_r}}{z_{i_r}z_{j_r}}\frac{\tilde{\omega}_{i_s}\tilde{\omega}_{j_s}}{\tilde{z}_{i_s}\tilde{z}_{j_s}} \times
\\\times g_{i_r,j_r}^{(l_r)}(z_{i_r},z_{j_r}) 
g_{i_s,j_s}^{(l_s)}(\tilde{z}_{i_s},\tilde{z}_{j_s}) 
\times  
\\
\times
\sigma_{i_r,j_r,i_s,j_s}
^{2}\left(  \omega_{i_r},\omega_{j_r},\tilde{\omega}_{i_s},\tilde{\omega}_{j_s}\right)  dz_{i_r}dz_{j_r} d\tilde{z}_{i_s} d\tilde{z}_{j_s} 
\end{multline}
where we are again using the short-hand notation $\omega_j = \omega_j(z_j), \tilde{\omega}_j=\omega_j(\tilde{z}_j)$, $j \in \mathcal{J}$, and where the variance function $\sigma^{2}_{i,j,m,n} ( \omega_{i},\omega_{j},\tilde{\omega}_{m},\tilde{\omega}_{n}
)$ is defined as follows for $i,j,m,n \in \mathcal{J}$, $i \neq j, m \neq n$. This function can be decomposed into six terms as follows
\begin{align}
\sigma^{2}_{i,j,m,n}&\left(  \omega_{i},\omega_{j},\tilde{\omega}_{m},\tilde{\omega}_{n}\right) = 
\nonumber \\ &=
\sigma_{i}^{2}\left(  \omega_{i},\tilde{\omega}_{m};\mathbf{Q}_{j}\left(
\omega_{j}\right)  ,\mathbf{Q}_{n}(\tilde{\omega}_{n})\right) \delta_{i=m} + 
\nonumber \\& 
+ \sigma_{j}^{2}\left(  \omega_{j},\tilde{\omega}_{n};\mathbf{Q}_{i}\left(
\omega_{i}\right)  ,\mathbf{Q}_{m}(\tilde{\omega}_{m})\right) \delta_{j=n} + 
\nonumber \\&
+ \sigma_{i}^{2}\left(  \omega_{i},\tilde{\omega}_{n};\mathbf{Q}_{j}\left(
\omega_{j}\right)  ,\mathbf{Q}_{m}(\tilde{\omega}_{m})\right) \delta_{i=n} + 
\nonumber \\&
+ \sigma_{j}^{2}\left(  \omega_{j},\tilde{\omega}_{m};\mathbf{Q}_{i}\left(
\omega_{i}\right)  ,\mathbf{Q}_{n}(\tilde{\omega}_{n})\right) \delta_{j=m} + \nonumber\\
&+ \varrho_{i,j}(\omega_i,\omega_j,\tilde{\omega}_m,\tilde{\omega}_n) \delta_{i=m}\delta_{j=n} +
\nonumber \\&
+ \varrho_{i,j}(\omega_i,\omega_j,\tilde{\omega}_n,\tilde{\omega}_m) \delta_{i=n}\delta_{j=m} \label{eq:Sigma2}
\end{align}
where each of the six terms is nonzero only when at least two of the original indices coincide. The functions $\sigma_{j}^{2}\left(  \omega,\tilde{\omega};\mathbf{A},\mathbf{B}\right) $ for $j \in \mathcal{J} $ are defined as
\begin{align}
\sigma_{j}^{2}&\left(  \omega,\tilde{\omega};\mathbf{A},\mathbf{B}\right)   
=\frac{1}{1-\Gamma_{j}(\omega,\tilde{\omega})}\frac{1}{N_{j}}\mathrm{tr}\Biggl[  \mathbf{R}_{j}\mathbf{Q}_{j}\left(  \omega\right)  \mathbf{Q}_{j}\left(  \tilde{\omega} \right)  
\times\nonumber\\&\times
\Omega_{j}\left(  \omega;\mathbf{A}\right) 
\mathbf{R}_{j}\mathbf{Q}_{j}\left(  \omega\right)  \mathbf{Q}_{j}\left( \tilde{\omega} \right)  \Omega_{j}\left(  \tilde{\omega} ;\mathbf{B}\right)  \Biggl] +
\nonumber \\
&  +\frac{1}{\left(  1-\Gamma_{j}(\omega,\tilde{\omega})\right)  ^{2}}
\frac{1}{N_{j}}\mathrm{tr}\left[  \mathbf{R}_{j}^{2}\mathbf{Q}_{j}^{2}\left(
\omega\right)  \mathbf{Q}_{j}\left(  \tilde{\omega}\right)  
\Omega_{j}\left(
\omega;\mathbf{A}\right)  \right]  \times 
\nonumber \\ & \times \frac{1}{N_{j}}\mathrm{tr}\left[
\mathbf{R}_{j}^{2}\mathbf{Q}_{j}\left(  \omega\right)  \mathbf{Q}_{j}
^{2}\left(  \tilde{\omega}\right)  \Omega_{j}\left(  \tilde{\omega} ;\mathbf{B}\right)  \right] \label{eq:omegabivariate}
\end{align} 
where we have introduced the quantities $\Gamma_{j}(\omega,\tilde{\omega})$ as defined in (\ref{eq:defGammabivariate}),  whereas the functions $\varrho_{i,j}(\omega_i,\omega_j,\tilde{\omega}_i,\tilde{\omega}_j) $ are defined as 
\begin{align}
\varrho_{i,j}&(\omega_i,\omega_j,\tilde{\omega}_i,\tilde{\omega}_j) 
\nonumber\\&
=
\frac{\mathrm{tr}^{2}\left[  \mathbf{R}_{i}\mathbf{Q}_{i}\left(  \omega
_{i}\right)  \mathbf{Q}_{i}\left(  \tilde{\omega}_{i}\right)  \mathbf{R}
_{j}\mathbf{Q}_{j}\left(  \omega_{j}\right)  \mathbf{Q}_{j}\left(  \tilde{\omega}
_{j}\right)  \right]  }{N_{i}N_{j}\left(  1-\Gamma_{i}(\omega
_{i},\tilde{\omega}_{i})\right)  \left(  1-\Gamma_{j}(\omega_{j},\tilde{\omega}
_{j})\right)  }.  \label{eq:defvarrho}
\end{align}

Now, we observe that some of the terms of the function $g_{i,j}^{(l)}(z_1,z_2)$ in (\ref{eq:defgfunction}) depend only on one of the integration variables, which will significantly allow us to simplify the expression of the asymptotic covariance. For example, the second term in (\ref{eq:defgfunction}) does not depend on $z_j$, and the corresponding integral in (\ref{eq:asymptVariance}) can be formulated as
\[
\frac{1}{2\pi\mathrm{j}}\oint\nolimits_{\mathcal{Z}_j} 
\frac{\omega_j}{z_j}\sigma^2_{i,j,m,n}(\omega_i,\omega_j,\tilde{\omega}_m,\tilde{\omega}_n) dz_j.
\]
This integral can be solved in closed form as follows. Since the function $\sigma^2_{i,j,m,n}(\omega_i,\omega_j,\tilde{\omega}_m,\tilde{\omega}_n) $ is essentially composed by a sum of terms in the form of (\ref{eq:omegabivariate}) and (\ref{eq:defvarrho}), we need to solve the integral for these two constituent functions. 
Let us first start with the integral with respect to the term $\varrho_{i,j}(\omega_i,\omega_j,\tilde{\omega}_i,\tilde{\omega}_j)$. By applying the change of variables $z_j \mapsto \omega_j(z_j)$ we have 
\begin{multline*}
    \frac{1}{2\pi\mathrm{j}}\oint\nolimits_{\mathcal{Z}_j} 
\frac{\omega_j}{z_j}\varrho_{i,j}(\omega_i,\omega_j,\tilde{\omega}_i,\tilde{\omega}_j) dz_j = 
\\
\frac{1}{2\pi\mathrm{j}}\oint\nolimits_{\mathcal{Z}_{\omega_j}}
\varrho_{i,j}(\omega_i,\omega_j,\tilde{\omega}_i,\tilde{\omega}_j) \frac{1-\Gamma_j(\omega_j)}{1-\frac{1}{N_j} \mathrm{tr}\left[\mathbf{R}_j \mathbf{Q}(\omega_j)\right] } d\omega_j 
\end{multline*}
where $\mathcal{Z}_{\omega_j}=\omega_j(\mathcal{Z}_j)$ encloses all the singularities of the integrand. Therefore, one can apply a second change of variables $\omega_j \mapsto \zeta(\omega_j) = \omega_j^{-1}$ leading to 
\begin{multline} 
\frac{1}{2\pi\mathrm{j}}\oint\nolimits_{\mathcal{Z}_j} 
\frac{\omega_j}{z_j}\varrho_{i,j}(\omega_i,\omega_j,\tilde{\omega}_i,\tilde{\omega}_j) dz_j = 
\\
\frac{1}{2\pi\mathrm{j}}\oint\nolimits_{\mathrm{C}_{0}} 
\varrho_{i,j}(\omega_i,\zeta^{-1},\tilde{\omega}_i,\tilde{\omega}_j) \frac{1-\Gamma_j(\zeta^{-1})}{1-\frac{1}{N_j} \mathrm{tr}\left[\mathbf{R}_j \mathbf{Q}(\zeta^{-1})\right] } \frac{1}{\zeta^2} d\zeta  
\label{eq:integralvarrho}
\end{multline}
where $\mathrm{C}_{0}$ is a negatively oriented contour enclosing $\{0\}$. Now, recalling the expression of $\varrho_{i,j}(\omega_i,\zeta^{-1},\tilde{\omega}_i,\tilde{\omega}_j)$ in (\ref{eq:defvarrho}) we readily see that the integrand on the right hand side of the above equation has a removable singularity at zero, implying that the corresponding integral is zero. Using exactly the same approach, one can readily see that 
\[
\frac{1}{2\pi\mathrm{j}}\oint\nolimits_{\mathcal{Z}_j} 
\frac{\omega_j}{z_j} \sigma_{j}^2(\omega_j,\tilde{\omega_j},\mathbf{A},\mathbf{B}) dz_j = 0.
\]
Regarding the terms $\sigma_{i}^2(\omega_i,\tilde{\omega_i},\mathbf{Q}_j(\omega_j),\mathbf{B})$, $i \neq j$,
we can use (\ref{eq:intToI}) and the linearity of $\sigma_{i}^2(\omega_i,\tilde{\omega_i},\mathbf{A},\mathbf{B})$ as a function of $\mathbf{A},\mathbf{B}$ to obtain 
\[
\frac{1}{2\pi\mathrm{j}}\oint\nolimits_{\mathcal{Z}_j} 
\frac{\omega_j}{z_j} \sigma_{i}^2(\omega_i,\tilde{\omega_i},\mathbf{Q}_j(\omega_j),\mathbf{B}) dz_j = \sigma_{i}^2(\omega_i,\tilde{\omega_i},\mathbf{I}_M,\mathbf{B}).
\]
From all the above, one can conclude that 
\begin{align} 
\frac{1}{2\pi\mathrm{j}}\oint\nolimits_{\mathcal{Z}_j} 
\frac{\omega_j}{z_j}\sigma^2_{i,j,m,n}(\omega_i,\omega_j,\tilde{\omega}_m,\tilde{\omega}_n) dz_j =
\nonumber \\
= \sigma_i^2(\omega_i,\tilde{\omega}_m;\mathbf{I}_M,\mathbf{Q}_n(\tilde{\omega}_n))\delta_{i=m} 
\nonumber \\
+\sigma_i^2(\omega_i,\tilde{\omega}_n;\mathbf{I}_M,\mathbf{Q}_m(\tilde{\omega}_m))\delta_{i=n}.
\label{eq:integralOutOneVar}
\end{align}
Proceeding in a similar way, one can readily see that 
\begin{align*}
\frac{-1}{4\pi^2}\oint\nolimits_{\mathcal{Z}_i}\oint\nolimits_{\mathcal{Z}_j} \frac{\omega_i {\omega}_j}{z_i {z}_j}\sigma^2_{i,j,m,n}(\omega_i,\omega_j,\tilde{\omega}_m,\tilde{\omega}_n) dz_j d\tilde{z}_m =0 \\
\frac{-1}{4\pi^2}\oint\nolimits_{\mathcal{Z}_j}\oint\nolimits_{\mathcal{Z}_m} \frac{\omega_j \tilde{\omega}_m}{z_j \tilde{z}_m}\sigma^2_{i,j,m,n}(\omega_i,\omega_j,\tilde{\omega}_m,\tilde{\omega}_n) dz_j d\tilde{z}_m =
\\
=\sigma_i^2(\omega_i,\tilde{\omega}_n;\mathbf{I}_M,\mathbf{I}_M)\delta_{i=n} \\
\frac{-1}{4\pi^2}\oint\nolimits_{\mathcal{Z}_j}\oint\nolimits_{\mathcal{Z}_n} \frac{\omega_j \tilde{\omega}_n}{z_j \tilde{z}_n}\sigma^2_{i,j,m,n}(\omega_i,\omega_j,\tilde{\omega}_m,\tilde{\omega}_n) dz_j d\tilde{z}_n =
\\
=\sigma_i^2(\omega_i,\tilde{\omega}_m;\mathbf{I}_M,\mathbf{I}_M)\delta_{i=m} 
\end{align*}
Let us see the implication of the above results in the simplification of the expression in (\ref{eq:Sigma2}). To that effect, let us consider the effect of the second term in (\ref{eq:defgfunction}), namely $f_i^{(l)}(\omega_i)\phi_i(\omega_i,f_j^{(l)}(\mathbf{R}_j))$, on the integral in (\ref{eq:asymptVariance}). It follows from (\ref{eq:integralOutOneVar}) that 
\begin{multline*}
\frac{-1}{4\pi^2}\oint\nolimits_{\mathrm{C}_i^{(l)}} \oint\nolimits_{\mathcal{Z}_j} \frac{\omega_i \omega_j}{z_i z_j} f_i^{(l)}(\omega_i)\phi_i(\omega_i,f_j^{(l)}(\mathbf{R}_j) )
\times \\ \times 
\sigma^2_{i,j,m,n}(\omega_i,\omega_j,\tilde{\omega}_m,\tilde{\omega}_m) dz_i dz_j =
\\
= \frac{1}{2\pi\mathrm{j}}\oint\nolimits_{\mathrm{C}_i^{(l)}} \frac{\omega_i}{z_i} f_i^{(l)}(\omega_i) \phi_i(\omega_i,f_j^{(l)}(\mathbf{R}_j)) 
\times \\ \times 
\Biggl[
\sigma^2_{i}\left(\omega_i,\tilde{\omega}_m; \mathbf{I}_M, \mathbf{Q}_n(\tilde{\omega}_n) \right) \delta_{i=m}+
\\  
\sigma^2_{i}\left(\omega_i,\tilde{\omega}_n; \mathbf{I}_M, \mathbf{Q}_m(\tilde{\omega}_m) \right) \delta_{i=n}
\Biggl] dz_i 
=
\\
= \frac{-1}{4\pi^2}\oint\nolimits_{\mathrm{C}_{\omega_i}^{(l)}} \oint\nolimits_{\mathrm{C}_{\omega_j}^{(l)}} 
f_i^{(l)}(\omega_i)  f_j^{(l)}(\omega_j)
\times \\ \times
\Bigg[
\sigma^2_{i}\left(\omega_i,\tilde{\omega}_m;  \frac{\omega_i }{z_i \omega_i^{\prime}} \phi_i(\omega_i,\mathbf{Q}_j(\omega_j)) \mathbf{I}_M, \mathbf{Q}_n(\tilde{\omega}_n) \right) \delta_{i=m}+ \\
+ \sigma^2_{i}\left(\omega_i,\tilde{\omega}_n; \frac{\omega_i }{z_i \omega_i^{\prime}} \phi_i(\omega_i,\mathbf{Q}_j(\omega_j))  \mathbf{I}_M, \mathbf{Q}_m(\tilde{\omega}_m) \right) \delta_{i=n}
\Bigg] d\omega_i d\omega_j
\end{multline*}
where the last identity follows from the linearity of $\phi_i(\omega_i;\mathbf{A} )$ as a function of $\mathbf{A}$, the fact that 
$$
f_j^{(l)}(\mathbf{R}_j)=\frac{1}{2\pi\mathrm{j}} \oint\nolimits_{\mathrm{C}_{\omega_j}^{(l)}} f_j^{(l)}(\omega_j) \mathbf{Q}_j(\omega_j)d\omega_j
$$ 
and the change of variable $z_i \mapsto \omega_i=\omega_i(z_i)$. 
Proceeding similarly for the rest of the cross terms, one can easily transform the identity in (\ref{eq:asymptVariance}) into an expression of the form
\begin{align*} 
 \{\bar{\boldsymbol{\Sigma}}_M\}_{r,s} &=
\sum_{l_r,l_s=1}^{L} \frac{1+\varsigma}{\left(  2\pi\mathrm{j}\right)  ^{4}}
\oint\nolimits_{\mathrm{C}^{(l_r)}_{\omega_{i_r}}} 
\oint\nolimits_{\mathrm{C}^{(l_r)}_{\omega_{j_r}}} \oint\nolimits_{\mathrm{C}^{(l_s)}_{\omega_{i_s}}} \oint\nolimits_{\mathrm{C}^{(l_s)}_{\omega_{j_s}}}
 f_{i_r}^{(l_r)}(\omega_{i_r})
 \times \\ &\times
 f_{j_r}^{(l_r)}(\omega_{j_r}) 
 f_{i_s}^{(l_s)}(\tilde{\omega}_{i_s}) 
 f_{j_s}^{(l_s)}(\tilde{\omega}_{j_s})
 \times  \\ &\times
\bar{\sigma}_{i_r,j_r,i_s,j_s}
^{2}\left(  \omega_{i_r},\omega_{j_r},\tilde{\omega}_{i_s},\tilde{\omega}_{j_s}\right)  d\omega_{i_r}d\omega_{j_r} d\tilde{\omega}_{i_s} d\tilde{\omega}_{j_s} 
\end{align*}
where now the function $\bar{\sigma}_{i_r,j_r,i_s,j_s}
^{2}\left(  \omega_{i_r},\omega_{j_r},\tilde{\omega}_{i_s},\tilde{\omega}_{j_s}\right) $ is defined as in (\ref{eq:Sigma2}) but replacing the all functions ${\sigma}^2_j(\omega,\tilde{\omega};\mathbf{A},\mathbf{B})$ by the expression 
\[
{\sigma}^2_j\left(\omega,\tilde{\omega};\mathbf{A}-\frac{\omega \phi_j(\omega;\mathbf{A})}{z \omega^{\prime}} \mathbf{I}_M,\mathbf{B} - \frac{\tilde{\omega} \phi_j(\tilde{\omega};\mathbf{A})}{\tilde{z} \tilde{\omega}^{\prime}} \mathbf{I}_M \right).
\]
Now, using the identity in (\ref{eq:OmegaIdentity}) we immediately see that the above quantity is exactly equal to $\bar{\sigma}^2_j(\omega,\tilde{\omega};\mathbf{A},\mathbf{B})$ as defined in (\ref{eq:omegabivariate}). From all this, we can conclude that $ \left[\bar{\boldsymbol{\Sigma}}_M\right]_{r,s}$ takes the form of final expression in the statement of the theorem.

\section{Evaluating the asymptotic covariance in (\ref{eq:asympvar}) using only one integral}
\label{sec:simplifiedIntegralsVar}

The evaluation of the different terms in (\ref{eq:asympvar}) involves a four-fold integral that is difficult to evaluate numerically for general functions $f_j^{(l)}(z)$. In this appendix, we show that in fact the evaluation can be carried out with integrals with respect to a single variable, which of course are much easier to evaluate. We begin by noticing that $\bar{\sigma}_j^2(\omega,\tilde{\omega},\mathbf{A},\mathbf{B})$ is a linear function with respect to the matrices $\mathbf{A},\mathbf{B}$. 
Therefore, when integrating terms involving this function, two of the integrals can be moved inside its argument. Hence, a closer look at the expression in (\ref{eq:asympvar}) reveals that all the integrals can be expressed in the following two alternative ways:
\begin{align*}
I_{j}^{(l,l^{\prime})} &  =\frac{-1}{4\pi^{2}}\oint_{\mathrm{C}_{\omega_{j}%
}^{(l)}}\oint_{\mathrm{C}_{\omega_{j}}^{(l^{\prime})}}f_{j}^{(l)}\left(
\omega\right)  f_{j}^{(l^{\prime})}\left(  \tilde{\omega}\right)  
\times \\ &\times
\bar{\sigma
}_{j}^{2}\left(  \omega,\tilde{\omega};\mathbf{A},\mathbf{B}\right)  d\omega
d\tilde{\omega}\\
J_{i,j}^{\left( l,l^{\prime}\right)  } &  =\frac{1}{\left(  2\pi
\mathrm{j}\right)  ^{4}}\oint\nolimits_{\mathrm{C}_{\omega_{i}}^{(l)}}
\oint_{\mathrm{C}_{\omega_{j}}^{(l)}}\oint_{\mathrm{C}_{\omega_{i}
}^{(l^{\prime})}}\oint_{\mathrm{C}_{\omega_{j}}^{(l^{\prime})}}f_{i}
^{(l)}\left(  \omega_{i}\right) \times \\ &\times  f_{j}^{(l)}\left(  \omega_{j}\right)
f_{i}^{(l^{\prime})}\left(  \tilde{\omega}_{i}\right)  f_{j}^{(l^{\prime}
)}\left(  \tilde{\omega}_{j}\right) 
\times \\ 
&\times
\varrho_{i,j}\left(  \omega_{i}
,\omega_{j},\tilde{\omega}_{i},\tilde{\omega}_{j}\right)  d\omega_{i}
d\omega_{j}d\tilde{\omega}_{i}d\tilde{\omega}_{j}.
\end{align*}
Now, consider the eigendecomposition $\mathbf{R}_{j}=\sum_{k=1}^{\bar{M}_{j}
}\gamma_{k}^{(j)}\mathbf{\Pi}_{k}^{(j)}$, where $\mathbf{\Pi}_{k}^{(j)}$ denotes the orthogonal projection matrix onto the subspaces spanned by the $K_k^{(j)}$ eigenvectors associated with $\gamma_k^{(j)}$. We can express the above integrals as
\begin{align*}
I_{j}^{(l,l^{\prime})} &  =
\sum_{k=1}^{\bar{M}_{j}}\sum_{r=1}^{\bar{M}_{j}
}\Biggl(  \gamma_{k}^{(j)}\gamma_{r}^{(j)} \frac{1}{N_{j}}\mathrm{tr}\left[
\mathbf{\Pi}_{k}^{(j)}\mathbf{A\Pi}_{r}^{(j)}\mathbf{B}\right]  \mathcal{I}
_{j}^{(l,l^{\prime})}\left(  k,r\right)  +
\\
&\left(\gamma_{k}^{(j)}\gamma_{r}^{(j)}\right)^2 \frac{1}{N_{j}}\mathrm{tr}\left[
\mathbf{\Pi}_{k}^{(j)}\mathbf{A}\right]  \frac{1}{N_{j}}\mathrm{tr}\left[
\mathbf{\Pi}_{r}^{(j)}\mathbf{B}\right]  \widetilde{\mathcal{I}}
_{j}^{(l,l^{\prime})}\left(  k,r\right)  \Biggl) 
\\
J_{i,j}^{\left(  l,l^{\prime}\right)  } &  =\frac{1}{N_{i}N_{j}}\sum_{m=1}^{\bar{M}_{i}}\sum_{k=1}^{\bar{M}_{j}}\sum_{n=1}^{\bar{M}_{i}}\sum_{r=1}^{\bar{M}_{j}}\gamma_{m}^{(i)}\gamma_{k}^{(j)}\gamma_{n}^{(i)}\gamma_{r}^{(j)}\mathrm{tr}\left[  \mathbf{\Pi}_{m}^{(i)}\mathbf{\Pi}_{k}^{(j)}\right]  
\times \\
& \times \mathrm{tr}\left[  \mathbf{\Pi}_{n}^{(i)}\mathbf{\Pi}_{r}^{(j)}\right]  \mathcal{I}_{i}^{(l,l^{\prime})}\left(  m,n\right)
\mathcal{I}_{j}^{(l,l^{\prime})}\left(  k,r\right)
\end{align*}
where we have now defined
\begin{align*}
\mathcal{I}_{j}^{(l,l^{\prime})}\left(  k,r\right)    &=\frac{-1}{4\pi^{2}
}\oint_{\mathrm{C}_{\omega_{j}}^{(l)}}\oint_{\mathrm{C}_{\omega_{j}
}^{(l^{\prime})}}
\frac{1}{1-\Gamma_{j}\left(  \omega,\tilde{\omega}\right)
}\frac{1}{\left(  \gamma_{k}^{(j)}-\omega\right) }
\times \\ 
&\times \frac{f_{j}^{(l)}\left(  \omega\right)  f_{j}^{(l^{\prime})}\left(
\tilde{\omega}\right)  }{ \left(
\gamma_{r}^{(j)}-\omega\right)  \left(  \gamma_{k}^{(j)}-\tilde{\omega
}\right)  \left(  \gamma_{r}^{(j)}-\tilde{\omega}\right)  }d\omega
d\tilde{\omega}
\\
\widetilde{\mathcal{I}}_{j}^{(l,l^{\prime})}\left(  k,r\right)   &  =\frac
{-1}{4\pi^{2}}\oint_{\mathrm{C}_{\omega_{j}}^{(l)}}\oint_{\mathrm{C}
_{\omega_{j}}^{(l^{\prime})}}\frac{1}{\left(  1-\Gamma_{j}\left(
\omega,\tilde{\omega}\right)  \right)  ^{2}}
\frac{1}{\left(
\gamma_{k}^{(j)}-\omega\right)^{2}}
\times \\ 
&\times
\frac{f_{j}^{(l)}\left(\omega\right)  f_{j}^{(l^{\prime})}\left(  \tilde{\omega}\right)  }
{\left(  \gamma_{r}^{(j)}-\omega\right)
\left(  \gamma_{k}^{(j)}-\tilde{\omega}\right)  \left(  \gamma_{r}
^{(j)}-\tilde{\omega}\right)  ^{2}}d\omega d\tilde{\omega}
\end{align*}
We next show that these two double integrals can be transformed into a single one.

\subsection{Solving $\mathcal{I}_{j}^{(l,l^{\prime})}\left(  k,r\right)  $}
\subsubsection{Case $r\neq k$}

Let us denote by $\varphi_{m}^{(j)}\left(  \omega\right)  $, $m=1,\ldots
,\bar{M}_{j}$, the solutions to the equation $\Gamma_{j}\left(  \omega
,\varphi_{m}^{(j)}\left(  \omega\right)  \right)  =1$. {}A direct application
of the Rouché's theorem allows us to prove that there are exactly $\bar{M}_{j}$
solutions inside the contour $\mathrm{C}_{\omega_{j}}^{(l)}$. Now, it is clear
that
\begin{equation}
\left(  1-\Gamma_{j}\left(  \omega,\tilde{\omega}\right)  \right)
\prod\nolimits_{m=1}^{\bar{M}_{j}}\left(  \gamma_{m}^{(j)}-\tilde{\omega
}\right)  =\prod\nolimits_{m=1}^{\bar{M}_{j}}\left(  \varphi_{m}^{(j)}\left(
\omega\right)  -\tilde{\omega}\right)  \label{eq:polynomialOriginalIdent}
\end{equation}
since both sides are polynomials in $\tilde{\omega}$ of degree $\bar{M}_{j}$,
have the same roots and the same leading coefficient. Hence, we can readily
write, using partial fraction decomposition,
\begin{multline*}
\frac{1}{1-\Gamma_{j}\left(  \omega,\tilde{\omega}\right)  }\frac{1}{\left(
\gamma_{k}^{(j)}-\tilde{\omega}\right)  \left(  \gamma_{r}^{(j)}-\tilde
{\omega}\right)  } =
\\
=
\frac{\prod\nolimits_{\substack{m=1\\m\neq r,k}}^{\bar{M}_{j}}\left(  \gamma_{m}^{(j)}-\tilde{\omega}\right)  }{\prod\nolimits_{m=1}^{\bar{M}_{j}}\left(  \varphi_{m}^{(j)}\left(\omega\right)  -\tilde{\omega}\right)  } 
\\
=\sum_{m=1}^{\bar{M}_{j}}\frac{1}{\left(  \varphi_{m}^{(j)}\left(  \omega\right)  -\tilde{\omega}\right)}\frac{\prod\nolimits_{\substack{l=1\\l\neq r,k}}^{\bar{M}_{j}}\left(
\gamma_{l}^{(j)}-\varphi_{m}^{(j)}\left(  \omega\right)  \right)  }{\prod\nolimits_{\substack{l=1\\l\neq m}}^{\bar{M}_{j}}\left(  \varphi_{l}^{(j)}\left(  \omega\right)  -\varphi_{m}^{(j)}\left(  \omega\right)
\right)  }.
\end{multline*}
The above expression can be simplified by using the identity 
\begin{equation}
\frac{\prod\nolimits_{l=1}^{\bar{M}_{j}}\left(  \gamma_{l}^{(j)}-\varphi
_{m}^{(j)}\left(  \omega\right)  \right)  }{\prod
\nolimits_{\substack{l=1\\l\neq m}}^{\bar{M}_{j}}\left(  \varphi_{l}%
^{(j)}\left(  \omega\right)  -\varphi_{m}^{(j)}\left(  \omega\right)  \right)
}=\frac{\left(  \omega-\varphi_{m}^{(j)}\left(  \omega\right)  \right)
}{z^{\prime}(\varphi_{m}^{(j)}\left(  \omega\right)  )}
\label{eq:quotientDifPhis}
\end{equation}
where we have defined
\[
z^{\prime}(\omega)=1-\frac{1}{N_{j}}\sum_{r=1}^{\bar{M}_{j}}K_{r}^{(j)}
\frac{\left(  \gamma_{r}^{(j)}\right)  ^{2}}{\left(  \gamma_{r}^{(j)}
-\omega\right)  ^{2}}.
\]
This identity essentially follows from taking derivatives on both sides of
(\ref{eq:polynomialOriginalIdent}) and forcing $\tilde{\omega}=\varphi
_{m}^{(j)}\left(  \omega\right)  $. This yields 
\begin{multline*}
\frac{1}{1-\Gamma_{j}\left(  \omega,\tilde{\omega}\right)  }\frac{1}{\left(
\gamma_{k}^{(j)}-\tilde{\omega}\right)  \left(  \gamma_{r}^{(j)}-\tilde
{\omega}\right)  }  =
\\
=
\sum_{m=1}^{\bar{M}_{j}}\frac{1}{\left(  \varphi_{m}^{(j)}\left(
\omega\right)  -\tilde{\omega}\right)  }
\frac{\left(  \omega-\varphi_{m}^{(j)}\left(  \omega\right)  \right)  }{\left(  \gamma_{r}^{(j)}-\varphi_{m}^{(j)}\left(  \omega\right)  \right)  \left(  \gamma_{k}^{(j)}-\varphi_{m}^{(j)}\left(  \omega\right)  \right)  }
\times \\ \times
\frac{1}{z^{\prime}(\varphi_{m}^{(j)}\left(  \omega\right)  )}.
\end{multline*}
Hence, we can insert the above expression
into $\mathcal{I}_{j}^{(l,l^{\prime})}\left(  k,r\right)  $  and carry out the
integral with respect to $\tilde{\omega}$ by evaluating the residues at the simple poles $\varphi_m(\omega)$, $m=1,\ldots,\bar{M}_j$,
\begin{multline*}
\mathcal{I}_{j}^{(l,l^{\prime})}\left(  k,r\right)  =\sum_{m=1}^{\bar{M}_{j}%
}\frac{1}{2\pi\mathrm{j}}
\oint_{\mathrm{C}_{\omega_{j}}^{(l)}}
\frac{f_{j}^{(l)}\left(  \omega\right)  }{\left(  \gamma_{r}^{(j)}-\varphi_{m}^{(j)}\left(  \omega\right)  \right)}
\times \\ \times
\frac{  
f_{j}^{(l^{\prime})} \left(  \varphi_{m}^{(j)}\left(  \omega\right)  \right)
\left(  \omega-\varphi_{m}^{(j)}\left(  \omega\right)  \right)  
}{  
\left(  \gamma_{k}^{(j)}-\varphi_{m}^{(j)}\left(  \omega\right)  \right) 
\left(  \gamma_{k}^{(j)}-\omega\right)  
\left(  \gamma_{r}^{(j)}-\omega\right)  
}
\times \\ \times 
\frac{1}{z^{\prime
}(\varphi_{m}^{(j)}\left(  \omega\right)  )}d\omega
\end{multline*}
or, alternatively, using the conventional change of variable $\omega\mapsto
z=z(\omega)$,
\begin{multline*}
\mathcal{I}_{j}^{(l,l^{\prime})}\left(  k,r\right)  =\sum_{m=1}^{\bar{M}_{j}%
}\frac{1}{2\pi\mathrm{j}}
\oint_{\mathrm{C}_{_{j}}^{(l)}}
\frac{
f_{j}^{(l^{\prime})}\left(  \varphi_{m}^{(j)}\left(  \omega\right)  \right)
}
{\left(  \gamma_{r}^{(j)}-\varphi_{m}^{(j)}\left(  \omega\right)  \right) }
\times \\ \times
\frac{f_{j}^{(l)}\left(  \omega\right)}
{
\left(  \gamma_{k}^{(j)}-\varphi_{m}^{(j)}\left(  \omega\right)  \right)  
\left(  \gamma_{k}^{(j)}-\omega\right)  
\left(  \gamma_{r}^{(j)}-\omega\right)  
}
\times \\ \times
\frac{\left(\omega-\varphi_{m}^{(j)}\left(  \omega\right)  \right)  }{z^{\prime}(\varphi_{m}^{(j)}\left(  \omega\right)  )z^{\prime}(\omega)}dz
\end{multline*}
where all the appearances of $\omega$ above should be understood as
$\omega(z)$. 
\subsubsection{Case $k=r$}
The case $k=r$ can also be dealt with by using a partial fraction
decomposition and the identity in (\ref{eq:quotientDifPhis}), which directly yields
\begin{multline*}
\frac{1}{1-\Gamma_{j}\left(  \omega,\tilde{\omega}\right)  }\frac{1}{\left(
\gamma_{k}^{(j)}-\tilde{\omega}\right)^{2}}   =
\\
=
\frac{1}{\left(  \gamma
_{k}^{(j)}-\tilde{\omega}\right)  }\frac{\prod\nolimits_{\substack{m=1\\m\neq
k}}^{\bar{M}_{j}}\left(  \gamma_{m}^{(j)}-\tilde{\omega}\right)  }
{\prod\nolimits_{m=1}^{\bar{M}_{j}}\left(  \varphi_{m}^{(j)}\left(
\omega\right)  -\tilde{\omega}\right)  }
\\
=
\frac{1}{\left(  \gamma_{k}^{(j)}-\tilde{\omega}\right)  }\frac
{\prod\nolimits_{\substack{m=1\\m\neq k}}^{\bar{M}_{j}}\left(  \gamma
_{m}^{(j)}-\gamma_{k}^{(j)}\right)  }{\prod\nolimits_{m=1}^{\bar{M}_{j}%
}\left(  \varphi_{m}^{(j)}\left(  \omega\right)  -\gamma_{k}^{(j)}\right)  }
\\
  +\sum_{m=1}^{\bar{M}_{j}}\frac{1}{\left(  \varphi_{m}^{(j)}\left(
\omega\right)  -\tilde{\omega}\right)  \left(  \gamma_{k}^{(j)}-\varphi
_{m}^{(j)}\left(  \omega\right)  \right)  ^{2}}\frac{\left(  \omega
-\varphi_{m}^{(j)}\left(  \omega\right)  \right)  }{z^{\prime}(\varphi
_{m}^{(j)}\left(  \omega\right)  )}.
\end{multline*}
Hence, inserting the above into the expression of $\mathcal{I}_{j}
^{(l,l^{\prime})}\left(  k,k\right)  $ and integrating with respect to
$\tilde{\omega}$ we obtain
\begin{align*}
\mathcal{I}_{j}^{(l,l^{\prime})}\left(  k,k\right)   &  
=\frac{1}{2\pi\mathrm{j}}\oint_{\mathrm{C}_{\omega_{j}}^{(l)}}\frac{\prod\nolimits_{\substack{m=1\\m\neq k}}^{\bar{M}_{j}}\left(  \gamma_{m}^{(j)}-\gamma_{k}^{(j)}\right)  }{\prod\nolimits_{m=1}^{\bar{M}_{j}}\left(\varphi_{m}^{(j)}\left(  \omega\right)  -\gamma_{k}^{(j)}\right)  }
\times \\ & \times
\frac{f_{j}^{(l^{\prime})}\left(  \gamma_{k}^{(j)}\right)  f_{j}^{(l)}\left(\omega\right)  }{\left(  \gamma_{k}^{(j)}-\omega\right)  ^{2}}d\omega
\\
&  
+\sum_{m=1}^{\bar{M}_{j}}\frac{1}{2\pi\mathrm{j}}\oint_{\mathrm{C}_{\omega_{j}}^{(l)}}
\frac{\left(  \omega-\varphi_{m}^{(j)}\left(  \omega\right)\right)}{\left(  \gamma_{k}^{(j)}-\varphi_{m}^{(j)}\left(  \omega\right)  \right)  ^{2}}
\times \\ & \times
\frac{f_{j}^{(l^{\prime})}\left(  \varphi_{m}^{(j)}\left(\omega\right)  \right)  f_{j}^{(l)}\left(  \omega\right)  }{\left(\gamma_{k}^{(j)}-\omega\right)  ^{2}}\frac{1}{z^{\prime}(\varphi_{m}^{(j)}\left(  \omega\right)  )}d\omega .
\end{align*}
Now, using the identity
\begin{equation} \label{eq:eq:quotientGammas}
-\frac{1}{N_{j}}K_{k}^{(j)}\frac{\left(  \gamma_{k}^{(j)}\right)  ^{2}%
}{\left(  \gamma_{k}^{(j)}-\omega\right)  }=\frac{\prod\nolimits_{m=1}%
^{\bar{M}_{j}}\left(  \varphi_{m}^{(j)}\left(  \omega\right)  -\gamma
_{k}^{(j)}\right)  }{\prod\nolimits_{\substack{m=1\\m\neq k}}^{\bar{M}_{j}%
}\left(  \gamma_{m}^{(j)}-\gamma_{k}^{(j)}\right)  }
\end{equation}
(which follows from evaluating (\ref{eq:polynomialOriginalIdent}) at
$\tilde{\omega}=\gamma_{k}^{(j)}$) we obtain%
\begin{align*}
\mathcal{I}_{j}^{(l,l^{\prime})}&\left(  k,k\right)     =-\frac{N_{j}}%
{K_{k}^{(j)}}\frac{f_{j}^{(l^{\prime})}\left(  \gamma_{k}^{(j)}\right)
f_{j}^{(l)}\left(  \gamma_{k}^{(j)}\right)  }{\left(  \gamma_{k}^{(j)}\right)
^{2}}\\
&  +\sum_{m=1}^{\bar{M}_{j}}\frac{1}{2\pi\mathrm{j}}\oint_{\mathrm{C}_{j}^{(l)}}\frac{f_{j}^{(l^{\prime})}\left(  \varphi_{m}^{(j)}\left(\omega\right)  \right)  f_{j}^{(l)}\left(  \omega\right)  }{\left(  \gamma_{k}^{(j)}-\varphi_{m}^{(j)}\left(  \omega\right)  \right)  ^{2}\left(\gamma_{k}^{(j)}-\omega\right)  ^{2}}
\times \\ &\times 
\frac{\left(  \omega-\varphi_{m}^{(j)}\left(  \omega\right)  \right)  }{z^{\prime}(\varphi_{m}^{(j)}\left(
\omega\right)  )z^{\prime}(\omega)}dz
\end{align*}
with the same conventions as above.

\subsection{Solving $\widetilde{\mathcal{I}}_{j}^{(l,l^{\prime})}\left(
k,r\right)  $}
\subsubsection{Case $k\neq r$}
As before, we begin by identifying $1-\Gamma_j(\omega,\tilde{\omega})$ as a quotient of polynomials in $\tilde{\omega}$ according to (\ref{eq:polynomialOriginalIdent}), so that 
\begin{multline*}
\frac{1}{\left(1-\Gamma_j(\omega,\tilde{\omega})\right)^2}\frac{1}{
\left(\gamma_k^{(j)}-\tilde{\omega}\right)^2
\left(\gamma_r^{(j)}-\tilde{\omega}\right)
}
=
\\ =
\frac{\left(  \gamma_{k}^{(j)}-\tilde{\omega}\right)  \prod
\nolimits_{\substack{m=1\\m\neq k,r}}^{\bar{M}_{j}}\left(  \gamma_{m}%
^{(j)}-\tilde{\omega}\right)  ^{2}}{\prod\nolimits_{m=1}^{\bar{M}_{j}}\left(
\varphi_{m}^{(j)}\left(  \omega\right)  -\tilde{\omega}\right)  ^{2}}.
\end{multline*}
We can therefore expand this quotient using partial fraction decomposition, that is
\begin{multline*}
\frac{\left(  \gamma_{k}^{(j)}-\tilde{\omega}\right)  \prod
\nolimits_{\substack{m=1\\m\neq k,r}}^{\bar{M}_{j}}\left(  \gamma_{m}%
^{(j)}-\tilde{\omega}\right)  ^{2}}{\prod\nolimits_{m=1}^{\bar{M}_{j}}\left(
\varphi_{m}^{(j)}\left(  \omega\right)  -\tilde{\omega}\right)  ^{2}} 
=
\\
=
\sum_{m=1}^{\bar{M}_{j}}
\frac{\left(  \gamma_{k}^{(j)}-\varphi_{m}^{(j)}\left(  \omega\right)  \right)}{\left(  \varphi_{m}^{(j)}\left(
\omega\right)  -\tilde{\omega}\right)  ^{2}}
\frac{ 
\prod
\nolimits_{\substack{l=1\\l\neq k,r}}^{\bar{M}_{j}}\left(  \gamma_{l}^{(j)}-\varphi_{m}^{(j)}\left(  \omega\right)  \right)  ^{2}}{\prod
\nolimits_{\substack{l=1\\l\neq m}}^{\bar{M}_{j}}\left(  \varphi_{l}^{(j)}\left(  \omega\right)  -\varphi_{m}^{(j)}\left(  \omega\right)  \right)^{2}}
\\
-\sum_{m=1}^{\bar{M}_{j}}
\frac{d}{d\tilde{\omega}}\left[
\frac{\left(  \gamma_{k}^{(j)}-\tilde{\omega}\right)  \prod
\nolimits_{\substack{l=1\\l\neq k,r}}^{\bar{M}_{j}}\left(  \gamma_{l}%
^{(j)}-\tilde{\omega}\right)  ^{2}}{\prod\nolimits_{\substack{l=1\\l\neq
m}}^{\bar{M}_{j}}\left(  \varphi_{l}^{(j)}\left(  \omega\right)
-\tilde{\omega}\right)  ^{2}}\right]  _{\varphi_{m}^{(j)}\left(
\omega\right)  }
\times 
\\
\times 
\frac{1}{\left(  \varphi_{m}^{(j)}\left(
\omega\right)  -\tilde{\omega}\right)  }.
\end{multline*}

The corresponding expression has first and second order poles at $\tilde{\omega} = \varphi^{(j)}_m(\omega)$, so that one can easily evaluate the integral with respect to $\tilde{\omega}$, which after some manipulation can be expressed as 
\begin{align*}
\widetilde{\mathcal{I}}_{j}^{(l,l^{\prime})}\left(  k,r\right)  &=-\sum_{m=1}^{\bar{M}_{j}}\frac{1}{2\pi\mathrm{j}}\oint_{\mathrm{C}_{\omega_{j}}^{(l)}}
\frac{d}{d\tilde{\omega}}\Biggl[  \frac{\left(  \varphi_{m}^{(j)}\left(  \omega\right)  -\tilde{\omega}\right)  ^{2}}{\left(1-\Gamma_{j}\left(  \omega,\tilde{\omega}\right)  \right)  ^{2}}
\times \\ &\times 
\frac{f_{j}^{(l^{\prime})}\left(  \tilde{\omega}\right)  }{\left(  \gamma_{k}^{(j)}-\tilde{\omega}\right)  \left(  \gamma_{r}^{(j)}-\tilde{\omega}\right)^{2}}
\Biggl]_{\tilde{\omega}=\varphi_{m}^{(j)}\left(  \omega\right)  }
\times \\ &\times
\frac{f_{j}^{(l)}\left(  \omega\right)  }{\left(  \gamma_{k}^{(j)}-\omega\right)  ^{2}\left(  \gamma_{r}^{(j)}-\omega\right)  }d\omega.
\end{align*}
Note that, by directly evaluating the derivatives and using
(\ref{eq:quotientDifPhis}), we can write
\begin{multline*}
\frac{d}{d\tilde{\omega}}\left[  \frac{\varphi_{m}^{(j)}\left(  \omega\right)
-\tilde{\omega}}{1-\Gamma_{j}\left(  \omega,\tilde{\omega}\right)  }\right]_{\tilde{\omega}=\varphi_{m}^{(j)}\left(  \omega\right)  }
=\\ 
= \frac{\omega-\varphi_{m}^{(j)}\left(  \omega\right)  }{z^{\prime}(\varphi_{m}^{(j)}\left(  \omega\right)  )}\Biggl(  \sum_{\substack{i=1\\i\neq m}}^{\bar{M}_{j}}\frac{1}{\varphi_{i}^{(j)}\left(  \omega\right) 
-\varphi_{m}^{(j)}\left(  \omega\right)  }
\\
-\sum_{i=1}^{\bar{M}_{j}}\frac{1}{\gamma_{i}^{(j)}-\varphi_{m}^{(j)}\left(  \omega\right)  }\Biggl).
\end{multline*}
The last term can be simplified by taking second order derivatives on both
sides of (\ref{eq:polynomialOriginalIdent}) and then forcing $\tilde{\omega
}=\varphi_{m}^{(j)}\left(  \omega\right)  $, leading to%
\begin{multline*}
\sum_{\substack{i=1\\i\neq m}}^{\bar{M}_{j}}\frac{1}{\varphi_{i}^{(j)}\left(\omega\right)  -\varphi_{m}^{(j)}\left(  \omega\right)  }-\sum_{i=1}^{\bar{M}_{j}}\frac{1}{\gamma_{i}^{(j)}-\varphi_{m}^{(j)}\left(  \omega\right)}
=
\\
= 
-\frac{\omega-\varphi_{m}^{(j)}\left(  \omega\right)  }{z^{\prime}\left(\varphi_{m}^{(j)}\left(  \omega\right)  \right)  }
\times \\ 
\times 
\frac{1}{N_{j}}\sum_{r=1}^{\bar{M}_{j}}K_{r}^{(j)}\frac{\left(  \gamma_{r}^{(j)}\right)  ^{2}}{\left(  \gamma_{r}^{(j)}-\omega\right)  \left(  \gamma_{r}^{(j)}-\varphi_{m}^{(j)}\left(  \omega\right)  \right) ^{3}}
\end{multline*}
so that we can conclude that, after some manipulation,
\begin{align*}
\frac{d}{d\tilde{\omega}}&\left[  \frac{\varphi_{m}^{(j)}\left(  \omega\right) -\tilde{\omega}}{1-\Gamma_{j}\left(  \omega,\tilde{\omega}\right)  }\right]_{\tilde{\omega}=\varphi_{m}^{(j)}\left(  \omega\right)  }  
=
\\
&=\frac
{d}{d\tilde{\omega}}\left[  \frac{\prod\nolimits_{l=1}^{\bar{M}_{j}}\left(
\gamma_{l}^{(j)}-\tilde{\omega}\right)  }{\prod\nolimits_{\substack{l=1\\l\neq m}}^{\bar{M}_{j}}\left(  \varphi_{l}^{(j)}\left(  \omega\right)-\tilde{\omega}\right)  }\right]_{\tilde{\omega}=\varphi_{m}^{(j)}\left(
\omega\right)  } 
\\
& =\frac{1}{2}\left[  \frac{d}{d\tilde{\omega}}\left[  \frac{\left(
\omega-\tilde{\omega}\right)  }{z^{\prime}\left(  \tilde{\omega}\right)
}\right]  _{\tilde{\omega}=\varphi_{m}^{(j)}\left(  \omega\right)  }-\frac
{1}{z^{\prime}\left(  \varphi_{m}^{(j)}\left(  \omega\right)  \right)
}\right].
\end{align*}
Observe that the first term can be also rewritten as%
\begin{align*}
&\frac{d}{d\tilde{\omega}}\left[  \frac{\omega-\tilde{\omega}}{z^{\prime
}(\tilde{\omega})}\right]  _{\tilde{\omega}=\varphi_{m}^{(j)}\left(
\omega\right)  } =
\\
& =\frac{-1}{z^{\prime}(\varphi_{m}^{(j)}\left(
\omega\right)  )}+\left(  \omega-\varphi_{m}^{(j)}\left(  \omega\right)
\right)  \left[  \frac{d}{d\tilde{\omega}}\frac{1}{z^{\prime}(\tilde{\omega}%
)}\right]  _{\tilde{\omega}=\varphi_{m}^{(j)}\left(  \omega\right)  }\\
& =\left(  \frac{d\varphi_{m}^{(j)}\left(  \omega\right)  }{d\omega}\right)
^{-1}\left(  \frac{d}{d\omega}\left[  \frac{\omega-\varphi_{m}^{(j)}\left(
\omega\right)  }{z^{\prime}(\varphi_{m}^{(j)}\left(  \omega\right)  )}\right]
-\frac{1}{z^{\prime}(\varphi_{m}^{(j)}\left(  \omega\right)  )}\right)  \\
& =\frac{z^{\prime}\left(  \varphi_{m}^{(j)}\left(  \omega\right)  \right)
}{z^{\prime}(\omega)}\frac{d}{d\omega}\left[  \frac{\omega-\varphi_{m}%
^{(j)}\left(  \omega\right)  }{z^{\prime}(\varphi_{m}^{(j)}\left(
\omega\right)  )}\right]  -\frac{1}{z^{\prime}(\omega)}%
\end{align*}
where we have used the fact that
\[
\frac{d\varphi_{m}^{(j)}\left(  \omega\right)  }{d\omega}
=
\frac{z^{\prime
}(\omega)}{z^{\prime}\left(  \varphi_{m}^{(j)}\left(  \omega\right)  \right)
}
\]
We can therefore express, by virtue of (\ref{eq:quotientDifPhis}),
\begin{multline*}
\frac{d}{d\tilde{\omega}}\left[  \left(  \frac{\varphi_{m}^{(j)}\left(
\omega\right)  -\tilde{\omega}}{1-\Gamma_{j}\left(  \omega,\tilde{\omega
}\right)  }\right)  ^{2}\right]  _{\tilde{\omega}
=
\varphi_{m}^{(j)}\left(
\omega\right)  } 
=
\\
=
\frac{\omega-\varphi_{m}^{(j)}\left(  \omega\right)}{z^{\prime}(\omega)}\frac{d}{d\omega}\left[  \frac{\omega-\varphi_{m}^{(j)}\left(  \omega\right)  }{z^{\prime}(\varphi_{m}^{(j)}\left(\omega\right)  )}\right]  
\\
-\frac{\omega-\varphi_{m}^{(j)}\left(\omega\right)  }{z^{\prime}(\varphi_{m}^{(j)}\left(  \omega\right)  )}\frac{1}{z^{\prime}(\omega)}-\frac{\omega-\varphi_{m}^{(j)}\left(  \omega\right)}{\left[  z^{\prime}(\varphi_{m}^{(j)}\left(  \omega\right)  )\right] ^{2}}
\end{multline*}
which allows us to express the original integral, using the integration by parts formula, as
\begin{multline*}
\widetilde{\mathcal{I}}_{j}^{(l,l^{\prime})}\left(  k,r\right)  
=\sum_{m=1}^{\bar{M}_{j}}\frac{1}{2\pi\mathrm{j}}\oint_{\mathrm{C}_{\omega
_{j}}^{(l)}}\frac{\omega-\varphi_{m}^{(j)}\left(  \omega\right)  }
\times \\ \times 
{z^{\prime
}(\varphi_{m}^{(j)}\left(  \omega\right)  )}\frac{f_{j}^{(l^{\prime})}\left(
\varphi_{m}^{(j)}\left(  \omega\right)  \right)  }{\left(  \gamma_{k}%
^{(j)}-\varphi_{m}^{(j)}\left(  \omega\right)  \right)  \left(  \gamma
_{r}^{(j)}-\varphi_{m}^{(j)}\left(  \omega\right)  \right)  ^{2}}
\times \\
\times \frac{d}{d\omega}\left[  \frac{\omega-\varphi_{m}^{(j)}\left(  \omega\right)
}{z^{\prime}(\omega)}\frac{f_{j}^{(l)}\left(  \omega\right)  }{\left(
\gamma_{k}^{(j)}-\omega\right)  ^{2}\left(  \gamma_{r}^{(j)}-\omega\right)
}\right]  d\omega\\
+\sum_{m=1}^{\bar{M}_{j}}\frac{1}{2\pi\mathrm{j}}\oint_{\mathrm{C}%
_{\omega_{j}}^{(l)}}\frac{\omega-\varphi_{m}^{(j)}\left(  \omega\right)
}{z^{\prime}(\varphi_{m}^{(j)}\left(  \omega\right)  )}\left[  \frac
{1}{z^{\prime}(\omega)}+\frac{1}{z^{\prime}(\varphi_{m}^{(j)}\left(
\omega\right)  )}\right] \times \\ \times \frac{f_{j}^{(l^{\prime})}\left(  \varphi_{m}^{(j)}\left(  \omega\right)  \right)  }{\left(  \gamma_{k}^{(j)}-\varphi
_{m}^{(j)}\left(  \omega\right)  \right)  \left(  \gamma_{r}^{(j)}-\varphi
_{m}^{(j)}\left(  \omega\right)  \right)  ^{2}}
\times \\ \times 
\frac{f_{j}^{(l)}\left(
\omega\right)  }{\left(  \gamma_{k}^{(j)}-\omega\right)  ^{2}\left(
\gamma_{r}^{(j)}-\omega\right)  }d\omega.
\end{multline*}
The only remaining derivative can be easily evaluated for each specific choice
of $f_{j}^{(l)}\left(  \omega\right)  $, and the result is then easily
evaluated using a single numerical integration.

\subsection{Case $k=r$}
In this case, we observe that we can express
\begin{multline*}
\frac{1}{\left(  1-\Gamma_{j}\left(  \omega,\tilde{\omega}\right)  \right)
^{2}\left(  \gamma_{k}^{(j)}-\tilde{\omega}\right)  ^{3}}
= \\ =
\frac{\prod
\nolimits_{\substack{m=1\\m\neq k}}^{\bar{M}_{j}}\left(  \gamma_{m}%
^{(j)}-\tilde{\omega}\right)  ^{2}}{\left(  \gamma_{k}^{(j)}-\tilde{\omega
}\right)  \prod\nolimits_{m=1}^{\bar{M}_{j}}\left(  \varphi_{m}^{(j)}\left(
\omega\right)  -\tilde{\omega}\right)  ^{2}}%
\end{multline*}
so that we can decompose the original integral using again partial fraction
decomposition, taking into account the additional root at $\tilde{\omega
}=\gamma_{k}^{(j)}$. Using the identity in (\ref{eq:eq:quotientGammas}) we can write
\begin{multline*}
\frac{1}{\left(  1-\Gamma_{j}\left(  \omega,\tilde{\omega}\right)  \right)^{2}}
\frac{1}{\left(  \gamma_{k}^{(j)}-\tilde{\omega}\right)^{3}}
= \\ 
= 
\frac{1}{\left(\gamma_{k}^{(j)}-\tilde{\omega}\right)  }\left(  \frac{N_{j}}{K_{k}^{(j)}}\right)  ^{2}\frac{\left(  \gamma_{k}^{(j)}-\omega\right)  ^{2}}{\left(\gamma_{k}^{(j)}\right)  ^{4}}
\\
+\sum_{m=1}^{\bar{M}_{j}}\frac{1}{\left(  \varphi_{m}^{(j)}\left(\omega\right)  -\tilde{\omega}\right)  ^{2}}
\times \\ 
\times 
\left[  \frac{\prod\nolimits_{\substack{l=1\\l\neq k}}^{\bar{M}_{j}}\left(  \gamma_{l}^{(j)}-\tilde{\omega}\right)  ^{2}}{\left(  \gamma_{k}^{(j)}-\tilde{\omega}\right)  \prod\nolimits_{\substack{l=1\\l\neq m}}^{\bar{M}_{j}}\left(  \varphi_{l}^{(j)}\left(\omega\right)  -\tilde{\omega}\right)  ^{2}}\right]_{\tilde{\omega}=\varphi_{m}^{(j)}\left(  \omega\right)  }
\\
-\sum_{m=1}^{\bar{M}_{j}}\frac{1}{\left(  \varphi_{m}^{(j)}\left(
\omega\right)  -\tilde{\omega}\right)  }
\times \\ 
\times 
\frac{d}{d\tilde{\omega}}\left[\frac{\prod\nolimits_{\substack{l=1\\l\neq k}}^{\bar{M}_{j}}\left(  \gamma_{l}^{(j)}-\tilde{\omega}\right)  ^{2}}{\left(  \gamma_{k}^{(j)}-\tilde{\omega}\right)  \prod\nolimits_{\substack{l=1\\l\neq m}}^{\bar{M}_{j}}\left(  \varphi_{l}^{(j)}\left(  \omega\right)  -\tilde{\omega}\right)^{2}}\right]
_{\tilde{\omega}=\varphi_{m}^{(j)}\left(  \omega\right)  }.
\end{multline*}
We can therefore insert the above expression into the original integral and
carry out the integration with respect to these three types of poles, so that
\begin{multline*}
\widetilde{\mathcal{I}}_{j}^{(l,l^{\prime})}\left(  k,k\right)   = \left(
\frac{N_{j}}{K_{k}^{(j)}}\right)  ^{2}\frac{f_{j}^{(l)}\left(  \gamma
_{k}^{(j)}\right)  f_{j}^{(l^{\prime})}\left(  \gamma_{k}^{(j)}\right)
}{\left(  \gamma_{k}^{(j)}\right)  ^{4}}\\
 -\sum_{m=1}^{\bar{M}_{j}}\frac{1}{2\pi\mathrm{j}}\oint_{\mathrm{C}%
_{\omega_{j}}^{(l)}}\frac{d}{d\tilde{\omega}}\Biggl[  \frac{\prod
\nolimits_{\substack{l=1\\l\neq k}}^{\bar{M}_{j}}\left(  \gamma_{l}%
^{(j)}-\tilde{\omega}\right)  ^{2}}{\left(  \gamma_{k}^{(j)}-\tilde{\omega
}\right)  \prod\nolimits_{\substack{l=1\\l\neq m}}^{\bar{M}_{j}}\left(  \varphi_{l}^{(j)}\left(
\omega\right)  -\tilde{\omega}\right)  ^{2}}
\times \\ \times
f_{j}^{(l^{\prime})}\left(
\tilde{\omega}\right)  \Biggl]  _{\varphi_{m}^{(j)}\left(  \omega\right)
}\frac{f_{j}^{(l)}\left(  \omega\right)  }{\left(  \gamma_{k}^{(j)}%
-\omega\right)  ^{3}}d\omega
\end{multline*}
For the evaluation of the second term, we can use exactly the same approach as
in the case $k\neq r$ and the resulting expression turns out to be the same,
particularized to the case  $k=r$.


\bibliographystyle{IEEEtran}
\bibliography{./bib/IEEEbib}

\begin{IEEEbiographynophoto}
{Roberto Pereira}
received his M.Sc. degree in Informatics from the Technical University of Munich, Germany, in 2019 and his PhD. degree in signal theory and communications from Technical University of Catalonia (UPC), Barcelona, Spain, 2023.

He joined the Centre Tecnol\`ogic de Telecomunicacions de Catalunya (CTTC/CERCA), Barcelona, Spain, in 2019, under a Marie Curie Fellowship as an Early Stage Researcher (ESR) in the H2020 ITN WindMill Project. During his PhD he was also visited the Connectivity Section of the Aalborg University for the period of 4 months. He is a now a Research Associate in CTTC. His main research activities are in wireless communications, machine learning and optimization in large dimensional settings.
\end{IEEEbiographynophoto}

\begin{IEEEbiographynophoto}
{Xavier Mestre}
received the MS and PhD in Electrical Engineering from the Technical University of Catalonia (UPC) in 1997 and 2002 respectively and the licenciate (5 year) degree in Mathematics in 2011. From January 1998 to December 2002, he was with UPC’s Communications Signal Processing Group, where he worked as a Research Assistant (1998-2000) and Research Associate (2001-2002). In January 2003 he joined the Telecommunications Technological Center of Catalonia (CTTC), where he currently holds a position as a Senior Research Associate. He has participated in more than 15 EU-funded projects and a number of other industrial and national projects. Since 2017 he has been adjunct lecturer at the Engineering Department of the Universitat Autònoma de Barcelona.

He has been Associate Editor of the IEEE Transactions on Signal Processing (2008-11, 2015-2019) and Senior Area Editor of the same journal (2019-2024). He is currently vice-chair of the Signal Processing Theory and Methods Technical Committee (elected member since 2019). He has been elected member of the IEEE Sensor Array and Multi-channel Signal Processing Technical Committee (2013-2018), the EURASIP Technical Area Committee on Signal Processing in Communications (2018-present) and vice-chair of the EURASIP Technical Area Committee on Theoretical and Methodological Trends in Signal Processing (2015-2022).

He has participated in the organization of multiple conferences and scientific events. He was general chair of the IEEE International Conference on Acoustics, Speech and Signal Processing 2020 and will also be general chair of the 2026 edition of this conference. He has also been involved in the organization of Eusipco 2024 (special session chair), IEEE WCNC 2018 (general vice-chair), IEEE ISPLC (technical chair), European Wireless 2014 (general co-chair), Eusipco 2011 (general technical chair).

\end{IEEEbiographynophoto}

\begin{IEEEbiographynophoto}
{David Gregoratti}
(S'02--M'10--SM'15) received his M.Sc.  degree in telecommunications engineering from ``Politecnico di Torino,'' Italy, in 2005, his Ph.D.\ degree in signal theory and communications from ``Universitat Polit\'ecnica de Catalunya'' (UPC), Barcelona, Spain, in 2010, and a Master in Advanced Mathematics from ``Universitat de Barcelona'' (UB), Barcelona, Spain, in 2021.

He was with the ``Centre Tecnol\`ogic de Telecomunicacions de Catalunya'' (CTTC/CERCA), Barcelona, Spain, first as a Ph.D.\ Candidate (2006--2010) and later as a Research Associate (2010--2020), contributing to several projects and contracts with both private and public funding. In 2021, he joined Software Radio Systems Spain SL, Barcelona, Spain. His current research interests are mainly focused on efficient digital signal processing algorithms for software defined radio 5G applications.

Dr.\ Gregoratti has volunteered in a number of IEEE-sponsored activities and
events, including ICASSP 2020 and WCNC 2018.
\end{IEEEbiographynophoto}

\end{document}

%% file: figs/final_double_column/histograms/top_hist_eu_M_40_N1_400_N2_80_08_04.tex
\begin{tikzpicture}

\definecolor{darkgray176}{RGB}{176,176,176}
\definecolor{lightgray204}{RGB}{204,204,204}
\definecolor{orange}{RGB}{255,165,0}
\definecolor{steelblue31119180}{RGB}{31,119,180}
\definecolor{mycolor1}{rgb}{0.00000,0.44700,0.74100}%
\definecolor{mycolor2}{RGB}{255,165,0}

\begin{axis}[%
width=0.3\linewidth,
height=1.0in,
at={(0in,2.1in)},
scale only axis,
scale only axis,
align =center,
title = {{$M=40$}},
scale only axis,
xmin=1, xmax=3,
ymin=0., ymax=3,
xtick={1, 2.0, 3},
ytick={0, 1.5, 3},
axis background/.style={fill=white}
]
\addplot[ybar interval, fill=mycolor1, fill opacity=0.6, draw=black, area legend] table[row sep=crcr] {%
x	y\\
0.96	0.00158730158730159\\
1.023	0.00158730158730159\\
1.086	0.00793650793650792\\
1.149	0.0222222222222222\\
1.212	0.0476190476190477\\
1.275	0.117460317460317\\
1.338	0.211111111111111\\
1.401	0.377777777777778\\
1.464	0.555555555555556\\
1.527	0.71904761904762\\
1.59	0.990476190476188\\
1.653	1.12063492063492\\
1.716	1.35555555555556\\
1.779	1.49841269841269\\
1.842	1.48412698412699\\
1.905	1.41111111111111\\
1.968	1.28571428571429\\
2.031	1.17460317460317\\
2.094	0.98253968253968\\
2.157	0.752380952380956\\
2.22	0.57142857142857\\
2.283	0.403174603174602\\
2.346	0.284126984126985\\
2.409	0.179365079365079\\
2.472	0.111111111111111\\
2.535	0.10952380952381\\
2.598	0.0428571428571427\\
2.661	0.0222222222222222\\
2.724	0.0111111111111112\\
2.787	0.00158730158730158\\
2.85	0.0095238095238095\\
2.913	0.00476190476190478\\
2.976	0\\
3.039	0.00158730158730159\\
3.102	0.00158730158730158\\
3.165	0.00158730158730159\\
3.228	0.00158730158730159\\
};
\addplot [color=mycolor2, forget plot, ultra thick]
  table[row sep=crcr]{%
0.822162869802932	0.000506385779279083\\
0.843519243028043	0.000697335147934404\\
0.864875616253154	0.000954038038276089\\
0.886231989478264	0.00129674298233387\\
0.907588362703375	0.00175108086601639\\
0.928944735928485	0.00234921399400998\\
0.950301109153596	0.00313114396631693\\
0.971657482378706	0.00414617447953503\\
0.993013855603817	0.00545451565279987\\
1.01437022882893	0.00712900453760461\\
1.03572660205404	0.0092569020939043\\
1.05708297527915	0.0119417102726444\\
1.07843934850426	0.0153049343314664\\
1.09979572172937	0.019487695782721\\
1.12115209495448	0.0246520813781848\\
1.14250846817959	0.0309820945207972\\
1.1638648414047	0.0386840589957107\\
1.18522121462981	0.0479863127119143\\
1.20657758785492	0.059138023192742\\
1.22793396108003	0.0724069588593092\\
1.24929033430514	0.0880760626403825\\
1.27064670753025	0.10643869878042\\
1.29200308075537	0.127792481114402\\
1.31335945398048	0.152431642085897\\
1.33471582720559	0.180637966105599\\
1.3560722004307	0.212670387170356\\
1.37742857365581	0.248753436552165\\
1.39878494688092	0.289064818224343\\
1.42014132010603	0.333722482809054\\
1.44149769333114	0.382771659542368\\
1.46285406655625	0.436172383706553\\
1.48421043978136	0.49378811750037\\
1.50556681300647	0.555376098871715\\
1.52692318623158	0.620580059547467\\
1.54827955945669	0.688925925692962\\
1.5696359326818	0.759821049379885\\
1.59099230590691	0.832557415585859\\
1.61234867913202	0.906319129563817\\
1.63370505235713	0.980194317570474\\
1.65506142558224	1.05319137726031\\
1.67641779880736	1.12425930207742\\
1.69777417203247	1.19231158819736\\
1.71913054525758	1.25625302573771\\
1.74048691848269	1.31500849125494\\
1.7618432917078	1.36755270865892\\
1.78319966493291	1.41293984181277\\
1.80455603815802	1.45033173307242\\
1.82591241138313	1.47902361351537\\
1.84726878460824	1.49846618454139\\
1.86862515783335	1.50828310479148\\
1.88998153105846	1.50828310479148\\
1.91133790428357	1.49846618454139\\
1.93269427750868	1.47902361351537\\
1.95405065073379	1.45033173307242\\
1.9754070239589	1.41293984181277\\
1.99676339718401	1.36755270865892\\
2.01811977040912	1.31500849125495\\
2.03947614363424	1.25625302573771\\
2.06083251685935	1.19231158819736\\
2.08218889008446	1.12425930207742\\
2.10354526330957	1.05319137726031\\
2.12490163653468	0.980194317570474\\
2.14625800975979	0.906319129563817\\
2.1676143829849	0.832557415585861\\
2.18897075621001	0.759821049379884\\
2.21032712943512	0.688925925692962\\
2.23168350266023	0.620580059547468\\
2.25303987588534	0.555376098871715\\
2.27439624911045	0.49378811750037\\
2.29575262233556	0.436172383706553\\
2.31710899556067	0.382771659542368\\
2.33846536878578	0.333722482809054\\
2.35982174201089	0.289064818224343\\
2.381178115236	0.248753436552165\\
2.40253448846112	0.212670387170356\\
2.42389086168623	0.180637966105599\\
2.44524723491134	0.152431642085898\\
2.46660360813645	0.127792481114402\\
2.48795998136156	0.10643869878042\\
2.50931635458667	0.0880760626403825\\
2.53067272781178	0.0724069588593094\\
2.55202910103689	0.059138023192742\\
2.573385474262	0.0479863127119144\\
2.59474184748711	0.0386840589957107\\
2.61609822071222	0.0309820945207972\\
2.63745459393733	0.0246520813781849\\
2.65881096716244	0.0194876957827211\\
2.68016734038755	0.0153049343314663\\
2.70152371361266	0.0119417102726444\\
2.72288008683777	0.0092569020939043\\
2.74423646006288	0.00712900453760459\\
2.76559283328799	0.00545451565279987\\
2.78694920651311	0.00414617447953502\\
2.80830557973822	0.00313114396631693\\
2.82966195296333	0.00234921399400998\\
2.85101832618844	0.00175108086601639\\
2.87237469941355	0.00129674298233387\\
2.89373107263866	0.000954038038276085\\
2.91508744586377	0.000697335147934406\\
2.93644381908888	0.000506385779279083\\
};
\end{axis}
\end{tikzpicture}%

%% file: figs/final_double_column/histograms/bot_hist_eu_M_80_N1_800_N2_160_08_04.tex
\begin{tikzpicture}
\definecolor{darkgray176}{RGB}{176,176,176}
\definecolor{lightgray204}{RGB}{204,204,204}
\definecolor{orange}{RGB}{255,165,0}
\definecolor{steelblue31119180}{RGB}{31,119,180}
\definecolor{mycolor1}{rgb}{0.00000,0.44700,0.74100}%
\definecolor{mycolor2}{RGB}{255,165,0}
\begin{axis}[%
width=0.3\linewidth,
height=1.0in,
at={(0in,2.1in)},
scale only axis,
scale only axis,
align =center,
title = {{$M=80$}},
scale only axis,
xmin=1, xmax=3,
ymin=0, ymax=3,
xtick={1., 2.0, 3},
ytick={0, 1.5, 3.0},
axis background/.style={fill=white}
]
\addplot[ybar interval, fill=mycolor1, fill opacity=0.6, draw=black, area legend] table[row sep=crcr] {%
x	y\\
1.41	0.00277777777777778\\
1.446	0.00277777777777778\\
1.482	0.00555555555555555\\
1.518	0.0166666666666668\\
1.554	0.0222222222222222\\
1.59	0.0583333333333333\\
1.626	0.155555555555555\\
1.662	0.275\\
1.698	0.547222222222222\\
1.734	0.763888888888888\\
1.77	1.19722222222222\\
1.806	1.70555555555555\\
1.842	2.15\\
1.878	2.47500000000001\\
1.914	2.76111111111111\\
1.95	2.95833333333333\\
1.986	2.71388888888887\\
2.022	2.45277777777781\\
2.058	2.28333333333333\\
2.094	1.71666666666667\\
2.13	1.27222222222222\\
2.166	0.841666666666666\\
2.202	0.563888888888888\\
2.238	0.377777777777777\\
2.274	0.188888888888889\\
2.31	0.144444444444444\\
2.346	0.0638888888888888\\
2.382	0.0444444444444444\\
2.418	0.013888888888889\\
2.454	0\\
2.49	0.00277777777777781\\
2.526	0.00277777777777781\\
};
\addplot [color=mycolor2, forget plot, ultra thick]
  table[row sep=crcr]{%
1.41251895863048	0.000965635482391707\\
1.42371838522093	0.00132976001601581\\
1.43491781181137	0.00181927103605142\\
1.44611723840181	0.00247278080570646\\
1.45731666499226	0.00333916528850769\\
1.4685160915827	0.00447975531930757\\
1.47971551817314	0.00597083061585338\\
1.49091494476358	0.0079064092189276\\
1.50211437135403	0.0104013068082263\\
1.51331379794447	0.0135944175712102\\
1.52451322453491	0.0176521408867875\\
1.53571265112535	0.0227718463502739\\
1.5469120777158	0.0291852343625823\\
1.55811150430624	0.0371614118876738\\
1.56931093089668	0.0470094648539953\\
1.58051035748712	0.0590802724173798\\
1.59170978407757	0.073767276842514\\
1.60290921066801	0.0915058994937351\\
1.61410863725845	0.112771282074141\\
1.6253080638489	0.138074036648828\\
1.63650749043934	0.167953711804444\\
1.64770691702978	0.20296972870822\\
1.65890634362022	0.243689612142384\\
1.67010577021067	0.29067443885752\\
1.68130519680111	0.344461548242853\\
1.69250462339155	0.405544705852604\\
1.70370404998199	0.474352074111584\\
1.71490347657244	0.55122251968829\\
1.72610290316288	0.636380964589992\\
1.73730232975332	0.72991365720079\\
1.74850175634376	0.831744388134351\\
1.75970118293421	0.941612791181828\\
1.77090060952465	1.05905593933997\\
1.78210003611509	1.18339445870088\\
1.79329946270553	1.31372432996788\\
1.80449888929598	1.44891542292887\\
1.81569831588642	1.58761761193725\\
1.82689774247686	1.72827505370141\\
1.83809716906731	1.86914888098215\\
1.84929659565775	2.00834819073986\\
1.86049602224819	2.14386880105606\\
1.87169544883863	2.27363883967928\\
1.88289487542908	2.39556983263092\\
1.89409430201952	2.50761160909759\\
1.90529372860996	2.60780905301478\\
1.9164931552004	2.69435853368895\\
1.92769258179085	2.76566175433914\\
1.93889200838129	2.8203747793607\\
1.95009143497173	2.85745014209772\\
1.96129086156217	2.87617019094031\\
1.97249028815262	2.87617019094031\\
1.98368971474306	2.85745014209772\\
1.9948891413335	2.8203747793607\\
2.00608856792394	2.76566175433914\\
2.01728799451439	2.69435853368895\\
2.02848742110483	2.60780905301478\\
2.03968684769527	2.50761160909759\\
2.05088627428572	2.39556983263092\\
2.06208570087616	2.27363883967928\\
2.0732851274666	2.14386880105606\\
2.08448455405704	2.00834819073986\\
2.09568398064749	1.86914888098215\\
2.10688340723793	1.7282750537014\\
2.11808283382837	1.58761761193725\\
2.12928226041881	1.44891542292887\\
2.14048168700926	1.31372432996787\\
2.1516811135997	1.18339445870088\\
2.16288054019014	1.05905593933997\\
2.17407996678058	0.941612791181826\\
2.18527939337103	0.831744388134351\\
2.19647881996147	0.729913657200788\\
2.20767824655191	0.636380964589992\\
2.21887767314236	0.55122251968829\\
2.2300770997328	0.474352074111586\\
2.24127652632324	0.405544705852604\\
2.25247595291368	0.344461548242853\\
2.26367537950413	0.290674438857519\\
2.27487480609457	0.243689612142384\\
2.28607423268501	0.202969728708221\\
2.29727365927545	0.167953711804444\\
2.3084730858659	0.138074036648828\\
2.31967251245634	0.112771282074142\\
2.33087193904678	0.0915058994937351\\
2.34207136563722	0.073767276842514\\
2.35327079222767	0.0590802724173802\\
2.36447021881811	0.0470094648539953\\
2.37566964540855	0.0371614118876734\\
2.386869071999	0.0291852343625822\\
2.39806849858944	0.0227718463502739\\
2.40926792517988	0.0176521408867876\\
2.42046735177032	0.0135944175712102\\
2.43166677836077	0.0104013068082263\\
2.44286620495121	0.00790640921892756\\
2.45406563154165	0.00597083061585338\\
2.46526505813209	0.00447975531930757\\
2.47646448472254	0.00333916528850771\\
2.48766391131298	0.00247278080570646\\
2.49886333790342	0.00181927103605143\\
2.51006276449386	0.0013297600160158\\
2.52126219108431	0.000965635482391707\\
};
\end{axis}
\end{tikzpicture}%

%% file: figs/final_double_column/histograms/top_hist_kl_M_40_N1_400_N2_80_08_04.tex
\definecolor{mycolor1}{rgb}{0.00000,0.44700,0.74100}%
\definecolor{mycolor2}{RGB}{255,165,0}
\definecolor{mycolor1}{rgb}{0.00000,0.44700,0.74100}%
\definecolor{mycolor2}{RGB}{255,165,0}
\begin{tikzpicture}
\begin{axis}[%
width=0.3\linewidth,
height=1.0in,
at={(0in,2.1in)},
scale only axis,
scale only axis,
align =center,
scale only axis,
xmin=0.5, xmax=0.935,
ymin=0, ymax=20,
xtick={0.5, 0.7, 0.9},
ytick={0, 10, 20},
axis background/.style={fill=white}
]
\addplot[ybar interval, fill=mycolor1, fill opacity=0.6, draw=black, area legend] table[row sep=crcr] {%
x	y\\
0.553	0.0641025641025639\\
0.5608	0.0897435897435907\\
0.5686	0.192307692307692\\
0.5764	0.358974358974358\\
0.5842	0.615384615384613\\
0.592	0.846153846153855\\
0.5998	1.29487179487179\\
0.6076	1.82051282051281\\
0.6154	2.32051282051281\\
0.6232	3.17948717948721\\
0.631	3.99999999999998\\
0.6388	4.76923076923075\\
0.6466	5.87179487179485\\
0.6544	6.98717948717956\\
0.6622	7.74358974358972\\
0.67	8.05128205128202\\
0.6778	8.82051282051291\\
0.6856	8.67948717948715\\
0.6934	8.76923076923074\\
0.7012	8.41025641025638\\
0.709	7.97435897435894\\
0.7168	7.0256410256411\\
0.7246	6.28205128205126\\
0.7324	5.29487179487178\\
0.7402	4.47435897435902\\
0.748	3.7179487179487\\
0.7558	2.76923076923076\\
0.7636	2.21794871794871\\
0.7714	1.57692307692307\\
0.7792	1.32051282051283\\
0.787	0.820512820512817\\
0.7948	0.717948717948715\\
0.8026	0.371794871794876\\
0.8104	0.269230769230768\\
0.8182	0.179487179487179\\
0.826	0.115384615384615\\
0.8338	0.0897435897435894\\
0.8416	0.0256410256410259\\
0.8494	0.0128205128205128\\
0.8572	0.0512820512820518\\
0.865	0\\
0.8728	0.0128205128205128\\
0.8806	0.0128205128205128\\
};
\addplot [color=mycolor2, forget plot, ultra thick]
  table[row sep=crcr]{%
0.515069581578184	0.00301527681052119\\
0.51865617225551	0.00415228583970393\\
0.522242762932836	0.00568082456277585\\
0.525829353610162	0.00772146297118361\\
0.529415944287488	0.0104268203111143\\
0.533002534964814	0.0139884071965343\\
0.53658912564214	0.0186444133669786\\
0.540175716319466	0.0246884179455335\\
0.543762306996792	0.0324789424061768\\
0.547348897674118	0.0424496953586298\\
0.550935488351444	0.0551202726521108\\
0.55452207902877	0.0711069772028938\\
0.558108669706096	0.0911333127125324\\
0.561695260383422	0.116039587975368\\
0.565281851060748	0.14679094941518\\
0.568868441738074	0.184483046271426\\
0.5724550324154	0.23034443461793\\
0.576041623092726	0.285734753745739\\
0.579628213770052	0.352137672995086\\
0.583214804447378	0.431147620850766\\
0.586801395124704	0.524449382483927\\
0.59038798580203	0.633789796845324\\
0.593974576479356	0.760941007095028\\
0.597561167156682	0.907655021879992\\
0.601147757834008	1.07560972796936\\
0.604734348511334	1.26634694922174\\
0.60832093918866	1.481203658288\\
0.611907529865986	1.72123799442049\\
0.615494120543312	1.98715229522494\\
0.619080711220638	2.27921587842759\\
0.622667301897964	2.59719077390459\\
0.62625389257529	2.94026396659375\\
0.629840483252616	3.30698992856726\\
0.633427073929942	3.69524725850182\\
0.637013664607268	4.10221308123264\\
0.640600255284594	4.52435847152487\\
0.64418684596192	4.9574675501698\\
0.647773436639246	5.39668206756535\\
0.651360027316572	5.83657226666711\\
0.654946617993898	6.27123364604538\\
0.658533208671224	6.6944079815845\\
0.66211979934855	7.09962568049492\\
0.665706390025876	7.48036530971847\\
0.669292980703202	7.83022504100421\\
0.672879571380528	8.14309986243135\\
0.676466162057854	8.4133577877026\\
0.68005275273518	8.63600800267757\\
0.683639343412506	8.80685395706645\\
0.687225934089832	8.92262484943852\\
0.690812524767158	8.9810797531741\\
0.694399115444484	8.9810797531741\\
0.69798570612181	8.92262484943852\\
0.701572296799136	8.80685395706645\\
0.705158887476462	8.63600800267757\\
0.708745478153788	8.4133577877026\\
0.712332068831114	8.14309986243135\\
0.71591865950844	7.8302250410042\\
0.719505250185765	7.48036530971847\\
0.723091840863092	7.09962568049492\\
0.726678431540417	6.6944079815845\\
0.730265022217743	6.27123364604538\\
0.73385161289507	5.83657226666709\\
0.737438203572395	5.39668206756535\\
0.741024794249721	4.9574675501698\\
0.744611384927047	4.52435847152487\\
0.748197975604373	4.10221308123264\\
0.751784566281699	3.69524725850182\\
0.755371156959025	3.30698992856726\\
0.758957747636351	2.94026396659375\\
0.762544338313677	2.59719077390459\\
0.766130928991003	2.27921587842759\\
0.769717519668329	1.98715229522494\\
0.773304110345655	1.72123799442049\\
0.776890701022981	1.481203658288\\
0.780477291700307	1.26634694922174\\
0.784063882377633	1.07560972796936\\
0.787650473054959	0.907655021879992\\
0.791237063732285	0.760941007095028\\
0.794823654409611	0.633789796845324\\
0.798410245086937	0.524449382483927\\
0.801996835764263	0.431147620850763\\
0.805583426441589	0.352137672995086\\
0.809170017118915	0.285734753745741\\
0.812756607796241	0.23034443461793\\
0.816343198473567	0.184483046271426\\
0.819929789150893	0.146790949415179\\
0.823516379828219	0.116039587975368\\
0.827102970505545	0.091133312712533\\
0.830689561182871	0.0711069772028938\\
0.834276151860197	0.0551202726521108\\
0.837862742537523	0.0424496953586298\\
0.841449333214849	0.0324789424061768\\
0.845035923892175	0.0246884179455337\\
0.848622514569501	0.0186444133669786\\
0.852209105246827	0.0139884071965343\\
0.855795695924153	0.0104268203111143\\
0.859382286601479	0.00772146297118361\\
0.862968877278805	0.00568082456277585\\
0.866555467956131	0.00415228583970393\\
0.870142058633457	0.00301527681052119\\
};
\end{axis}
\end{tikzpicture}%

%% file: figs/final_double_column/histograms/bot_hist_kl_M_80_N1_800_N2_160_08_04.tex
%
%
\definecolor{mycolor1}{rgb}{0.00000,0.44700,0.74100}%
\definecolor{mycolor2}{RGB}{255,165,0}
\definecolor{mycolor1}{rgb}{0.00000,0.44700,0.74100}%
\definecolor{mycolor2}{RGB}{255,165,0}
\begin{tikzpicture}
\begin{axis}[%
width=0.3\linewidth,
height=1.0in,
at={(0in,2.1in)},
scale only axis,
scale only axis,
align =center,
scale only axis,
xmin=0.5, xmax=0.935,
ymin=0, ymax=20,
xtick={0.5, 0.7, 0.9},
ytick={0, 10, 20},
axis background/.style={fill=white}
]
\addplot[ybar interval, fill=mycolor1, fill opacity=0.6, draw=black, area legend] table[row sep=crcr] {%
x	y\\
0.605	0.0163934426229508\\
0.6111	0.0327868852459017\\
0.6172	0.0327868852459017\\
0.6233	0.098360655737705\\
0.6294	0.213114754098361\\
0.6355	0.704918032786886\\
0.6416	1.31147540983607\\
0.6477	2.59016393442623\\
0.6538	4.32786885245902\\
0.6599	6.29508196721301\\
0.666	9.39344262295083\\
0.6721	12.0819672131148\\
0.6782	15.1967213114754\\
0.6843	16.2950819672131\\
0.6904	18.1803278688525\\
0.6965	16.7868852459017\\
0.7026	16.016393442623\\
0.7087	13.2786885245902\\
0.7148	10.344262295082\\
0.7209	7.26229508196722\\
0.727	5.31147540983607\\
0.7331	3.37704918032787\\
0.7392	2.42622950819672\\
0.7453	1.11475409836066\\
0.7514	0.622950819672132\\
0.7575	0.229508196721308\\
0.7636	0.229508196721312\\
0.7697	0.0655737704918033\\
0.7758	0.0491803278688525\\
0.7819	0.0327868852459017\\
0.788	0.0163934426229508\\
0.7941	0.0163934426229508\\
};
\addplot [color=mycolor2, forget plot, ultra thick]
  table[row sep=crcr]{%
0.606200922848981	0.00599503733136181\\
0.60800484217432	0.00825566280769012\\
0.609808761499659	0.0112947359286962\\
0.611612680824998	0.0153519765095704\\
0.613416600150337	0.0207308253737831\\
0.615220519475676	0.0278120479874011\\
0.617024438801015	0.0370692182443633\\
0.618828358126354	0.0490860363862098\\
0.620632277451693	0.0645753224144363\\
0.622436196777032	0.0843993849891132\\
0.624240116102371	0.109591295602187\\
0.62604403542771	0.14137640078821\\
0.627847954753049	0.181193185957566\\
0.629651874078389	0.230712370884427\\
0.631455793403727	0.291852880166558\\
0.633259712729067	0.366793106868812\\
0.635063632054406	0.457975692244067\\
0.636867551379745	0.56810389997895\\
0.638671470705084	0.700127592935495\\
0.640475390030423	0.857216847656982\\
0.642279309355762	1.04272138976764\\
0.644083228681101	1.26011432153295\\
0.64588714800644	1.51291905558423\\
0.647691067331779	1.80461897268663\\
0.649494986657118	2.13855008291515\\
0.651298905982457	2.51777787317916\\
0.653102825307796	2.94496054087032\\
0.654906744633135	3.42220190090128\\
0.656710663958474	3.95089835580159\\
0.658514583283813	4.5315853687888\\
0.660318502609152	5.16378980261358\\
0.662122421934491	5.84589520347907\\
0.66392634125983	6.57502754208869\\
0.665730260585169	7.34696900332067\\
0.667534179910508	8.15610708687789\\
0.669338099235847	8.99542550873341\\
0.671142018561186	9.85654216839401\\
0.672945937886525	10.7297977909209\\
0.674749857211864	11.6043968181522\\
0.676553776537203	12.4685997950668\\
0.678357695862543	13.3099639876939\\
0.680161615187882	14.1156264143805\\
0.681965534513221	14.8726210235519\\
0.68376945383856	15.5682195644486\\
0.685573373163899	16.1902839228365\\
0.687377292489238	16.7276164640627\\
0.689181211814577	17.1702943455604\\
0.690985131139916	17.5099738970031\\
0.692789050465255	17.740152041588\\
0.694592969790594	17.8563733214629\\
0.696396889115933	17.8563733214629\\
0.698200808441272	17.740152041588\\
0.700004727766611	17.5099738970031\\
0.70180864709195	17.1702943455604\\
0.703612566417289	16.7276164640627\\
0.705416485742628	16.1902839228365\\
0.707220405067967	15.5682195644486\\
0.709024324393306	14.8726210235519\\
0.710828243718645	14.1156264143805\\
0.712632163043984	13.3099639876939\\
0.714436082369323	12.4685997950668\\
0.716240001694662	11.6043968181522\\
0.718043921020001	10.7297977909209\\
0.71984784034534	9.85654216839401\\
0.72165175967068	8.99542550873341\\
0.723455678996018	8.15610708687789\\
0.725259598321358	7.34696900332067\\
0.727063517646697	6.57502754208869\\
0.728867436972036	5.84589520347907\\
0.730671356297375	5.16378980261358\\
0.732475275622714	4.5315853687888\\
0.734279194948053	3.95089835580159\\
0.736083114273392	3.42220190090128\\
0.737887033598731	2.94496054087032\\
0.73969095292407	2.51777787317916\\
0.741494872249409	2.13855008291515\\
0.743298791574748	1.80461897268663\\
0.745102710900087	1.51291905558423\\
0.746906630225426	1.26011432153295\\
0.748710549550765	1.04272138976764\\
0.750514468876104	0.857216847656982\\
0.752318388201443	0.700127592935495\\
0.754122307526782	0.56810389997895\\
0.755926226852121	0.457975692244067\\
0.75773014617746	0.366793106868812\\
0.759534065502799	0.291852880166558\\
0.761337984828138	0.230712370884427\\
0.763141904153477	0.181193185957566\\
0.764945823478816	0.14137640078821\\
0.766749742804155	0.109591295602187\\
0.768553662129494	0.0843993849891132\\
0.770357581454834	0.0645753224144363\\
0.772161500780173	0.0490860363862098\\
0.773965420105512	0.0370692182443633\\
0.775769339430851	0.0278120479874011\\
0.77757325875619	0.0207308253737831\\
0.779377178081529	0.0153519765095704\\
0.781181097406868	0.0112947359286962\\
0.782985016732207	0.00825566280769012\\
0.784788936057546	0.00599503733136181\\
};
\end{axis}

\end{tikzpicture}%

%% file: figs/final_double_column/histograms/top_hist_le_M_40_N1_400_N2_80_08_04.tex
%
%
\definecolor{mycolor1}{rgb}{0.00000,0.44700,0.74100}%
\definecolor{mycolor2}{RGB}{255,165,0}
\definecolor{mycolor1}{rgb}{0.00000,0.44700,0.74100}%
\definecolor{mycolor2}{RGB}{255,165,0}
\begin{tikzpicture}
\begin{axis}[%
width=0.3\linewidth,
height=1.0in,
at={(0in,2.1in)},
scale only axis,
scale only axis,
align =center,
scale only axis,
xmin=0.9, xmax=1.55,
ymin=0, ymax=15,
xtick={0.9, 1.2, 1.5},
ytick={0, 7.5, 15},
axis background/.style={fill=white}
]
\addplot[ybar interval, fill=mycolor1, fill opacity=0.6, draw=black, area legend] table[row sep=crcr] {%
x	y\\
1.02	0.160982452912632\\
1.042	0.333463652461881\\
1.064	0.586436078467446\\
1.086	1.20736839684474\\
1.108	2.35724306050643\\
1.13	3.23114780488923\\
1.152	4.62249614791992\\
1.174	5.79536830485477\\
1.196	6.0943357174068\\
1.218	5.64588459857875\\
1.24	5.0594485201113\\
1.262	3.94407009635949\\
1.284	2.58721799323873\\
1.306	1.74780948876572\\
1.328	1.05788469056873\\
1.35	0.597934825104069\\
1.372	0.275969919278796\\
1.394	0.149483706276017\\
1.416	0.149483706276017\\
};
\addplot [color=mycolor2, forget plot,  ultra thick]
  table[row sep=crcr]{%
0.951555186785156	0.00186095552116521\\
0.956846481327376	0.0025729548140711\\
0.962137775869596	0.00353392212531819\\
0.967429070411816	0.00482181385424789\\
0.972720364954036	0.00653570671883101\\
0.978011659496256	0.00880041839937667\\
0.983302954038476	0.0117717962019538\\
0.988594248580696	0.0156426669556496\\
0.993885543122916	0.0206494035971869\\
0.999176837665136	0.0270790158344764\\
1.00446813220736	0.0352766137017587\\
1.00975942674958	0.0456530242322478\\
1.0150507212918	0.05869226431105\\
1.02034201583402	0.074958489563931\\
1.02563331037624	0.0951019536554561\\
1.03092460491846	0.119863429686452\\
1.03621589946068	0.150076471843326\\
1.0415071940029	0.186666838531\\
1.04679848854512	0.230648366215621\\
1.05208978308734	0.283114584790195\\
1.05738107762956	0.345225408910262\\
1.06267237217178	0.418188332920259\\
1.067963666714	0.503233705396747\\
1.07325496125622	0.601583865980151\\
1.07854625579844	0.71441619151752\\
1.08383755034066	0.842820415775188\\
1.08912884488288	0.987750947461873\\
1.0944201394251	1.14997530033627\\
1.09971143396732	1.33002014717086\\
1.10500272850954	1.52811689242443\\
1.11029402305176	1.74414899955362\\
1.11558531759398	1.97760357922568\\
1.1208766121362	2.22752991583536\\
1.12616790667842	2.49250765578352\\
1.13145920122064	2.77062728097911\\
1.13675049576286	3.05948523127124\\
1.14204179030508	3.35619561559334\\
1.1473330848473	3.65741986993154\\
1.15262437938952	3.95941499890807\\
1.15791567393174	4.25810020652266\\
1.16320696847396	4.54914082077646\\
1.16849826301618	4.82804749552702\\
1.1737895575584	5.09028778577389\\
1.17908085210062	5.33140639661246\\
1.18437214664284	5.54714975636719\\
1.18966344118506	5.73359010983502\\
1.19495473572728	5.88724410689286\\
1.2002460302695	6.00518090009739\\
1.20553732481172	6.08511507134277\\
1.21082861935394	6.12548027364321\\
1.21611991389616	6.12548027364321\\
1.22141120843838	6.08511507134277\\
1.2267025029806	6.00518090009739\\
1.23199379752282	5.88724410689286\\
1.23728509206504	5.73359010983502\\
1.24257638660726	5.54714975636719\\
1.24786768114948	5.33140639661246\\
1.2531589756917	5.09028778577389\\
1.25845027023392	4.82804749552702\\
1.26374156477614	4.54914082077646\\
1.26903285931836	4.25810020652266\\
1.27432415386058	3.95941499890807\\
1.2796154484028	3.65741986993154\\
1.28490674294502	3.35619561559334\\
1.29019803748724	3.05948523127124\\
1.29548933202946	2.77062728097911\\
1.30078062657168	2.49250765578352\\
1.3060719211139	2.22752991583536\\
1.31136321565612	1.97760357922568\\
1.31665451019834	1.74414899955362\\
1.32194580474056	1.52811689242443\\
1.32723709928278	1.33002014717086\\
1.332528393825	1.14997530033627\\
1.33781968836722	0.987750947461873\\
1.34311098290944	0.842820415775188\\
1.34840227745166	0.71441619151752\\
1.35369357199388	0.601583865980151\\
1.3589848665361	0.503233705396743\\
1.36427616107832	0.418188332920259\\
1.36956745562054	0.345225408910265\\
1.37485875016276	0.283114584790195\\
1.38015004470498	0.230648366215621\\
1.3854413392472	0.186666838531\\
1.39073263378942	0.150076471843326\\
1.39602392833164	0.119863429686452\\
1.40131522287386	0.0951019536554561\\
1.40660651741608	0.074958489563931\\
1.4118978119583	0.05869226431105\\
1.41718910650052	0.0456530242322478\\
1.42248040104274	0.0352766137017587\\
1.42777169558496	0.0270790158344765\\
1.43306299012718	0.0206494035971869\\
1.4383542846694	0.0156426669556495\\
1.44364557921162	0.0117717962019537\\
1.44893687375384	0.00880041839937667\\
1.45422816829606	0.00653570671883105\\
1.45951946283828	0.00482181385424789\\
1.4648107573805	0.00353392212531819\\
1.47010205192272	0.00257295481407108\\
1.47539334646494	0.00186095552116521\\
};
\end{axis}
\end{tikzpicture}%

%% file: figs/final_double_column/histograms/bot_hist_le_M_80_N1_800_N2_160_08_04.tex
%
%
\definecolor{mycolor1}{rgb}{0.00000,0.44700,0.74100}%
\definecolor{mycolor2}{RGB}{255,165,0}
\definecolor{mycolor1}{rgb}{0.00000,0.44700,0.74100}%
\definecolor{mycolor2}{RGB}{255,165,0}
\begin{tikzpicture}
\begin{axis}[%
width=0.3\linewidth,
height=1.0in,
at={(0in,2.1in)},
scale only axis,
scale only axis,
align =center,
scale only axis,
xmin=0.9, xmax=1.55,
ymin=0, ymax=15,
xtick={0.9, 1.2, 1.5},
ytick={0, 7.5, 15},
axis background/.style={fill=white}
]
\addplot[ybar interval, fill=mycolor1, fill opacity=0.6, draw=black, area legend] table[row sep=crcr] {%
x	y\\
1.09	0.0611194641045381\\
1.1011	0\\
1.1122	0.122238928209076\\
1.1233	0.183358392313618\\
1.1344	0.244477856418152\\
1.1455	0.855672497463551\\
1.1566	1.83358392313614\\
1.1677	3.7282873103769\\
1.1788	6.35642426687197\\
1.1899	8.00664979769465\\
1.201	10.5125478259806\\
1.2121	12.1627733568033\\
1.2232	12.2238928209076\\
1.2343	10.8181451465035\\
1.2454	8.25112765411265\\
1.2565	6.90649944381295\\
1.2676	4.03388463089952\\
1.2787	1.83358392313618\\
1.2898	1.10015035388169\\
1.3009	0.427836248731775\\
1.312	0.183358392313614\\
1.3231	0.244477856418157\\
1.3342	0.244477856418157\\
};
\addplot [color=mycolor2, forget plot, ultra thick]
  table[row sep=crcr]{%
1.095018311921	0.0049555828245349\\
1.0976276824017	0.00677236886274191\\
1.10023705288239	0.00919640706823024\\
1.10284642336309	0.0124087343132236\\
1.10545579384378	0.0166367504472608\\
1.10806516432447	0.022163647563298\\
1.11067453480517	0.0293390255945735\\
1.11328390528586	0.0385906286550459\\
1.11589327576656	0.0504370570403613\\
1.11850264624725	0.0655012125160651\\
1.12111201672795	0.0845241200153836\\
1.12372138720864	0.108378639168908\\
1.12633075768934	0.138082437945231\\
1.12894012817003	0.174809453818928\\
1.13154949865073	0.219898923135248\\
1.13415886913142	0.274860926654165\\
1.13676823961212	0.341377290502495\\
1.13937761009281	0.421296610342518\\
1.1419869805735	0.516622146770271\\
1.1445963510542	0.629491386087364\\
1.14720572153489	0.762146185846689\\
1.14981509201559	0.91689263976573\\
1.15242446249628	1.0960501087499\\
1.15503383297698	1.3018892756958\\
1.15764320345767	1.53655958678877\\
1.16025257393837	1.80200702899517\\
1.16286194441906	2.09988384207715\\
1.16547131489976	2.43145244522818\\
1.16808068538045	2.79748653724175\\
1.17069005586115	3.19817296262079\\
1.17329942634184	3.6330184776655\\
1.17590879682253	4.10076595239979\\
1.17851816730323	4.59932476023259\\
1.18112753778392	5.12572009720667\\
1.18373690826462	5.67606570577143\\
1.18634627874531	6.24556393654039\\
1.18895564922601	6.82853626390178\\
1.1915650197067	7.41848629432482\\
1.1941743901874	8.00819600560727\\
1.19678376066809	8.58985448565837\\
1.19939313114879	9.15521687183995\\
1.20200250162948	9.69578961059712\\
1.20461187211017	10.203036654429\\
1.20722124259087	10.668599883658\\
1.20983061307156	11.0845259739641\\
1.21243998355226	11.4434912060304\\
1.21504935403295	11.7390153921232\\
1.21765872451365	11.9656562143488\\
1.22026809499434	12.1191758423184\\
1.22287746547504	12.1966727069849\\
1.22548683595573	12.1966727069849\\
1.22809620643643	12.1191758423184\\
1.23070557691712	11.9656562143488\\
1.23331494739781	11.7390153921232\\
1.23592431787851	11.4434912060304\\
1.2385336883592	11.0845259739641\\
1.2411430588399	10.668599883658\\
1.24375242932059	10.203036654429\\
1.24636179980129	9.69578961059712\\
1.24897117028198	9.15521687183995\\
1.25158054076268	8.58985448565837\\
1.25418991124337	8.00819600560727\\
1.25679928172407	7.41848629432482\\
1.25940865220476	6.82853626390178\\
1.26201802268546	6.24556393654039\\
1.26462739316615	5.67606570577143\\
1.26723676364684	5.12572009720667\\
1.26984613412754	4.59932476023259\\
1.27245550460823	4.10076595239979\\
1.27506487508893	3.6330184776655\\
1.27767424556962	3.19817296262079\\
1.28028361605032	2.79748653724175\\
1.28289298653101	2.43145244522818\\
1.28550235701171	2.09988384207715\\
1.2881117274924	1.80200702899517\\
1.2907210979731	1.53655958678877\\
1.29333046845379	1.3018892756958\\
1.29593983893448	1.0960501087499\\
1.29854920941518	0.91689263976573\\
1.30115857989587	0.762146185846677\\
1.30376795037657	0.629491386087364\\
1.30637732085726	0.516622146770271\\
1.30898669133796	0.421296610342518\\
1.31159606181865	0.341377290502495\\
1.31420543229935	0.274860926654165\\
1.31681480278004	0.219898923135248\\
1.31942417326074	0.174809453818928\\
1.32203354374143	0.138082437945231\\
1.32464291422212	0.10837863916891\\
1.32725228470282	0.0845241200153836\\
1.32986165518351	0.0655012125160651\\
1.33247102566421	0.0504370570403613\\
1.3350803961449	0.0385906286550459\\
1.3376897666256	0.0293390255945735\\
1.34029913710629	0.022163647563298\\
1.34290850758699	0.0166367504472608\\
1.34551787806768	0.0124087343132236\\
1.34812724854838	0.00919640706823024\\
1.35073661902907	0.00677236886274191\\
1.35334598950977	0.0049555828245349\\
};
\end{axis}

\end{tikzpicture}%

%% file: figs/final_double_column/icassp/icassp_left.tex
%


\definecolor{mycolor1}{RGB}{164,66,0}
\definecolor{mycolor2}{RGB}{27, 153, 139} 
\definecolor{mycolor3}{RGB}{255,186,8}

\begin{tikzpicture}

\begin{axis}[%
width=0.75\linewidth,
height=1.in,
at={(0in,1.7in)},
scale only axis,
xmin=4,
xmax=80,
xlabel style={font=\color{white!15!black}},
ymode=log,
ymin=1e-3,
ymax=1e2,
yminorticks=true,
axis background/.style={fill=white},
legend style={at={(1.35in,1.in)},anchor=north,
    legend columns=3, font=\small},
xlabel style={text width=\textwidth, align=center},
xlabel={
Growing $N$ \\ (a) $c = 1/3$ and $\rho_1 \neq \rho_2$},
ylabel={Normalized MSE},
ytick={100, 10, 1, 0.1, 0.01, 0.001}
]

\addplot [color=mycolor1, dashed, line width=1.5pt]
  table[row sep=crcr]{%
4 22.0531825403572\\
8 10.0063880999464\\
12 8.00707824669646\\
16 7.06463355304826\\
20 6.70698336096864\\
24 6.32179181489975\\
28 6.20404470539219\\
32 6.03199177696995\\
40 5.84723458168331\\
48 5.72752602618718\\
56 5.64088281611225\\
64 5.59819804006593\\
72 5.55711595993819\\
80 5.5122086390736\\
};
\addlegendentry{LE (TRAD)}

\addplot [color=mycolor2, dashed, line width=1.5pt]
  table[row sep=crcr]{%
4 54.0697477013531\\
8 21.3878965382415\\
12 16.6760210058229\\
16 14.5220064074477\\
20 13.7398006518458\\
24 12.9124237432465\\
28 12.6526658275354\\
32 12.2826163316711\\
40 11.8766766090565\\
48 11.6016883895621\\
56 11.4406144314719\\
64 11.3477013241997\\
72 11.2641224407674\\
80 11.169781087246\\
};
\addlegendentry{KL (TRAD)}

\addplot [color=mycolor3, dashed, line width=1.5pt]
  table[row sep=crcr]{%
4 34.5834776200675\\
8 8.08780740322365\\
12 5.32545791493304\\
16 3.90372258754803\\
20 3.51654222723525\\
24 3.33047328712539\\
28 3.19643958024283\\
32 2.98451569460762\\
40 2.80322558090831\\
48 2.68993329159016\\
56 2.56629402785797\\
64 2.53399306332127\\
72 2.50080173633129\\
80 2.48553779711568\\
};
\addlegendentry{EU (TRAD)}

\addplot [color=mycolor1, line width=1.5pt]
  table[row sep=crcr]{%
4 5.84338106474569\\
8 0.972004374438551\\
12 0.397757810559794\\
16 0.198610470202216\\
20 0.132070870417101\\
24 0.0815747931593199\\
28 0.0647431653410693\\
32 0.0486366907479247\\
40 0.0295614814106786\\
48 0.0195999924362036\\
56 0.0140968764680147\\
64 0.0116077374958384\\
72 0.0092701328300272\\
80 0.0071080244454003\\
};
\addlegendentry{LE (PROP)}

\addplot [color=mycolor2, line width=1.5pt]
  table[row sep=crcr]{%
4 9.85359394243842\\
8 1.43285049203657\\
12 0.562261163456425\\
16 0.2785759641503\\
20 0.184412258234564\\
24 0.112260194623464\\
28 0.0888592583402862\\
32 0.0662011418429323\\
40 0.0398982389984758\\
48 0.026107688576185\\
56 0.0190839977841018\\
64 0.0157429586763576\\
72 0.0126136904676363\\
80 0.00954041772373593\\
};
\addlegendentry{KL (PROP)}

\addplot [color=mycolor3, line width=1.5pt]
  table[row sep=crcr]{%
4 14.0682657348788\\
8 2.13072538726328\\
12 0.968789889366881\\
16 0.444747423699632\\
20 0.272855218661045\\
24 0.219578764180637\\
28 0.162780951609694\\
32 0.118823320339624\\
40 0.0747008852727229\\
48 0.0527550244545615\\
56 0.036035587953284\\
64 0.0263951191535045\\
72 0.0219666697379293\\
80 0.0175506768788776\\
};
\addlegendentry{EU (PROP)}

\end{axis}


\begin{axis}[%
width=0.75\linewidth,
height=1.in,
at={(0in,0in)},
scale only axis,
xmin=4,
xmax=80,
xlabel style={font=\color{white!15!black}},
ymode=log,
ymin=0.0001,
ymax=10,
yminorticks=true,
axis background/.style={fill=white},
ylabel={MSE},
title style={font=\bfseries},
xlabel style={text width=\textwidth, align=center},
xlabel={Growing $N$ \\ (c) $c = 1/3$ and $\rho_1 = \rho_2$},
ytick={10, 1, 0.1, 0.01, 0.001, 0.0001}
]

\addplot [color=mycolor1, dashed, line width=1.5pt]
  table[row sep=crcr]{%
4 1.28144382781095\\
8 0.801369531836439\\
12 0.706028379146016\\
16 0.660410904716977\\
20 0.641183194712889\\
24 0.619262167507512\\
28 0.609178586362755\\
32 0.5998644504104\\
40 0.58692417203017\\
48 0.582032281666876\\
56 0.578626553384435\\
64 0.574024257857787\\
72 0.571262324537235\\
80 0.569237671181249\\
};

\addplot [color=mycolor1, line width=1.5pt]
  table[row sep=crcr]{%
4 0.276569289863618\\
8 0.0558363183548939\\
12 0.0233958302395401\\
16 0.0127768082045006\\
20 0.00856649430675689\\
24 0.00555307562763639\\
28 0.00412578231871091\\
32 0.00306388935918001\\
40 0.00187898550729448\\
48 0.00136347147770375\\
56 0.000971729540490136\\
64 0.000784495450222966\\
72 0.000601436675243671\\
80 0.000493676757494008\\
};

\addplot [color=mycolor2, dashed, line width=1.5pt]
  table[row sep=crcr]{%
4 0.708221465717377\\
8 0.384611969954738\\
12 0.331067744139953\\
16 0.307291087241193\\
20 0.296755245382806\\
24 0.285179875945366\\
28 0.280248588005225\\
32 0.275167711880635\\
40 0.268695619360602\\
48 0.26592687704918\\
56 0.264612126065344\\
64 0.262099979896965\\
72 0.260865394121295\\
80 0.259897544620848\\
};

\addplot [color=mycolor2, line width=1.5pt]
  table[row sep=crcr]{%
4 0.108360044289842\\
8 0.018873658878946\\
12 0.00761666534481442\\
16 0.00423415100663355\\
20 0.00279965228284662\\
24 0.00179741685995095\\
28 0.00133419526537582\\
32 0.000971250727684459\\
40 0.000601079856256231\\
48 0.000423706649650817\\
56 0.000311315746736184\\
64 0.000248768644344777\\
72 0.000189690411396198\\
80 0.000154182837346761\\
};

\addplot [color=mycolor3, dashed, line width=1.5pt]
  table[row sep=crcr]{%
4 1.8371700998529\\
8 0.809511417696475\\
12 0.693471380167245\\
16 0.593047975509328\\
20 0.558223483429914\\
24 0.549396261208662\\
28 0.528178700575557\\
32 0.50777605189945\\
40 0.497875754558159\\
48 0.491974771684385\\
56 0.479162071813936\\
64 0.475847907873733\\
72 0.469585161537523\\
80 0.468021576796178\\
};

\addplot [color=mycolor3, line width=1.5pt]
  table[row sep=crcr]{%
4 0.739753090666081\\
8 0.156968246374703\\
12 0.0827754718488103\\
16 0.0387641635397611\\
20 0.0258743999971622\\
24 0.0198049636640436\\
28 0.0133238577024983\\
32 0.00967908970074142\\
40 0.00707917279179438\\
48 0.00513851845853683\\
56 0.00311790424009016\\
64 0.00242556493862796\\
72 0.00191675219112341\\
80 0.00150232492623233\\
};

\end{axis}

\end{tikzpicture}%

%% file: figs/final_double_column/icassp/icassp_right.tex
%


\definecolor{mycolor1}{RGB}{164,66,0}
\definecolor{mycolor2}{RGB}{27, 153, 139} 
\definecolor{mycolor3}{RGB}{255,186,8}
\begin{tikzpicture}

\begin{axis}[%
width=0.75\textwidth,
height=1.0in,
at={(0,1.7in)},
scale only axis,
xmin=4,
xmax=80,
xlabel style={font=\color{white!15!black}},
xlabel style={text width=\linewidth, align=center},
xlabel={Growing $N$ \\ (b) $c = 3/8$ and $\rho_1 \neq \rho_2$ },
ymode=log,
ymin=1e-3,
ymax=1e2,
yminorticks=true,
axis background/.style={fill=white},
title style={font=\bfseries},
ytick={100, 10, 1, 0.1, 0.01, 0.001}
]

\addplot [color=mycolor1, dashed, line width=1.5pt]
  table[row sep=crcr]{%
4 28.751202571078\\
8 12.5466973328863\\
12 10.5407775006092\\
16 9.49794349737744\\
20 8.82004011534334\\
24 8.55073049225466\\
28 8.30867803246589\\
32 8.04815399310414\\
40 7.81639819204837\\
48 7.80018310134461\\
56 7.54153747981358\\
64 7.57921759336775\\
72 7.48374443819728\\
80 7.38707586986541\\
};

\addplot [color=mycolor1, line width=1.5pt]
  table[row sep=crcr]{%
4 7.49721827534185\\
8 1.20809567440562\\
12 0.481317956920199\\
16 0.274883989063116\\
20 0.162631065989593\\
24 0.103070090953272\\
28 0.084226270872424\\
32 0.0603835878236734\\
40 0.0357331320629898\\
48 0.0277761571187664\\
56 0.0187003096795027\\
64 0.0151712651022601\\
72 0.0107137866387587\\
80 0.0089017449097175\\
};

\addplot [color=mycolor2, dashed, line width=1.5pt]
  table[row sep=crcr]{%
4 79.4978608302166\\
8 29.202497370727\\
12 23.7003159406506\\
16 21.0205095872767\\
20 19.2898299697531\\
24 18.605434649745\\
28 18.1038548818065\\
32 17.4219631664153\\
40 16.8920543992458\\
48 16.9001914933384\\
56 16.2863528902182\\
64 16.4040469329423\\
72 16.1622290456345\\
80 15.920103615255\\
};

\addplot [color=mycolor2, line width=1.5pt]
  table[row sep=crcr]{%
4 13.9136108319826\\
8 2.02512249992942\\
12 0.723452875837275\\
16 0.406725039698846\\
20 0.233131916800517\\
24 0.142143325205252\\
28 0.121075554762093\\
32 0.083967282098086\\
40 0.050060655031280\\
48 0.038489778813934\\
56 0.026423911866817\\
64 0.021273152351684\\
72 0.014546225458089\\
80 0.012027363044946\\
};

\addplot [color=mycolor3, dashed, line width=1.5pt]
  table[row sep=crcr]{%
4 44.5974491981763\\
8 9.69816362007871\\
12 6.33663526533726\\
16 5.06323156631602\\
20 4.56177449259415\\
24 4.19659806589398\\
28 3.87868496163333\\
32 3.65955044845865\\
40 3.50490286789355\\
48 3.44547900842038\\
56 3.232739549509\\
64 3.21037180961416\\
72 3.14369399703393\\
80 3.1079006682084\\
};

\addplot [color=mycolor3, line width=1.5pt]
  table[row sep=crcr]{%
4 18.6426379119171\\
8 2.53866363120297\\
12 1.01245629953334\\
16 0.545930076508487\\
20 0.390711107498871\\
24 0.247612705410133\\
28 0.176870352078256\\
32 0.134873975099294\\
40 0.085796025112040\\
48 0.060805629560959\\
56 0.042064503895839\\
64 0.032563806239785\\
72 0.023763591443557\\
80 0.020173525298515\\
};

\end{axis}


\begin{axis}[%
width=0.75\linewidth,
height=1.in,
at={(0in,0in)},
scale only axis,
xmin=4,
xmax=80,
xlabel style={font=\color{white!15!black}},
xlabel style={text width=\linewidth, align=center},
xlabel={Growing $N$ \\ (d) $c=3/8$ and $\rho_1 = \rho_2$},
ymode=log,
ymin=1e-4,
ymax=1e1,
yminorticks=true,
axis background/.style={fill=white},
title style={font=\bfseries},
ytick={10, 1, 0.1, 0.01, 0.001, 0.0001}
]

\addplot [color=mycolor1, dashed, line width=1.5pt]
  table[row sep=crcr]{%
4 1.70384181336775\\
8 1.01233395168541\\
12 0.948751478612864\\
16 0.888695285506406\\
20 0.84029227895075\\
24 0.835049102398601\\
28 0.819428497231644\\
32 0.79964339696066\\
40 0.78731177310467\\
48 0.789966542297312\\
56 0.772800909726213\\
64 0.77692629298753\\
72 0.771445851992993\\
80 0.763871732874348\\
};

\addplot [color=mycolor1, line width=1.5pt]
  table[row sep=crcr]{%
4 0.381607757962164\\
8 0.0716673346210305\\
12 0.0321535464249558\\
16 0.0183476508256875\\
20 0.0115888275214852\\
24 0.00713773591921731\\
28 0.00567519060715087\\
32 0.00409864629374256\\
40 0.00246616987907075\\
48 0.00182470104591854\\
56 0.00144404775307625\\
64 0.00105440000073531\\
72 0.000749254351752995\\
80 0.000615529723958386\\
};

\addplot [color=mycolor2, dashed, line width=1.5pt]
  table[row sep=crcr]{%
4 1.04938301849011\\
8 0.520404408257198\\
12 0.482923075880042\\
16 0.443405555585209\\
20 0.413208073717047\\
24 0.411061071196459\\
28 0.402486472716688\\
32 0.390424365686607\\
40 0.383522324770411\\
48 0.385621449723859\\
56 0.376087180510243\\
64 0.378371536053034\\
72 0.375356928420968\\
80 0.371026011424398\\
};

\addplot [color=mycolor2, line width=1.5pt]
  table[row sep=crcr]{%
4 0.15852810551121\\
8 0.0259468443342064\\
12 0.0113380004100251\\
16 0.00621609377688842\\
20 0.00379407210836343\\
24 0.00233217153469493\\
28 0.00190554552196032\\
32 0.00134282489769875\\
40 0.000787424196741551\\
48 0.000586123490018031\\
56 0.000469300047040188\\
64 0.000331250878454299\\
72 0.000239546700331839\\
80 0.000193970320616403\\
};

\addplot [color=mycolor3, dashed, line width=1.5pt]
  table[row sep=crcr]{%
4 2.22645835330622\\
8 0.993573038882083\\
12 0.853600520277775\\
16 0.76163269152405\\
20 0.708071499028766\\
24 0.674386017813006\\
28 0.653217622823807\\
32 0.635656762099673\\
40 0.624273633143486\\
48 0.622248780695074\\
56 0.598515172782713\\
64 0.598500706837513\\
72 0.595032341284263\\
80 0.588412058103731\\
};

\addplot [color=mycolor3, line width=1.5pt]
  table[row sep=crcr]{%
4 0.881183904128799\\
8 0.18634268966809\\
12 0.0940366525853945\\
16 0.0552755052815264\\
20 0.0349027746339335\\
24 0.0220231471396096\\
28 0.0164546329159128\\
32 0.012165428349051\\
40 0.00836738982560332\\
48 0.00571556082385228\\
56 0.00393762687334373\\
64 0.00304117326274118\\
72 0.00242603500237173\\
80 0.00189067295550246\\
};

\end{axis}

\end{tikzpicture}%

%% file: figs/results_prob_clust_emp_theo_metric_comp/legend_metrics.tex
\definecolor{darkgray176}{RGB}{176,176,176}
\definecolor{darkorange25512714}{RGB}{255,127,14}
\definecolor{forestgreen4416044}{RGB}{44,160,44}
\definecolor{lightgray204}{RGB}{204,204,204}
\definecolor{magenta}{RGB}{255,0,255}
\definecolor{orange2551868}{RGB}{255,186,8}
\definecolor{saddlebrown164660}{RGB}{164,66,0}
\definecolor{steelblue31119180}{RGB}{31,119,180}
\definecolor{saddlebrown164660}{RGB}{164,66,0}

\begin{tikzpicture} 
\begin{axis}[
hide axis,
width=0.8\linewidth,
height=0.1\linewidth,
at={(0,0)},
xmin=10,
xmax=50,
ymin=0,
ymax=0.4,
legend style={fill opacity=1, draw opacity=1, text opacity=1, draw=lightgray204, legend columns=3,
    anchor=south,
    at={(0.7, 0)},
}
]
\addlegendimage{darkorange25512714, line width=2pt};
\addlegendentry{LE (EMP) \hspace{1em}};

\addlegendimage{forestgreen4416044, line width=2pt};
\addlegendentry{KL (EMP)\hspace{1em}};

\addlegendimage{steelblue31119180, line width=2pt};
\addlegendentry{EU (EMP)\hspace{1em}};

\addlegendimage{darkorange25512714, dashed,line width=2pt};
\addlegendentry{LE (THE) \hspace{1em}};

\addlegendimage{forestgreen4416044, dashed,line width=2pt};
\addlegendentry{KL (THE) \hspace{1em}};

\addlegendimage{steelblue31119180, dashed, line width=2pt};
\addlegendentry{EU (THE) \hspace{1em}};

\coordinate (legend) at (axis description cs:0.97,0.03);

\end{axis}

\end{tikzpicture}

%% file: figs/results_prob_clust_emp_theo_metric_comp/M_150_c1_0.67_c2_0.67_c3_0.67_c_pergroup.tex
\begin{tikzpicture}

\definecolor{darkgray176}{RGB}{176,176,176}
\definecolor{lightgray204}{RGB}{204,204,204}

\definecolor{blue}{RGB}{31,119,180}
\definecolor{red}{RGB}{255,127,14}
\definecolor{green}{RGB}{44,160,44}

\begin{axis}[
width=0.8\linewidth,
height=1.5in,
at={(0in,2.3in)},
scale only axis,
legend cell align={left},
legend style={
  fill opacity=0.8,
  draw opacity=1,
  text opacity=1,
  at={(0.91,0.5)},
  anchor=east,
  draw=lightgray204
},
tick align=outside,
tick pos=left,
x grid style={darkgray176},
xtick style={color=black},
y grid style={darkgray176},
ymin=-0.01, ymax=1.001,
ytick style={color=black},
xmin=10, xmax=150,
ytick={0, 0.25, 0.5, 0.75, 1},
ylabel={Prob. Correct Clustering},
xlabel={Growing $M$}
]
\addplot [semithick, blue, line width=1.2pt]
table {%
10 0.00781302216091821
10 0.00781302216091821
16 0.0213363222233046
26 0.0727284564057784
30 0.106484299080488
30 0.106484299080488
36 0.171714689574617
38 0.196895019448373
46 0.309399000850855
50 0.369123228284365
50 0.369123228284365
70 0.636767630001947
90 0.806253501409712
90 0.806253501409712
110 0.901079365530342
130 0.95277923943442
130 0.95277923943442
150 0.979493861282436
150 0.979493861282436
};
\addplot [semithick, blue, dashed, line width=1.2pt]
table {%
10 0.013
10 0.013
16 0.025
26 0.085
30 0.133
30 0.133
36 0.173
38 0.21
46 0.339
50 0.386
50 0.386
70 0.661
90 0.787
90 0.787
110 0.894
130 0.949
130 0.949
150 0.971
150 0.971
};
\addplot [semithick, green, line width=1.2pt]
table {%
10 0.00704835551692823
10 0.00704835551692823
16 0.0182191729729036
26 0.053726027722808
30 0.0746201601653582
30 0.0746201601653582
36 0.113751507998189
38 0.128826799021584
46 0.198539379617419
50 0.238069940920949
50 0.238069940920949
70 0.457160129633777
90 0.663762645132365
90 0.663762645132365
110 0.819788483890557
130 0.916636973601159
130 0.916636973601159
150 0.96655818767949
150 0.96655818767949
};
\addplot [semithick, green, dashed, line width=1.2pt]
table {%
10 0.003
10 0.003
16 0.00800000000000001
26 0.045
30 0.045
30 0.045
36 0.0669999999999999
38 0.106
46 0.165
50 0.204
50 0.204
70 0.438
90 0.696
90 0.696
110 0.844
130 0.944
130 0.944
150 0.961
150 0.961
};
\addplot [semithick, red, line width=1.2pt]
table {%
10 0.005
16 0.013
26 0.074
30 0.092
36 0.157
38 0.179
46 0.317
70 0.656
90 0.883
110 0.954
130 0.981
150 0.998
};
\addplot [semithick, red, dashed, line width=1.2pt]
table {%
10 0.00385038502176739
16 0.0144662959051173
26 0.0590912616220176
30 0.0949304824554701
36 0.155337900512059
38 0.178462993796234
46 0.313614663540532
50 0.342924146050797
70 0.657781066890964
90 0.86835291849172
110 0.957640688887887
130 0.988402241544113
150 0.996016218459019
};

\addplot [semithick, lightgray, dashed]
table {%
0 0.25
150 0.25 
};

\addplot [semithick, lightgray, dashed]
table {%
0 0.5
150 0.5 
};

\addplot [semithick, lightgray, dashed]
table {%
0 0.75
150 0.75 
};
\end{axis}

\end{tikzpicture}

%% file: figs/results_prob_clust_emp_theo_metric_comp/M_150_c1_0.25_c2_0.33_c3_0.50_c_pergroup.tex
\begin{tikzpicture}

\definecolor{darkgray176}{RGB}{176,176,176}
\definecolor{lightgray204}{RGB}{204,204,204}

\definecolor{blue}{RGB}{31,119,180}
\definecolor{red}{RGB}{255,127,14}
\definecolor{green}{RGB}{44,160,44}

\begin{axis}[
width=0.8\linewidth,
height=1.5in,
at={(0in,2.3in)},
scale only axis,
legend cell align={left},
legend style={
  fill opacity=0.8,
  draw opacity=1,
  text opacity=1,
  at={(0.91,0.5)},
  anchor=east,
  draw=lightgray204
},
tick align=outside,
tick pos=left,
x grid style={darkgray176},
xtick style={color=black},
y grid style={darkgray176},
ymin=-0.01, ymax=1.001,
ytick style={color=black},
xmin=10, xmax=150,
ytick={0, 0.25, 0.5, 0.75, 1},
xlabel={Growing $M$}
]
\addplot [semithick, blue, line width=1.2pt]
table {%
10 0.0218967686340733
10 0.0218967686340733
20 0.127693169259333
30 0.331838024488915
30 0.331838024488915
40 0.542529692796407
50 0.695307984934852
50 0.695307984934852
60 0.795871623033789
70 0.864835425127263
80 0.913644191054176
90 0.947484731918091
90 0.947484731918091
100 0.969765050245605
130 0.995916940841742
130 0.995916940841742
150 0.999203591080766
150 0.999203591080766
};
\addplot [semithick, blue, dashed, line width=1.2pt]
table {%
10 0.028
10 0.028
20 0.169
30 0.408
30 0.408
40 0.593
50 0.727
50 0.727
60 0.809
70 0.873
80 0.897
90 0.924
90 0.924
100 0.959
130 0.99
130 0.99
150 0.995
150 0.995
};
\addplot [semithick, green, line width=1.2pt]
table {%
10 0.0388978257114615
10 0.0388978257114615
20 0.231646839721973
30 0.519209479096969
30 0.519209479096969
40 0.740480162853171
50 0.869857187661777
50 0.869857187661777
60 0.940081178747237
70 0.975457779980987
80 0.99123164564313
90 0.997296651355034
90 0.997296651355034
100 0.99927370226991
130 0.999994042361338
130 0.999994042361338
150 1.00000150202306
150 1.00000150202306
};
\addplot [semithick, green, dashed, line width=1.2pt]
table {%
10 0.031
10 0.031
20 0.24
30 0.561
30 0.561
40 0.752
50 0.863
50 0.863
60 0.93
70 0.966
80 0.99
90 0.998
90 0.998
100 0.998
130 1
130 1
150 1
150 1
};
\addplot [semithick, red, line width=1.2pt]
table {%
10 0.042
10 0.042
20 0.313
30 0.643
30 0.643
40 0.823
50 0.927
50 0.927
60 0.97
70 0.988
80 0.999
90 0.999
90 0.999
100 0.999
130 1
130 1
150 1
150 1
};
\addplot [semithick, red, dashed, line width=1.2pt]
table {%
10 0.028662850949692
10 0.0286564585008298
20 0.294017570026878
30 0.628423584243769
30 0.628416444110533
40 0.827020295543702
50 0.924976079215114
50 0.924972632164806
60 0.971760151791799
70 0.99025125599673
80 0.998363619544333
90 0.999544402005646
90 0.999543729275969
100 0.999898520181567
130 1.00000091219098
130 0.999996769373473
150 0.999998496094731
150 0.999990594263119
};

\addplot [semithick, lightgray, dashed]
table {%
0 0.25
150 0.25 
};

\addplot [semithick, lightgray, dashed]
table {%
0 0.5
150 0.5 
};

\addplot [semithick, lightgray, dashed]
table {%
0 0.75
150 0.75 
};
\end{axis}

\end{tikzpicture}

%% file: figs/results_prob_clust_emp_theo_metric_comp/M_150_c1_0.50_c2_0.33_c3_0.25_c_pergroup.tex
\begin{tikzpicture}

\definecolor{darkgray176}{RGB}{176,176,176}
\definecolor{lightgray204}{RGB}{204,204,204}

\definecolor{blue}{RGB}{31,119,180}
\definecolor{red}{RGB}{255,127,14}
\definecolor{green}{RGB}{44,160,44}

\begin{axis}[
width=0.8\linewidth,
height=1.5in,
at={(0in,2.3in)},
scale only axis,
legend cell align={left},
legend style={
  fill opacity=0.8,
  draw opacity=1,
  text opacity=1,
  at={(0.91,0.5)},
  anchor=east,
  draw=lightgray204
},
tick align=outside,
tick pos=left,
x grid style={darkgray176},
xtick style={color=black},
y grid style={darkgray176},
ymin=-0.01, ymax=1.001,
ytick style={color=black},
xmin=10, xmax=150,
ytick={0, 0.25, 0.5, 0.75, 1},
xlabel={Growing $M$}
]
\addplot [semithick, blue, line width=1.2pt]
table {%
10 0.0235852089869021
10 0.0235852089869021
20 0.15335597296106
30 0.412542730681812
30 0.412542730681812
40 0.671547971909476
50 0.8440868877437
50 0.8440868877437
60 0.935884983788541
70 0.977000093991809
80 0.992739678746709
90 0.997961689801271
150 1
};
\addplot [semithick, blue, dashed, line width=1.2pt]
table {%
10 0.027
10 0.027
20 0.148
30 0.441
30 0.441
40 0.692
50 0.826
50 0.826
60 0.929
70 0.971
80 0.986
90 0.997
150 1
};
\addplot [semithick, green, line width=1.2pt]
table {%
10 0.0557118155049418
10 0.0557118155049418
20 0.251672296663488
30 0.496416924848528
30 0.496416924848528
40 0.712944441530559
50 0.861347858540026
50 0.861347858540026
60 0.942456416598431
70 0.979250674609726
80 0.993475699560698
90 0.998212544467733
150 1
};
\addplot [semithick, green, dashed, line width=1.2pt]
table {%
10 0.038
10 0.038
20 0.211
30 0.475
30 0.475
40 0.735
50 0.869
50 0.869
60 0.958
70 0.979
80 0.998
90 0.998
150 1
};
\addplot [semithick, red, line width=1.2pt]
table {%
10 0.045
10 0.045
20 0.266
30 0.548
30 0.548
40 0.812
50 0.92
50 0.92
60 0.978
70 0.993
80 1
90 0.999
150 1
};
\addplot [semithick, red, dashed, line width=1.2pt]
table {%
10 0.0390398836364324
10 0.0390377967421003
20 0.252046547084184
30 0.550943321460429
30 0.550951970314834
40 0.802740939160876
50 0.915295172578657
50 0.915296616502779
60 0.977902566193795
70 0.994098502476453
80 0.998622937534837
90 0.999805964763581
150 1
};

\addplot [semithick, lightgray, dashed]
table {%
0 0.25
150 0.25 
};

\addplot [semithick, lightgray, dashed]
table {%
0 0.5
150 0.5 
};

\addplot [semithick, lightgray, dashed]
table {%
0 0.75
150 0.75 
};
\end{axis}

\end{tikzpicture}

%% file: figs/results_prob_clust_plug_vs_consistent/legend_multiple_setups.tex
\definecolor{darkgray176}{RGB}{176,176,176}
\definecolor{darkorange25512714}{RGB}{255,127,14}
\definecolor{forestgreen4416044}{RGB}{44,160,44}
\definecolor{lightgray204}{RGB}{204,204,204}
\definecolor{magenta}{RGB}{255,0,255}
\definecolor{saddlebrown164660}{RGB}{164,66,0}
\definecolor{steelblue31119180}{RGB}{31,119,180}
\definecolor{saddlebrown164660}{RGB}{164,66,0}

\definecolor{orange2551868}{RGB}{164,66,0}

\begin{tikzpicture} 
\begin{axis}[
hide axis,
width=0.8\linewidth,
height=0.1\linewidth,
at={(0,0)},
xmin=10,
xmax=50,
ymin=0,
ymax=0.4,
legend style={fill opacity=1, draw opacity=1, text opacity=1, draw=lightgray204, legend columns=3,
    anchor=south,
    at={(0.7, 0)},
}
]

\addlegendimage{steelblue31119180, mark=square*, only marks};
\addlegendentry{$c_1=2/3, c_3=2/3, c_5=2/3$ \hspace{1em}};

\addlegendimage{orange2551868, mark=square*, only marks};
\addlegendentry{$c_1=1/2, c_3=1/2, c_5=1/2$ \hspace{1em}};

\addlegendimage{darkorange25512714, mark=square*, only marks};
\addlegendentry{$c_1=1/2, c_3=1/3, c_5=1/4$ \hspace{1em}};

\addlegendimage{forestgreen4416044, mark=square*, only marks};
\addlegendentry{$c_1=1/4, c_3=1/3, c_5=1/2$ \hspace{1em}};



\addlegendimage{magenta, mark=square*, only marks};
\addlegendentry{$c_1=1/3, c_3=1/2, c_5=1/2$ \hspace{1em}};

    \coordinate (legend) at (axis description cs:0.97,0.03);

\end{axis}

\end{tikzpicture}

%% file: figs/results_prob_clust_plug_vs_consistent/tkz_eu_plugin_vs_consistent_rho_0.30_0.50_0.70.tex
\begin{tikzpicture}

\definecolor{darkgray176}{RGB}{176,176,176}
\definecolor{darkorange25512714}{RGB}{255,127,14}
\definecolor{forestgreen4416044}{RGB}{44,160,44}
\definecolor{lightgray204}{RGB}{204,204,204}
\definecolor{magenta}{RGB}{255,0,255}
\definecolor{orange2551868}{RGB}{255,186,8}
\definecolor{steelblue31119180}{RGB}{31,119,180}

\definecolor{orange2551868}{RGB}{164,66,0}
\begin{axis}[
width=0.25\linewidth,
height=1.5in,
at={(0in,2.3in)},
scale only axis,
legend cell align={left},
legend style={
  fill opacity=0.8,
  draw opacity=1,
  text opacity=1,
  at={(0.91,0.5)},
  anchor=east,
  draw=lightgray204
},
tick align=outside,
tick pos=left,
x grid style={darkgray176},
xtick style={color=black},
y grid style={darkgray176},
ymin=-0.01, ymax=1.001,
ytick style={color=black},
xmin=10, xmax=150,
ytick={0, 0.25, 0.5, 0.75, 1},
ylabel={Prob. Correct Clustering},
xlabel={Growing $M$},
title={EU}
]

\addplot [semithick, steelblue31119180, line width=1.2pt]
table {%
10 0.013
12.8571428571429 0.0192062970480652
15.7142857142857 0.0244098498017897
18.5714285714286 0.0318560052488769
21.4285714285714 0.04479011037703
24.2857142857143 0.0664575121739525
27.1428571428571 0.09973099567458
58.5714285714286 0.505190244828412
61.4285714285714 0.547750030159145
64.2857142857143 0.588996274812993
67.1428571428571 0.627291943267448
70 0.661
72.8571428571429 0.688938242476408
75.7142857142857 0.711743800115508
78.5714285714286 0.730508635324402
81.4285714285714 0.746324710510195
84.2857142857143 0.76028398807999
87.1428571428571 0.773478430440891
90 0.787
92.8571428571429 0.801682538491943
95.7142857142857 0.817327404961428
98.5714285714286 0.833477837780684
101.428571428571 0.84967707532194
104.285714285714 0.865468355957425
107.142857142857 0.880394918059369
110 0.894
112.857142857143 0.905929271194303
115.714285714286 0.916238125228285
118.571428571429 0.925084386730705
121.428571428571 0.932625880330325
124.285714285714 0.939020430655907
127.142857142857 0.944425862336212
130 0.949
132.857142857143 0.952900668276033
135.714285714286 0.956285691793072
138.571428571429 0.959312895179879
141.428571428571 0.962140103065213
144.285714285714 0.964925140077838
147.142857142857 0.967825830846513
150 0.971
};
\addplot [semithick, steelblue31119180, dashed, line width=1.2pt]
table {%
10 0.00900000000000001
12.8571428571429 0.0122493770403933
15.7142857142857 0.0138218844155242
18.5714285714286 0.0166549942955776
21.4285714285714 0.0236861788507384
24.2857142857143 0.0378529102511912
27.1428571428571 0.0617814559494834
30 0.086
32.8571428571429 0.0944449509049248
35.7142857142857 0.106498651998409
38.5714285714286 0.142430598682039
41.4285714285714 0.187042932155211
44.2857142857143 0.230862951542525
47.1428571428571 0.267420313236926
50 0.298
52.8571428571429 0.33171193625613
55.7142857142857 0.369341332052652
58.5714285714286 0.409423895842675
61.4285714285714 0.45049533607931
64.2857142857143 0.491091361215669
67.1428571428571 0.529747679704862
70 0.565
72.8571428571429 0.595761716260626
75.7142857142857 0.622456965472012
78.5714285714286 0.645887570325863
81.4285714285714 0.666855353513883
84.2857142857143 0.686162137727776
87.1428571428571 0.704609745659247
90 0.723
92.8571428571429 0.741935076252386
95.7142857142857 0.76121856116134
98.5714285714286 0.780454394282444
101.428571428571 0.79924651517128
104.285714285714 0.81719886338343
107.142857142857 0.833915378474477
110 0.849
112.857142857143 0.862162701761901
115.714285714286 0.873537594547352
118.571428571429 0.883364823389842
121.428571428571 0.891884533322862
124.285714285714 0.899336869379902
127.142857142857 0.905961976594451
130 0.912
132.857142857143 0.917691084630039
135.714285714286 0.923275375518058
138.571428571429 0.928993017697546
141.428571428571 0.935084156201995
144.285714285714 0.941788936064893
147.142857142857 0.949347502319732
150 0.958
};
\addplot [semithick, darkorange25512714, line width=1.2pt]
table {%
10 0.027
12.8571428571429 0.0301941758000708
15.7142857142857 0.0612554002898668
18.5714285714286 0.11465318052411
21.4285714285714 0.184857023557523
24.2857142857143 0.266336436444827
27.1428571428571 0.353560926240745
30 0.441
32.8571428571429 0.523662404078276
35.7142857142857 0.598713842035109
38.5714285714286 0.663859256731001
41.4285714285714 0.71693921918938
44.2857142857143 0.758913748181057
47.1428571428571 0.793862310224223
50 0.826
52.8571428571429 0.858417673715614
55.7142857142857 0.889709233682716
58.5714285714286 0.917344343739062
61.4285714285714 0.9388809663766
64.2857142857143 0.953907933133705
67.1428571428571 0.964044944595174
70 0.971
72.8571428571429 0.976272491497174
75.7142857142857 0.980527382875685
78.5714285714286 0.984221030834524
81.4285714285714 0.987793726992642
84.2857142857143 0.991316266128123
87.1428571428571 0.994489946178183
90 0.997
92.8571428571429 0.99862461249263
95.7142857142857 0.999513776722645
98.5714285714286 0.999910437798495
101.428571428571 1.00005336895479
104.285714285714 1.00008539032766
107.142857142857 1.00005336895479
110 1
112.857142857143 0.99996035449073
115.714285714286 0.999939006908815
118.571428571429 0.999932907599697
121.428571428571 0.999939006908815
124.285714285714 0.999954255181612
127.142857142857 0.999975602763526
130 1
132.857142857143 1.00002439723647
135.714285714286 1.00004574481839
138.571428571429 1.00006099309118
141.428571428571 1.0000670924003
144.285714285714 1.00006099309118
147.142857142857 1.00003964550927
150 1
};
\addplot [semithick, darkorange25512714, dashed, line width=1.2pt]
table {%
10 0.011
12.8571428571429 0.0223639274177599
15.7142857142857 0.0291802539483721
18.5714285714286 0.0324190403548591
21.4285714285714 0.033050347400243
24.2857142857143 0.0320442358475463
27.1428571428571 0.0303707664597912
30 0.029
32.8571428571429 0.0287249393836247
35.7142857142857 0.0296303561358354
38.5714285714286 0.031623963934232
41.4285714285714 0.0345999645638731
44.2857142857143 0.0381417862813666
47.1428571428571 0.0415220838148702
50 0.044
52.8571428571429 0.0450242910083789
55.7142857142857 0.0448021663556556
58.5714285714286 0.0437304488934854
61.4285714285714 0.0422121302549131
64.2857142857143 0.0407920840449391
67.1428571428571 0.0401570658405196
70 0.041
72.8571428571429 0.0436034729260991
75.7142857142857 0.0466087191990272
78.5714285714286 0.0482466354433677
81.4285714285714 0.0468421622119068
84.2857142857143 0.0428832504060892
87.1428571428571 0.0390208612760182
90 0.038
92.8571428571429 0.0415992570707104
95.7142857142857 0.0477315642383044
98.5714285714286 0.0533434385673067
101.428571428571 0.0554819307992455
104.285714285714 0.0535063662467228
107.142857142857 0.0490883447934146
110 0.044
112.857142857143 0.0397030521559624
115.714285714286 0.0364175684660019
118.571428571429 0.0340532028636225
121.428571428571 0.0325196092823284
124.285714285714 0.0317264416556238
127.142857142857 0.031583353917013
130 0.032
132.857142857143 0.0328860338380891
135.714285714286 0.0341511093647844
138.571428571429 0.03570488051359
141.428571428571 0.0374570012180103
144.285714285714 0.0393171254115492
147.142857142857 0.0411949070277111
150 0.043
};
\addplot [semithick, forestgreen4416044, line width=1.2pt]
table {%
10 0.028
12.8571428571429 0.0469223307485282
15.7142857142857 0.0852080450342425
18.5714285714286 0.138311789442268
21.4285714285714 0.201688210557732
24.2857142857143 0.270791954965758
27.1428571428571 0.341077669251472
30 0.408
32.8571428571429 0.468013757706969
35.7142857142857 0.52157440851001
38.5714285714286 0.570137582457257
41.4285714285714 0.615110722685325
44.2857142857143 0.656792973365893
47.1428571428571 0.694375179705709
50 0.727
52.8571428571429 0.754258615341455
55.7142857142857 0.777536298012607
58.5714285714286 0.798666843093448
61.4285714285714 0.819410011993087
64.2857142857143 0.839822791690275
67.1428571428571 0.858259394733409
70 0.873
72.8571428571429 0.882939218589441
75.7142857142857 0.889429390488646
78.5714285714286 0.89443728790641
81.4285714285714 0.89988600498465
84.2857142857143 0.906694040327105
87.1428571428571 0.914775296999328
90 0.924
92.8571428571429 0.934135342869903
95.7142857142857 0.944536393318232
98.5714285714286 0.954455187596291
101.428571428571 0.963162354398388
104.285714285714 0.970356148608054
107.142857142857 0.976162451298034
110 0.980725735984085
112.857142857143 0.984190476181965
115.714285714286 0.986701145407429
118.571428571429 0.988402217176234
121.428571428571 0.989438165004138
124.285714285714 0.989953462406895
127.142857142857 0.990092582900264
130 0.99
132.857142857143 0.989820187221861
135.714285714286 0.989697618081602
138.571428571429 0.98977676609498
141.428571428571 0.990202104777753
144.285714285714 0.991118107645676
147.142857142857 0.992669248214506
150 0.995
};
\addplot [semithick, forestgreen4416044, dashed, line width=1.2pt]
table {%
10 0.012
12.8571428571429 0.0143217549169594
15.7142857142857 0.0153217549169594
18.5714285714286 0.0153001305382123
21.4285714285714 0.0145570123189305
24.2857142857143 0.0133925307973263
27.1428571428571 0.012106816511612
30 0.011
32.8571428571429 0.010258670183773
35.7142857142857 0.00961524951649576
38.5714285714286 0.00868861883480338
41.4285714285714 0.00711652890916022
44.2857142857143 0.00497073898810216
47.1428571428571 0.00275701679823655
50 0.001
52.8571428571429 8.388421562956e-05
55.7142857142857 -0.000168903085435723
58.5714285714286 -7.63764719563163e-05
61.4285714285714 5.1068794152215e-05
64.2857142857143 7.82815084070338e-05
67.1428571428571 4.53545237580314e-05
70 -1.01643953670516e-20
72.8571428571429 -2.01512232682002e-05
75.7142857142857 -1.77935872314391e-05
78.5714285714286 -5.70285327051899e-06
81.4285714285714 3.68202238186661e-06
84.2857142857143 5.66860189952487e-06
87.1428571428571 3.31096586276377e-06
90 4.2351647362715e-22
92.8571428571429 -1.56985450389662e-06
95.7142857142857 -1.48993463824369e-06
98.5714285714286 -5.48021935905727e-07
101.428571428571 4.85227755749865e-07
104.285714285714 1.23304935578789e-06
107.142857142857 1.7125685497054e-06
110 1.95803670849651e-06
112.857142857143 2.00370520315532e-06
115.714285714286 1.88382540467594e-06
118.571428571429 1.63264868405248e-06
121.428571428571 1.28442641227905e-06
124.285714285714 8.73409960349753e-07
127.142857142857 4.33850699258702e-07
130 4.2351647362715e-22
132.857142857143 -3.93890766432242e-07
135.714285714286 -7.13570229043916e-07
138.571428571429 -9.24787016840916e-07
141.428571428571 -9.93289758829131e-07
144.285714285714 -8.84827084014456e-07
147.142857142857 -5.65147621402782e-07
150 0
};
\addplot [semithick, magenta, line width=1.2pt]
table {%
10 0.019
12.0408163265306 0.0288340137310415
14.0816326530612 0.0419829908386408
16.1224489795918 0.0584619649020636
18.1632653061224 0.0782859695005754
20.2040816326531 0.101470038213442
22.2448979591837 0.128029204619928
24.2857142857143 0.1579785022993
26.3265306122449 0.191332964830823
28.3673469387755 0.228107625793762
30.4081632653061 0.268315562529043
32.4489795918367 0.311555129849378
34.4897959183673 0.356499381832365
36.530612244898 0.4016961733018
38.5714285714286 0.445693359081482
40.6122448979592 0.487045466538462
42.6530612244898 0.52482328018263
44.6938775510204 0.558972676215018
46.734693877551 0.589524296849102
48.7755102040816 0.616508784298358
50.8163265306122 0.639968460109326
52.8571428571429 0.660399679901416
54.8979591836735 0.678888605613781
56.9387755102041 0.696560816934665
58.9795918367347 0.714541893552313
61.0204081632653 0.733914965146632
63.0612244897959 0.754786811205759
65.1020408163265 0.776287861026066
67.1428571428571 0.797506093895585
69.1836734693878 0.817529489102347
71.2244897959184 0.835481542247852
73.265306122449 0.85101717821656
75.3061224489796 0.864200417207291
77.3469387755102 0.875105802771007
79.3877551020408 0.883807878458665
81.4285714285714 0.890426595233086
83.469387755102 0.89555067270256
85.5102040816327 0.900045378823415
87.5510204081633 0.904779555896348
89.5918367346939 0.910622046222062
91.6326530612245 0.918299629125707
93.6734693877551 0.927489149753771
95.7142857142857 0.937396869646239
97.7551020408163 0.947226830609099
99.7959183673469 0.956183074448344
101.836734693878 0.963469642969963
103.877551020408 0.968290577979948
105.918367346939 0.969849921284289
107.959183673469 0.967351714688976
110 0.97
130 0.99
150 0.99
};
\addplot [semithick, magenta, dashed, line width=1.2pt]
table {%
10 0.018
12.0408163265306 0.0217319796870627
14.0816326530612 0.0240875016779086
16.1224489795918 0.0253767865994792
18.1632653061224 0.0259100550787157
20.2040816326531 0.0259975277425596
22.2448979591837 0.0259494252179523
24.2857142857143 0.0260759681318351
26.3265306122449 0.0266873771111496
28.3673469387755 0.028093872782837
30.4081632653061 0.0306042139919935
32.4489795918367 0.0342172618325215
34.4897959183673 0.0382404545855185
36.530612244898 0.0418876764939859
38.5714285714286 0.0443728118009249
40.6122448979592 0.0449168565125156
42.6530612244898 0.0432910463860631
44.6938775510204 0.0401993117498319
46.734693877551 0.0364359286643199
48.7755102040816 0.0327951731900253
50.8163265306122 0.0300585608822359
52.8571428571429 0.0285115426561957
54.8979591836735 0.027795163914037
56.9387755102041 0.0275074033528079
58.9795918367347 0.0272462396695565
61.0204081632653 0.0266286064676472
63.0612244897959 0.0257074001957171
65.1020408163265 0.0249714801476763
67.1428571428571 0.0249286605237509
69.1836734693878 0.0260867555241672
71.2244897959184 0.0289005131662014
73.265306122449 0.0330306541370639
75.3061224489796 0.0375266553129488
77.3469387755102 0.0414222702565835
79.3877551020408 0.0437512525306956
81.4285714285714 0.0436487384981344
83.469387755102 0.041296501534087
85.5102040816327 0.0374937746914511
87.5510204081633 0.0330477715933961
89.5918367346939 0.0287657058630919
91.6326530612245 0.0253669101348635
93.6734693877551 0.0229212216098582
95.7142857142857 0.021207371713676
97.7551020408163 0.0200027187314664
99.7959183673469 0.0190846209483788
101.836734693878 0.0182304366495626
103.877551020408 0.0172175241201672
105.918367346939 0.015823241645342
107.959183673469 0.0138249475102365
110 0.011
150 0.0
};
\addplot [semithick, orange2551868, line width=1.2pt]
table {%
10 0.02
12.8571428571429 0.0359194600714552
15.7142857142857 0.0523684396632919
18.5714285714286 0.0711596207632759
21.4285714285714 0.094105685359173
24.2857142857143 0.123019315438749
27.1428571428571 0.159713192989769
30 0.206
32.8571428571429 0.262392856002969
35.7142857142857 0.324206630715251
38.5714285714286 0.385456631399184
41.4285714285714 0.440344389692913
44.2857142857143 0.487354597878144
47.1428571428571 0.529255108880148
50 0.569
52.8571428571429 0.608778774698968
55.7142857142857 0.647722641079081
58.5714285714286 0.684198233402557
61.4285714285714 0.716642884044227
64.2857142857143 0.745119981968969
67.1428571428571 0.771318972731714
70 0.797
72.8571428571429 0.823218444593082
75.7142857142857 0.848210635937071
78.5714285714286 0.869508140609793
81.4285714285714 0.884760943194484
84.2857142857143 0.894342642398798
87.1428571428571 0.901350451054809
90 0.909
92.8571428571429 0.919684697417589
95.7142857142857 0.932509060873733
98.5714285714286 0.945755385280322
101.428571428571 0.957745588622212
104.285714285714 0.967712919562424
107.142857142857 0.975801957442142
110 0.982196904675516
112.857142857143 0.987081963676694
115.714285714286 0.990641336859826
118.571428571429 0.99305922663906
121.428571428571 0.994519835428546
124.285714285714 0.995207365642431
127.142857142857 0.995306019694867
130 0.995
132.857142857143 0.994473508971981
135.714285714286 0.993910749024958
138.571428571429 0.99349592257308
141.428571428571 0.993413232030496
144.285714285714 0.993846879811356
147.142857142857 0.994981068329807
150 0.997
};
\addplot [semithick, orange2551868, dashed, line width=1.2pt]
table {%
10 0.017
12.8571428571429 0.0272204858142968
15.7142857142857 0.0377919143857254
18.5714285714286 0.0504024800400045
21.4285714285714 0.0667403771028528
24.2857142857143 0.088493799899989
27.1428571428571 0.117350942757132
30 0.155
32.8571428571429 0.202000738392442
35.7142857142857 0.254399214450825
38.5714285714286 0.307113057129645
41.4285714285714 0.355248015869875
44.2857142857143 0.398236611301399
47.1428571428571 0.439838135243016
50 0.484
52.8571428571429 0.533206761300882
55.7142857142857 0.584091548567222
58.5714285714286 0.631824634643835
61.4285714285714 0.67174642233487
64.2857142857143 0.703110303509095
67.1428571428571 0.729082659099898
70 0.753
72.8571428571429 0.777366223026909
75.7142857142857 0.801354768697282
78.5714285714286 0.823306463452565
81.4285714285714 0.841627552794987
84.2857142857143 0.856228920624764
87.1428571428571 0.868526089240105
90 0.88
92.8571428571429 0.891803743000036
95.7142857142857 0.903779003282186
98.5714285714286 0.915439614625022
101.428571428571 0.926310435969615
104.285714285714 0.936169904994576
107.142857142857 0.945050038116049
110 0.952993876912683
112.857142857143 0.960044462963124
115.714285714286 0.966244837846021
118.571428571429 0.971638043140021
121.428571428571 0.976267120423772
124.285714285714 0.98017511127592
127.142857142857 0.983405057275114
130 0.986
132.857142857143 0.988002981029227
135.714285714286 0.989457041941442
138.571428571429 0.990405224315293
141.428571428571 0.990890569729427
144.285714285714 0.99095611976249
147.142857142857 0.990644915993133
150 0.99
};

\addplot [semithick, lightgray, dashed]
table {%
0 0.25
150 0.25 
};

\addplot [semithick, lightgray, dashed]
table {%
0 0.5
150 0.5 
};

\addplot [semithick, lightgray, dashed]
table {%
0 0.75
150 0.75 
};
\end{axis}

\end{tikzpicture}

%% file: figs/results_prob_clust_plug_vs_consistent/tkz_kl_plugin_vs_consistent_rho_0.30_0.50_0.70.tex
\begin{tikzpicture}

\definecolor{darkgray176}{RGB}{176,176,176}
\definecolor{darkorange25512714}{RGB}{255,127,14}
\definecolor{forestgreen4416044}{RGB}{44,160,44}
\definecolor{lightgray204}{RGB}{204,204,204}
\definecolor{magenta}{RGB}{255,0,255}
\definecolor{orange2551868}{RGB}{255,186,8}
\definecolor{steelblue31119180}{RGB}{31,119,180}
\definecolor{orange2551868}{RGB}{164,66,0}

\begin{axis}[
width=0.25\linewidth,
height=1.5in,
at={(0in,2.3in)},
scale only axis,
legend cell align={left},
legend style={
  fill opacity=0.8,
  draw opacity=1,
  text opacity=1,
  at={(0.91,0.5)},
  anchor=east,
  draw=lightgray204
},
tick align=outside,
tick pos=left,
x grid style={darkgray176},
xtick style={color=black},
y grid style={darkgray176},
ymin=-0.01, ymax=1.001,
ytick style={color=black},
xmin=10, xmax=150,
ytick={0, 0.25, 0.5, 0.75, 1},
xlabel={Growing $M$},
title={Symmetrized KL}
]

\addplot [semithick, steelblue31119180, line width=1.2pt]
table {%
10 0.003
12.8571428571429 0.000972230532544558
15.7142857142857 0.00707098609820188
18.5714285714286 0.0179762528894705
21.4285714285714 0.0303680170988491
24.2857142857143 0.0409262649188363
27.1428571428571 0.0463715153428487
41.4285714285714 0.144120516273484
44.2857142857143 0.156288848572033
47.1428571428571 0.173973477771389
50 0.204
52.8571428571429 0.236058985709318
55.7142857142857 0.268069878433209
58.5714285714286 0.300335451106778
61.4285714285714 0.333158476665135
64.2857142857143 0.366841728043386
67.1428571428571 0.401687978176639
70 0.438
72.8571428571429 0.475897888061885
75.7142857142857 0.514771023363937
78.5714285714286 0.553826108521108
81.4285714285714 0.592269846148349
84.2857142857143 0.62930893886061
87.1428571428571 0.664150089272843
90 0.696
92.8571428571429 0.724285322101451
95.7142857142857 0.74931250041425
98.5714285714286 0.771607928219868
101.428571428571 0.791697998799779
104.285714285714 0.810109105435456
107.142857142857 0.827367641408373
110 0.844
112.857142857143 0.860406887672252
115.714285714286 0.876486263608803
118.571428571429 0.892010400173764
121.428571428571 0.906751569731251
124.285714285714 0.920482044645377
127.142857142857 0.932974097280256
130 0.944
132.857142857143 0.953332025168724
138.571428571429 0.96003532309565
150 0.971
};
\addplot [semithick, steelblue31119180, dashed, line width=1.2pt]
table {%
10 0.003
12.8571428571429 0.000972230532544558
15.7142857142857 0.00707098609820188
18.5714285714286 0.0179762528894705
21.4285714285714 0.0303680170988491
24.2857142857143 0.0409262649188363
27.1428571428571 0.0463715153428487
32.8571428571429 0.0420548716949534
35.7142857142857 0.0625161403521857
38.5714285714286 0.115617616357856
41.4285714285714 0.144120516273484
44.2857142857143 0.156288848572033
47.1428571428571 0.173973477771389
50 0.204
52.8571428571429 0.236058985709318
55.7142857142857 0.268069878433209
58.5714285714286 0.300335451106778
61.4285714285714 0.333158476665135
64.2857142857143 0.366841728043386
67.1428571428571 0.401687978176639
70 0.438
72.8571428571429 0.475897888061885
75.7142857142857 0.514771023363937
78.5714285714286 0.553826108521108
81.4285714285714 0.592269846148349
84.2857142857143 0.62930893886061
87.1428571428571 0.664150089272843
90 0.696
92.8571428571429 0.724285322101451
95.7142857142857 0.74931250041425
98.5714285714286 0.771607928219868
101.428571428571 0.791697998799779
104.285714285714 0.810109105435456
107.142857142857 0.827367641408373
110 0.844
112.857142857143 0.860406887672252
115.714285714286 0.876486263608803
118.571428571429 0.892010400173764
121.428571428571 0.906751569731251
124.285714285714 0.920482044645377
127.142857142857 0.932974097280256
130 0.944
132.857142857143 0.953332025168724
135.714285714286 0.960742445150541
138.571428571429 0.966003532309565
141.428571428571 0.968887559009909
144.285714285714 0.969166797615688
147.142857142857 0.966613520491013
150 0.961
};
\addplot [semithick, darkorange25512714, line width=1.2pt]
table {%
10 0.038
12.8571428571429 0.0740262522640271
15.7142857142857 0.121597680835456
18.5714285714286 0.179067643762754
21.4285714285714 0.244789499094389
24.2857142857143 0.31711660487883
27.1428571428571 0.394402319164544
30 0.475
32.8571428571429 0.556693903944964
35.7142857142857 0.634991881604397
38.5714285714286 0.704832682094562
41.4285714285714 0.761360888074903
44.2857142857143 0.804455253698106
47.1428571428571 0.838728704610101
50 0.869
52.8571428571429 0.898805021041267
55.7142857142857 0.926548136844768
58.5714285714286 0.949350838505722
61.4285714285714 0.964467015586903
64.2857142857143 0.972195722404863
67.1428571428571 0.975881178029933
70 0.979
72.8571428571429 0.984150734656093
75.7142857142857 0.990419643551803
78.5714285714286 0.996014917043861
81.4285714285714 0.999209400112586
84.2857142857143 0.999762994080757
87.1428571428571 0.998922656613618
90 0.998
92.8571428571429 0.998002057146952
95.7142857142857 0.998717543434408
98.5714285714286 0.999630594860522
101.428571428571 1.00024009619338
104.285714285714 1.00038415390941
107.142857142857 1.00024009619338
110 1
112.857142857143 0.999821642827776
115.714285714286 0.999725604350424
118.571428571429 0.999698164785467
121.428571428571 0.999725604350424
124.285714285714 0.999794203262818
127.142857142857 0.99989024174017
130 1
132.857142857143 1.00010975825983
135.714285714286 1.00020579673718
138.571428571429 1.00027439564958
141.428571428571 1.00030183521453
144.285714285714 1.00027439564958
147.142857142857 1.00017835717222
150 1
};
\addplot [semithick, darkorange25512714, dashed, line width=1.2pt]
table {%
10 0.001
12.8571428571429 0.000538312000445386
15.7142857142857 0.000232189551465793
18.5714285714286 5.20594940557052e-05
21.4285714285714 -3.16513307903991e-05
24.2857142857143 -4.85160820780384e-05
27.1428571428571 -2.81079188127322e-05
30 -1.6940658945086e-21
32.8571428571429 1.24840591998084e-05
35.7142857142857 1.10188190070212e-05
38.5714285714286 3.52838348713619e-06
41.4285714285714 -2.27246332189002e-06
44.2857142857143 -3.48329801554428e-06
47.1428571428571 -2.01805782275709e-06
50 3.4410713482206e-22
52.8571428571429 8.96315232398798e-07
55.7142857142857 7.91115798684363e-07
58.5714285714286 2.53326659245085e-07
61.4285714285714 -1.63155716061284e-07
64.2857142857143 -2.50090139581559e-07
67.1428571428571 -1.44890705867126e-07
70 -6.61744490042422e-24
72.8571428571429 6.43540537747492e-08
75.7142857142857 5.68021745598565e-08
78.5714285714286 1.81897422950071e-08
81.4285714285714 -1.17167029679563e-08
84.2857142857143 -1.79639385975518e-08
87.1428571428571 -1.04120593826592e-08
90 -4.13590306276514e-25
92.8571428571429 4.64152044769146e-09
95.7142857142857 4.11464515362918e-09
98.5714285714286 1.32973288501431e-09
101.428571428571 -8.7812549010379e-10
104.285714285714 -1.40500078416606e-09
107.142857142857 -8.78125490103791e-10
110 1.29246970711411e-25
112.857142857143 6.5232179264853e-10
115.714285714286 1.00357198869005e-09
118.571428571429 1.10392918755905e-09
121.428571428571 1.00357198869004e-09
124.285714285714 7.52678991517533e-10
127.142857142857 4.01428795476018e-10
130 -2.06795153138257e-25
132.857142857143 -4.01428795476019e-10
135.714285714286 -7.52678991517535e-10
138.571428571429 -1.00357198869005e-09
141.428571428571 -1.10392918755905e-09
144.285714285714 -1.00357198869005e-09
147.142857142857 -6.52321792648531e-10
150 0
};
\addplot [semithick, forestgreen4416044, line width=1.2pt]
table {%
10 0.031
12.8571428571429 0.0605838548156242
15.7142857142857 0.118012426244196
18.5714285714286 0.195804970497678
21.4285714285714 0.286480743788036
24.2857142857143 0.382559002327233
27.1428571428571 0.476559002327233
30 0.561
32.8571428571429 0.63023768047673
35.7142857142857 0.685973444565542
38.5714285714286 0.73174512199379
41.4285714285714 0.771023639982563
44.2857142857143 0.805741168108902
47.1428571428571 0.836291118305804
50 0.863
52.8571428571429 0.88617224328716
55.7142857142857 0.906023960938701
58.5714285714286 0.922749186394976
61.4285714285714 0.936552839874549
64.2857142857143 0.947890237494775
67.1428571428571 0.957467091271806
70 0.966
72.8571428571429 0.974019206593072
75.7142857142857 0.981309530442155
78.5714285714286 0.987469435057741
81.4285714285714 0.992112619718846
84.2857142857143 0.995203206380633
87.1428571428571 0.997055739674406
90 0.998
92.8571428571429 0.998344462426921
95.7142857142857 0.998312380703368
98.5714285714286 0.998105703247216
101.428571428571 0.997921836182805
104.285714285714 0.997853712883268
107.142857142857 0.997889793970526
110 0.998013997772968
112.857142857143 0.998210242618984
115.714285714286 0.998462446836966
118.571428571429 0.998754528755303
121.428571428571 0.999070406702384
124.285714285714 0.999393999006601
127.142857142857 0.999709223996343
130 1
132.857142857143 1.00025024534596
135.714285714286 1.00044387836262
138.571428571429 1.00056481737837
141.428571428571 1.00059698072159
144.285714285714 1.00052428672067
147.142857142857 1.00033065370401
150 1
};
\addplot [semithick, forestgreen4416044, dashed, line width=1.2pt]
table {%
10 0.001
12.8571428571429 0.000538312000516799
15.7142857142857 0.000232189551537206
18.5714285714286 5.20594940842704e-05
21.4285714285714 -3.16513308189642e-05
24.2857142857143 -4.85160821494512e-05
27.1428571428571 -2.81079188841451e-05
30 5.0821976835258e-21
32.8571428571429 1.24840593426341e-05
35.7142857142857 1.1018819249825e-05
38.5714285714286 3.52838364424448e-06
41.4285714285714 -2.27246356469375e-06
44.2857142857143 -3.4832987725206e-06
47.1428571428571 -2.01805867971142e-06
50 0
52.8571428571429 8.96317160546026e-07
55.7142857142857 7.91119126523653e-07
58.5714285714286 2.53328830196038e-07
61.4285714285714 -1.63159086748291e-07
64.2857142857143 -2.50100665837169e-07
67.1428571428571 -1.44902631814797e-07
70 4.96308367531817e-24
72.8571428571429 6.43809050102242e-08
75.7142857142857 5.68485215061951e-08
78.5714285714286 1.82199785000607e-08
81.4285714285714 -1.17636497823214e-08
84.2857142857143 -1.81105491997599e-08
87.1428571428571 -1.05781656957309e-08
90 8.27180612553028e-25
92.8571428571429 5.01550959711379e-09
95.7142857142857 4.76017456307889e-09
98.5714285714286 1.75086880481063e-09
101.428571428571 -1.55024842092608e-09
104.285714285714 -3.93945481082393e-09
107.142857142857 -5.47146501503322e-09
110 -6.25570833385465e-09
112.857142857143 -6.40161406758887e-09
115.714285714286 -6.01861151653654e-09
118.571428571429 -5.21612998099834e-09
121.428571428571 -4.10359876127491e-09
124.285714285714 -2.79044715766694e-09
127.142857142857 -1.38610447047509e-09
130 -8.27180612553028e-25
132.857142857143 1.25843695345764e-09
135.714285714286 2.27977708959718e-09
138.571428571429 2.95459110811794e-09
141.428571428571 3.17344970871927e-09
144.285714285714 2.8269235911005e-09
147.142857142857 1.80558345496096e-09
150 0
};
\addplot [semithick, magenta, line width=1.2pt]
table {%
10 0.017
12.0408163265306 0.0287977756367757
14.0816326530612 0.0452054318241083
16.1224489795918 0.0661302376270365
18.1632653061224 0.0914794621105983
20.2040816326531 0.121160374339832
22.2448979591837 0.155080243379777
24.2857142857143 0.19314633829547
26.3265306122449 0.23526592815195
28.3673469387755 0.281346282014255
30.4081632653061 0.331291602857735
32.4489795918367 0.384356082643598
34.4897959183673 0.438343652909992
36.530612244898 0.490862015454948
38.5714285714286 0.539518872076497
40.6122448979592 0.581943580109132
42.6530612244898 0.61744099376392
44.6938775510204 0.64815605075598
46.734693877551 0.67650879431918
48.7755102040816 0.704919267687385
50.8163265306122 0.735757274089845
52.8571428571429 0.769439536576255
54.8979591836735 0.803845657963065
56.9387755102041 0.836685681051137
58.9795918367347 0.865669648641335
61.0204081632653 0.888589474928487
63.0612244897959 0.905120116168672
65.1020408163265 0.916819570679219
67.1428571428571 0.925327708171424
69.1836734693878 0.932284398356581
71.2244897959184 0.939252634615402
73.265306122449 0.946645112641356
75.3061224489796 0.953989026690448
77.3469387755102 0.960788792846664
79.3877551020408 0.966548827193983
81.4285714285714 0.97083311614147
83.469387755102 0.973820627209283
85.5102040816327 0.976053133774986
87.5510204081633 0.978077098425409
89.5918367346939 0.980438983747379
91.6326530612245 0.983607534917868
93.6734693877551 0.987477116881622
95.7142857142857 0.991684655663233
97.7551020408163 0.995865862952768
99.7959183673469 0.999656450440291
101.836734693878 1.00269212981587
103.877551020408 1.00460861276956
105.918367346939 1.00504161099144
107.959183673469 1.00362683617156
110 1
};
\addplot [semithick, magenta, dashed, line width=1.2pt]
table {%
10 0.00800000000000001
12.0408163265306 0.0090057777953144
14.0816326530612 0.00925527814376483
16.1224489795918 0.00888902178456472
18.1632653061224 0.00804752945692744
20.2040816326531 0.00687132190006642
22.2448979591837 0.00550091985319503
24.2857142857143 0.0040768440555267
26.3265306122449 0.0027396152462748
28.3673469387755 0.00162975416465275
30.4081632653061 0.000887405371618438
32.4489795918367 0.000572963637963008
34.4897959183673 0.000568891419624419
36.530612244898 0.000733575764188404
38.5714285714286 0.000925403719240695
40.6122448979592 0.0010037098069767
42.6530612244898 0.000901135011058769
44.6938775510204 0.000674579855625314
46.734693877551 0.000392981301522811
48.7755102040816 0.00012527630959774
50.8163265306122 -6.06762770078507e-05
52.8571428571429 -0.000138929440911648
54.8979591836735 -0.000137981108805077
56.9387755102041 -8.99678546320187e-05
58.9795918367347 -2.70262523363503e-05
61.0204081632653 1.95560147705281e-05
63.0612244897959 3.80157479242535e-05
65.1020408163265 3.61142329074987e-05
67.1428571428571 2.24616461354161e-05
69.1836734693878 5.66816402315812e-06
71.2244897959184 -6.04895210687053e-06
73.265306122449 -1.03516149126397e-05
75.3061224489796 -9.4275368241402e-06
77.3469387755102 -5.58084955810305e-06
79.3877551020408 -1.1156848312592e-06
81.4285714285714 1.8302082036265e-06
83.469387755102 2.83674995356686e-06
85.5102040816327 2.49719370057493e-06
87.5510204081633 1.41788989642201e-06
89.5918367346939 2.05188992879405e-07
91.6326530612245 -5.96648844542959e-07
93.6734693877551 -9.02249472235697e-07
95.7142857142857 -8.31912819830224e-07
97.7551020408163 -5.06908977680808e-07
99.7959183673469 -4.85080361417031e-08
101.836734693878 4.22019914432828e-07
103.877551020408 7.83404783688522e-07
105.918367346939 9.14376481271123e-07
107.959183673469 6.93664916826367e-07
110 0
};
\addplot [semithick, orange2551868, line width=1.2pt]
table {%
10 0.011
12.8571428571429 0.0137321661154287
15.7142857142857 0.0306505334623675
18.5714285714286 0.0591826623645388
21.4285714285714 0.0967561131456652
24.2857142857143 0.140798446129469
27.1428571428571 0.188737221639673
30 0.238
32.8571428571429 0.286884157303744
35.7142857142857 0.337166332722487
38.5714285714286 0.391492981197383
41.4285714285714 0.452335125003461
44.2857142857143 0.518128835094823
47.1428571428571 0.583275231104651
50 0.642
52.8571428571429 0.690170817478092
55.7142857142857 0.730223314156821
58.5714285714286 0.766235109384252
61.4285714285714 0.802130972179752
64.2857142857143 0.838320114002716
67.1428571428571 0.871696188752557
70 0.899
72.8571428571429 0.918052476893912
75.7142857142857 0.93099505089575
78.5714285714286 0.941049279045117
81.4285714285714 0.951335220206952
84.2857142857143 0.962638475228909
87.1428571428571 0.973410186941357
90 0.982
92.8571428571429 0.987222751164957
95.7142857142857 0.989754045613999
98.5714285714286 0.99073468062931
101.428571428571 0.99128110988376
104.285714285714 0.991949884035947
107.142857142857 0.992737650730198
110 0.993616714001526
112.857142857143 0.994559377884944
115.714285714286 0.995537946415463
118.571428571429 0.996524723628095
121.428571428571 0.997492013557852
124.285714285714 0.998412120239748
127.142857142857 0.999257347708793
130 1
132.857142857143 1.00061238114838
135.714285714286 1.00106679518895
138.571428571429 1.00133554615671
141.428571428571 1.00139093808669
144.285714285714 1.00120527501389
147.142857142857 1.00075086097332
150 1
};
\addplot [semithick, orange2551868, dashed, line width=1.2pt]
table {%
10 0.011
12.8571428571429 0.0137321661154287
15.7142857142857 0.0306505334623675
18.5714285714286 0.0591826623645388
21.4285714285714 0.0967561131456652
24.2857142857143 0.140798446129469
27.1428571428571 0.188737221639673
30 0.238
32.8571428571429 0.286884157303744
35.7142857142857 0.337166332722487
38.5714285714286 0.391492981197383
41.4285714285714 0.452335125003461
44.2857142857143 0.518128835094823
47.1428571428571 0.583275231104651
50 0.642
52.8571428571429 0.690170817478092
55.7142857142857 0.730223314156821
58.5714285714286 0.766235109384252
61.4285714285714 0.802130972179752
64.2857142857143 0.838320114002716
67.1428571428571 0.871696188752557
70 0.899
72.8571428571429 0.918052476893912
75.7142857142857 0.93099505089575
78.5714285714286 0.941049279045117
81.4285714285714 0.951335220206952
84.2857142857143 0.962638475228909
87.1428571428571 0.973410186941357
90 0.982
92.8571428571429 0.987222751164957
95.7142857142857 0.989754045613999
98.5714285714286 0.99073468062931
101.428571428571 0.99128110988376
104.285714285714 0.991949884035947
107.142857142857 0.992737650730198
110 0.993616714001526
112.857142857143 0.994559377884944
115.714285714286 0.995537946415463
118.571428571429 0.996524723628095
121.428571428571 0.997492013557852
124.285714285714 0.998412120239748
127.142857142857 0.999257347708793
130 1
132.857142857143 1.00061238114838
135.714285714286 1.00106679518895
138.571428571429 1.00133554615671
141.428571428571 1.00139093808669
144.285714285714 1.00120527501389
147.142857142857 1.00075086097332
150 1
};

\addplot [semithick, lightgray, dashed]
table {%
0 0.25
150 0.25 
};

\addplot [semithick, lightgray, dashed]
table {%
0 0.5
150 0.5 
};

\addplot [semithick, lightgray, dashed]
table {%
0 0.75
150 0.75 
};
\end{axis}

\end{tikzpicture}

%% file: figs/results_prob_clust_plug_vs_consistent/tkz_le_plugin_vs_consistent_rho_0.30_0.50_0.70.tex
\begin{tikzpicture}

\definecolor{darkgray176}{RGB}{176,176,176}
\definecolor{darkorange25512714}{RGB}{255,127,14}
\definecolor{forestgreen4416044}{RGB}{44,160,44}
\definecolor{lightgray204}{RGB}{204,204,204}
\definecolor{magenta}{RGB}{255,0,255}
\definecolor{orange2551868}{RGB}{255,186,8}
\definecolor{steelblue31119180}{RGB}{31,119,180}

\definecolor{orange2551868}{RGB}{164,66,0}

\begin{axis}[
width=0.25\linewidth,
height=1.5in,
at={(0in,2.3in)},
scale only axis,
legend cell align={left},
legend style={
  fill opacity=0.8,
  draw opacity=1,
  text opacity=1,
  at={(0.91,0.5)},
  anchor=east,
  draw=lightgray204
},
tick align=outside,
tick pos=left,
x grid style={darkgray176},
xtick style={color=black},
y grid style={darkgray176},
ymin=-0.01, ymax=1.001,
ytick style={color=black},
xmin=10, xmax=150,
ytick={0, 0.25, 0.5, 0.75, 1},
xlabel={Growing $M$},
title = {LE}
]
\addplot [semithick, steelblue31119180, line width=1.2pt]
table {%
10 0.00385705381451632
10 0.00385681312703091
16 0.0144780252021844
26 0.0590810393514087
30 0.0949221444612329
30 0.0949272385398195
36 0.155346403849123
38 0.178461047668808
46 0.29360926015118
50 0.342931968036015
70 0.657788362840946
90 0.848362873022423
110 0.95763915554308
130 0.988404989305072
130 0.988404638317677
150 0.996014843003849
150 0.996025208082725
};
\addplot [semithick, steelblue31119180, dashed, line width=1.2pt]
table {%
10 0.00600000000000001
10 0.00600000000000001
16 0.021
26 0.0659999999999999
30 0.091
30 0.091
36 0.125
38 0.149
46 0.243
50 0.259
50 0.259
70 0.459
90 0.653
90 0.653
110 0.767
130 0.883
130 0.883
150 0.918
150 0.918
};
\addplot [semithick, darkorange25512714, line width=1.2pt]
table {%
10 0.0390366453028962
10 0.0390362884489475
20 0.252047341042916
30 0.550942431586467
30 0.550941466509454
40 0.802751626775648
50 0.91530064390948
50 0.915295614137382
60 0.977902078167426
70 0.994096969041111
80 0.998619122053197
90 0.999803865912957
90 0.999807480243494
};
\addplot [semithick, darkorange25512714, dashed, line width=1.2pt]
table {%
10 0
10 0
20 0
30 0
30 0
40 0
50 0
50 0
60 0
70 0
80 0
90 0
90 0
};
\addplot [semithick, forestgreen4416044, line width=1.2pt]
table {%
10 0.0286590823968757
20 0.294012308877831
30 0.628421141185282
40 0.827024505689929
50 0.924977422833833
50 0.924983323104392
60 0.971767786945861
70 0.990252366133427
80 0.998370695476048
90 0.999543390044083
100 0.999899369683797
130 1.00000250884008
150 0.999997710584186
};
\addplot [semithick, forestgreen4416044, dashed, line width=1.2pt]
table {%
10 0.003
10 0.003
20 0
30 0
30 0
40 0
50 0
50 0
60 0
70 0
80 0
90 0
90 0
100 0
130 0
130 0
150 0
150 0
};
\addplot [semithick, magenta, line width=1.2pt]
table {%
10 0.0190400437077476
10 0.0190434913522922
20 0.149056820968944
30 0.390380602602429
30 0.390384717415026
40 0.627182188741088
50 0.807161910968298
50 0.807163915518295
60 0.923817679164622
70 0.975276361220345
80 0.992464249024621
90 0.998185238105448
90 0.998190515009491
};
\addplot [semithick, magenta, dashed, line width=1.2pt]
table {%
10 0.005
10 0.005
20 0.002
30 0
30 0
40 0
50 0
50 0
60 0
70 0
80 0
90 0
100 0.00
150 0.00
};
\addplot [semithick, orange2551868, line width=1.2pt]
table {%
10 0.017
20 0.105
30 0.314
40 0.525
50 0.745
60 0.878
70 0.945
80 0.986
90 0.991
100 0.997
130 1
150 1
};
\addplot [semithick, orange2551868, dashed, line width=1.2pt]
table {%
10 0.021
20 0.111
30 0.314
40 0.53
50 0.696
60 0.851
70 0.916
80 0.964
90 0.986
100 0.994
130 0.999
150 1
};

\addplot [semithick, lightgray, dashed]
table {%
0 0.25
150 0.25 
};

\addplot [semithick, lightgray, dashed]
table {%
0 0.5
150 0.5 
};

\addplot [semithick, lightgray, dashed]
table {%
0 0.75
150 0.75 
};
\end{axis}

\end{tikzpicture}

%% file: figs/results_prob_clust_plug_vs_consistent/other_rho_val/tkz_eu_plugin_vs_consistent_rho_0.30_0.50_0.70_other.tex
\begin{tikzpicture}

\definecolor{darkgray176}{RGB}{176,176,176}
\definecolor{darkorange25512714}{RGB}{255,127,14}
\definecolor{forestgreen4416044}{RGB}{44,160,44}
\definecolor{lightgray204}{RGB}{204,204,204}
\definecolor{magenta}{RGB}{255,0,255}
\definecolor{orange2551868}{RGB}{255,186,8}
\definecolor{steelblue31119180}{RGB}{31,119,180}
\definecolor{orange2551868}{RGB}{164,66,0}

\begin{axis}[
width=0.25\linewidth,
height=1.5in,
at={(0in,2.3in)},
scale only axis,
legend cell align={left},
legend style={
  fill opacity=0.8,
  draw opacity=1,
  text opacity=1,
  at={(0.91,0.5)},
  anchor=east,
  draw=lightgray204
},
tick align=outside,
tick pos=left,
x grid style={darkgray176},
xtick style={color=black},
y grid style={darkgray176},
ymin=-0.01, ymax=1.001,
ytick style={color=black},
xmin=10, xmax=150,
ytick={0, 0.25, 0.5, 0.75, 1},
ylabel={Prob. Correct Clustering},
xlabel={Growing $M$}
]

\addplot [semithick, steelblue31119180, line width=1.2pt]
table {%
10 0.081
20 0.2934
25 0.4014
30 0.4824
35 0.5414
40 0.608
50 0.694
70 0.779
90 0.8191
100 0.8569
110 0.8847
120 0.9023
130 0.9242
140 0.9428
150 0.958
};
\addplot [semithick, steelblue31119180, dashed, line width=1.2pt]
table {%
10 0.052
20 0.2278
25 0.332
30 0.4262
35 0.4932
40 0.573
50 0.662
70 0.76
90 0.8034
100 0.8414
110 0.8686
120 0.8897
130 0.9143
140 0.9299
150 0.9504
};
\addplot [semithick, darkorange25512714, line width=1.2pt]
table {%
10 0.198
20 0.568
25 0.688
30 0.7678
35 0.8124
40 0.854
50 0.905
60 0.949
70 0.968
80 0.987
90 0.9893
100 0.9967
110 0.9984
120 0.9991
130 0.9999
140 1
150 1
};
\addplot [semithick, darkorange25512714, dashed, line width=1.2pt]
table {%
10 0.147
20 0.4762
25 0.6158
30 0.7136
35 0.809
40 0.884
50 0.972
60 0.994
70 0.996
80 1
90 0.9997
100 1
110 1
120 1
130 1
140 1
150 1
};
\addplot [semithick, forestgreen4416044, line width=1.2pt]
table {%
10 0.208
20 0.5414
25 0.6024
30 0.6554
35 0.6888
40 0.711
60 0.806
80 0.881
90 0.9103
100 0.931
110 0.95
120 0.9635
130 0.9767
140 0.982
150 0.9899
};
\addplot [semithick, forestgreen4416044, dashed, line width=1.2pt]
table {%
10 0.147
20 0.3022
25 0.3106
30 0.3494
35 0.357
40 0.35
50 0.354
60 0.342
70 0.338
90 0.3557
110 0.3582
120 0.3576
130 0.3641
140 0.3555
150 0.3672
};
\addplot [semithick, magenta, line width=1.2pt]
table {%
10 0.139
20 0.4458
25 0.5526
30 0.5958
35 0.6668
40 0.695
50 0.758
70 0.841
80 0.881
90 0.8994
100 0.9218
110 0.9498
120 0.9643
130 0.9736
140 0.9833
150 0.987
};
\addplot [semithick, magenta, dashed, line width=1.2pt]
table {%
10 0.091
20 0.2636
25 0.3292
30 0.3686
35 0.4358
40 0.484
50 0.552
60 0.609
70 0.662
90 0.7405
100 0.7713
110 0.8087
120 0.8389
130 0.866
140 0.8878
150 0.9045

};
\addplot [semithick, orange2551868, line width=1.2pt]
table {%
10 0.122
20 0.4088
25 0.5188
30 0.581
35 0.6536
40 0.684
50 0.723
60 0.792
70 0.841
80 0.883
90 0.8983
100 0.9251
110 0.9444
120 0.9617
130 0.9736
140 0.9843
150 0.9888

};
\addplot [semithick, orange2551868, dashed, line width=1.2pt]
table {%
10 0.085
20 0.356
25 0.4708
30 0.5434
35 0.6294
40 0.66
50 0.708
60 0.777
70 0.832
80 0.87
90 0.8873
100 0.9194
110 0.9386
120 0.9576
130 0.9696
140 0.9803
150 0.9856
};

\addplot [semithick, lightgray, dashed]
table {%
0 0.25
150 0.25 
};

\addplot [semithick, lightgray, dashed]
table {%
0 0.5
150 0.5 
};

\addplot [semithick, lightgray, dashed]
table {%
0 0.75
150 0.75 
};
\end{axis}

\end{tikzpicture}

%% file: figs/results_prob_clust_plug_vs_consistent/other_rho_val/tkz_kl_plugin_vs_consistent_esitmator_rho2.tex
\begin{tikzpicture}

\definecolor{darkgray176}{RGB}{176,176,176}
\definecolor{darkorange25512714}{RGB}{255,127,14}
\definecolor{forestgreen4416044}{RGB}{44,160,44}
\definecolor{lightgray204}{RGB}{204,204,204}
\definecolor{magenta}{RGB}{255,0,255}
\definecolor{orange2551868}{RGB}{255,186,8}
\definecolor{steelblue31119180}{RGB}{31,119,180}
\definecolor{orange2551868}{RGB}{164,66,0}

\begin{axis}[
width=0.25\linewidth,
height=1.5in,
at={(0in,2.3in)},
scale only axis,
legend cell align={left},
legend style={
  fill opacity=0.8,
  draw opacity=1,
  text opacity=1,
  at={(0.91,0.5)},
  anchor=east,
  draw=lightgray204
},
tick align=outside,
tick pos=left,
x grid style={darkgray176},
xtick style={color=black},
y grid style={darkgray176},
ymin=-0.01, ymax=1.001,
ytick style={color=black},
xmin=10, xmax=150,
ytick={0, 0.25, 0.5, 0.75, 1},
xlabel={Growing $M$}
]

\addplot [semithick, steelblue31119180, line width=1.2pt]
table {%
10 0.0570000000000001
20 0.2364
25 0.366
30 0.4946
35 0.6066
40 0.741
50 0.879
60 0.97
70 0.993
80 0.997
150 1
};
\addplot [semithick, steelblue31119180, dashed, line width=1.2pt]
table {%
10 0.0570000000000001
20 0.2364
25 0.366
30 0.4946
35 0.6066
40 0.741
50 0.879
60 0.97
70 0.993
80 0.997
150 1
};
\addplot [semithick, darkorange25512714, line width=1.2pt]
table {%
10 0.299
20 0.8278
25 0.9434
30 0.979
35 0.9954
40 0.998
50 1
60 1
70 1
80 1
150 1
};
\addplot [semithick, darkorange25512714, dashed, line width=1.2pt]
table {%
10 0.113
20 0.1102
25 0.101
30 0.0842000000000001
35 0.0884
40 0.073
50 0.0610000000000001
60 0.0649999999999999
70 0.038
80 0.041
150 0
};
\addplot [semithick, forestgreen4416044, line width=1.2pt]
table {%
10 0.382
20 0.843
25 0.9292
30 0.9696
35 0.9904
40 0.998
50 1
60 1
70 1
80 1
150 1
};
\addplot [semithick, forestgreen4416044, dashed, line width=1.2pt]
table {%
10 0.015
20 0.002
25 0.000199999999999978
30 0.000199999999999978
35 0
40 0
50 0
60 0
70 0
80 0
150 0
};
\addplot [semithick, magenta, line width=1.2pt]
table {%
10 0.217
20 0.6928
25 0.8544
30 0.9426
35 0.9788
40 0.991
50 1
60 1
70 1
80 1
150 1
};
\addplot [semithick, magenta, dashed, line width=1.2pt]
table {%
10 0.11
20 0.2418
25 0.3454
30 0.4664
35 0.575
40 0.677
50 0.799
60 0.924
70 0.959
80 0.986
150 1
};
\addplot [semithick, orange2551868, line width=1.2pt]
table {%
10 0.15
20 0.5902
25 0.7636
30 0.9
35 0.9648
40 0.991
50 1
60 1
70 1
80 1
150 1
};
\addplot [semithick, orange2551868, dashed, line width=1.2pt]
table {%
10 0.15
20 0.5902
25 0.7636
30 0.9
35 0.9648
40 0.991
50 1
60 1
70 1
80 1
150 1
};

\addplot [semithick, lightgray, dashed]
table {%
0 0.25
150 0.25 
};

\addplot [semithick, lightgray, dashed]
table {%
0 0.5
150 0.5 
};

\addplot [semithick, lightgray, dashed]
table {%
0 0.75
150 0.75 
};
\end{axis}

\end{tikzpicture}

%% file: figs/results_prob_clust_plug_vs_consistent/other_rho_val/tkz_le_plugin_vs_consistent_rho_0.30_0.50_0.70_other.tex
\begin{tikzpicture}

\definecolor{darkgray176}{RGB}{176,176,176}
\definecolor{darkorange25512714}{RGB}{255,127,14}
\definecolor{forestgreen4416044}{RGB}{44,160,44}
\definecolor{lightgray204}{RGB}{204,204,204}
\definecolor{magenta}{RGB}{255,0,255}
\definecolor{orange2551868}{RGB}{255,186,8}
\definecolor{steelblue31119180}{RGB}{31,119,180}
\definecolor{orange2551868}{RGB}{164,66,0}

\begin{axis}[
width=0.25\linewidth,
height=1.5in,
at={(0in,2.3in)},
scale only axis,
legend cell align={left},
legend style={
  fill opacity=0.8,
  draw opacity=1,
  text opacity=1,
  at={(0.91,0.5)},
  anchor=east,
  draw=lightgray204
},
tick align=outside,
tick pos=left,
x grid style={darkgray176},
xtick style={color=black},
y grid style={darkgray176},
ymin=-0.01, ymax=1.001,
ytick style={color=black},
xmin=10, xmax=150,
ytick={0, 0.25, 0.5, 0.75, 1},
xlabel={Growing $M$}
]
\addplot [semithick, steelblue31119180, line width=1.2pt]
table {%
10 0.091
20 0.3714
25 0.5664
30 0.7112
35 0.817
40 0.911
50 0.978
60 0.997
70 1
80 1
90 1
150 1
};
\addplot [semithick, steelblue31119180, dashed, line width=1.2pt]
table {%
10 0.115
20 0.4184
25 0.5906
30 0.7078
35 0.7924
40 0.87
50 0.941
60 0.985
70 0.997
80 0.995
90 1
150 1
};
\addplot [semithick, darkorange25512714, line width=1.2pt]
table {%
10 0.339
20 0.8872
25 0.9682
30 0.9916
35 0.9988
40 1
50 1
60 1
70 1
80 1
90 1
150 1
};
\addplot [semithick, darkorange25512714, dashed, line width=1.2pt]
table {%
10 0.113
20 0.0956
25 0.0814
30 0.0679999999999999
35 0.0646
40 0.0580000000000001
50 0.047
60 0.042
70 0.013
80 0.015
90 0.011
150 0
};
\addplot [semithick, forestgreen4416044, line width=1.2pt]
table {%
10 0.424
20 0.9142
25 0.975
30 0.9912
35 0.9992
40 1
50 1
60 1
70 1
80 1
90 1
150 1
};
\addplot [semithick, forestgreen4416044, dashed, line width=1.2pt]
table {%
10 0.092
20 0.09
25 0.0669999999999999
30 0.0600000000000001
35 0.0488
40 0.044
50 0.021
60 0.011
70 0.00600000000000001
80 0.004
90 0
150 0
};
\addplot [semithick, magenta, line width=1.2pt]
table {%
10 0.265
20 0.8006
25 0.9344
30 0.9814
35 0.9958
40 0.999
50 1
60 1
70 1
80 1
90 1
150 1
};
\addplot [semithick, magenta, dashed, line width=1.2pt]
table {%
10 0.249
20 0.6528
25 0.8238
30 0.9196
35 0.9658
40 0.987
50 0.999
60 1
70 1
80 1
90 1
150 1
};
\addplot [semithick, orange2551868, line width=1.2pt]
table {%
10 0.206
20 0.72
25 0.8682
30 0.9562
35 0.993
40 1
50 1
60 1
70 1
80 1
90 1
150 1
};
\addplot [semithick, orange2551868, dashed, line width=1.2pt]
table {%
10 0.238
20 0.7384
25 0.8764
30 0.9488
35 0.9854
40 0.999
50 1
60 1
70 1
80 1
90 1
150 1
};

\addplot [semithick, lightgray, dashed]
table {%
0 0.25
150 0.25 
};

\addplot [semithick, lightgray, dashed]
table {%
0 0.5
150 0.5 
};

\addplot [semithick, lightgray, dashed]
table {%
0 0.75
150 0.75 
};
\end{axis}

\end{tikzpicture}